\documentclass[sigconf,9pt]{acmart}

\setcopyright{none}
\renewcommand\footnotetextcopyrightpermission[1]{}
\AtBeginDocument{%
  }
\usepackage{makecell}
\usepackage{array}
\usepackage{microtype}
\usepackage{graphicx}
\usepackage{subfigure}
\usepackage{booktabs} 
\usepackage{listings}
\usepackage{xcolor}
\usepackage{amsmath}
\usepackage{multirow}
\usepackage{booktabs}
\usepackage{booktabs}
\usepackage{float}
\usepackage{pifont}
\usepackage[most]{tcolorbox}
\usepackage{xcolor}
\usepackage{placeins}

\begin{document}

\title{MobileFineTuner: A Mobile-Native Framework for On-Device LLM Fine-Tuning in Real-World Embedded AI Applications}

\author{*Jiaxiang Geng$^{1,2}$ \quad *Lunyu Zhao$^1$ \quad *Yiyi Lu$^1$ \quad Bing Luo$^1$}
\affiliation{
    \institution{$^1$Duke Kunshan University, Kunshan, China}
    \institution{$^2$The University of Hong Kong, Hong Kong, China}
    \country{} 
}
\email{{jg645,lz269,yl996,bl291}@duke.edu}
\thanks{* These authors contribute equally to this work. Corresponding Author: Bing Luo.}

\renewcommand{\shortauthors}{Geng et al.}
\renewcommand{\shorttitle}{MobileFineTuner}

\begin{abstract}
Large language models (LLMs) are moving from cloud-centric services toward on-device embedded AI, where models interact with private, longitudinal signals sensed from users and their physical environments. Mobile phones are a natural platform for such applications because they are continuously carried by users, connected to wearable sensors, and deeply integrated with daily mobile applications. However, practical LLM fine-tuning on commodity phones remains difficult. Existing fine-tuning frameworks are largely Python-based and server-oriented, making them hard to deploy inside mobile applications. We present MobileFineTuner, a mobile-native open-source framework for end-to-end LLM fine-tuning on commodity mobile phones. MobileFineTuner is implemented in C++ and provides a reusable training stack. To make fine-tuning feasible under mobile resource constraints, MobileFineTuner integrates a resource-aware training runtime with memory-efficient attention, activation checkpointing, gradient accumulation, parameter sharding, and energy-aware scheduling. We evaluate MobileFineTuner on real mobile phones using GPT-2, Gemma 3, and Qwen2.5 models across multiple fine-tuning tasks. The results show that MobileFineTuner reproduces standard Full-FT and LoRA fine-tuning behavior, substantially reduces memory pressure and improves executability on memory-constrained phones. We further demonstrate MobileFineTuner through a private campus health-agent application, where a local LLM is fine-tuned on user-specific wearable-sensing records to provide more personalized responses while keeping raw records on the phone. These results establish MobileFineTuner as a practical toolkit for studying and building on-device LLM fine-tuning applications in embedded AI and sensing systems.
\end{abstract}

\ccsdesc[500]{Computing methodologies~Distributed artificial intelligence}

\keywords{Fine-tuning, Mobile phones, Large language models}


\maketitle

\section{Introduction}

Large language models (LLMs) are moving from cloud-centric services toward on-device embedded AI, where model intelligence is coupled with signals sensed from users and their physical environments~\cite{ren2025sensorloop}. Among edge platforms, the mobile phone is a natural host for personal AI: it is carried throughout the day, connected to wearable sensors, and exposed to longitudinal local context such as physical activity, sleep, heart rate, mobility, screen time, calendar routines, and app interactions~\cite{google2024personalhealth}. This enables a new class of personal embedded AI agents that serve a single user. For example, a personal health agent application on the phone could learn the user's behavior from locally stored data, detect deviations in recent activity patterns, and generate feedback tailored to the user's current state.

Realizing such agent applications requires adapting the base LLM with the user's own data. Existing personal-agent pipelines remain largely cloud-centric: models are trained or fine-tuned on servers using public datasets or personal data collected from end devices, while the mobile phone mainly serves as an inference endpoint~\cite{li2024personalllm,merrill2026wearablehealth}. However, this design creates a fundamental tension. If adaptation relies only on public or population-level data, the agent cannot fully capture the user's individual context; if raw personal traces are uploaded for server-side training, the system exposes sensitive user data and may create serious privacy risks~\cite{Thirunavukarasu2023,GDPR2016a}. To resolve this tension, model adaptation should happen where the personal data is generated and stored: on the user's mobile phone. This requires mobile phones to support practical \emph{on-device LLM fine-tuning}, allowing the model to adapt to private user data without exposing raw personal traces.

Despite this need, practical LLM fine-tuning on commodity mobile phones remains difficult because it must address two system requirements: 

\noindent \textbf{(RQ1) Mobile-native and reusable framework.} 
The fine-tuning stack should run directly in mobile environments (e.g. Android) and be reusable by researchers and developers across different applications, rather than relying on server-side runtimes.

\noindent \textbf{(RQ2) Resource-aware mobile training.} 
Unlike inference, LLM fine-tuning involves repeated forward and backward propagation, activation storage, gradient computation, and parameter updates, which introduce substantial memory and energy overhead on resource-constrained phones.

However, existing mature LLM fine-tuning frameworks fail to satisfy \textbf{RQ1} in mobile environments. Popular frameworks, such as PyTorch and Hugging Face Transformers, rely on Python and native Linux/Windows runtimes~\cite{NEURIPS2019_bdbca288,DBLP:journals/corr/abs-1910-03771}, making them difficult to deploy directly inside mobile operating systems and mobile applications. Mobile phones, in contrast, do not natively support Python-centered training stacks~\cite{python3docs}. As a result, these mature frameworks cannot be directly reused for mobile-phone LLM fine-tuning.

A few recent efforts have attempted LLM fine-tuning directly on mobile phones~\cite{peng-etal-2024-pocketllm,10.1145/3711875.3729132}. However, they still fall short of both \textbf{RQ1} and \textbf{RQ2}. Termux-based solutions emulate a Python environment on the phone, which introduces runtime overhead and is difficult to integrate directly into normal mobile applications. ONNX-based workflows rely on a server-side runtime, where the user's local data must be sent to the server and processed by Python-based toolkits to construct an ONNX computation graph, which is then sent back to the mobile phone to complete the final parameter updates. Moreover, these approaches do not provide a general mobile training runtime with built-in support for managing the memory and energy costs of LLM fine-tuning, making the training process prone to resource-related failures such as out-of-memory errors during implementation.

\begin{figure}[t]
\centering
\includegraphics[width=\linewidth]{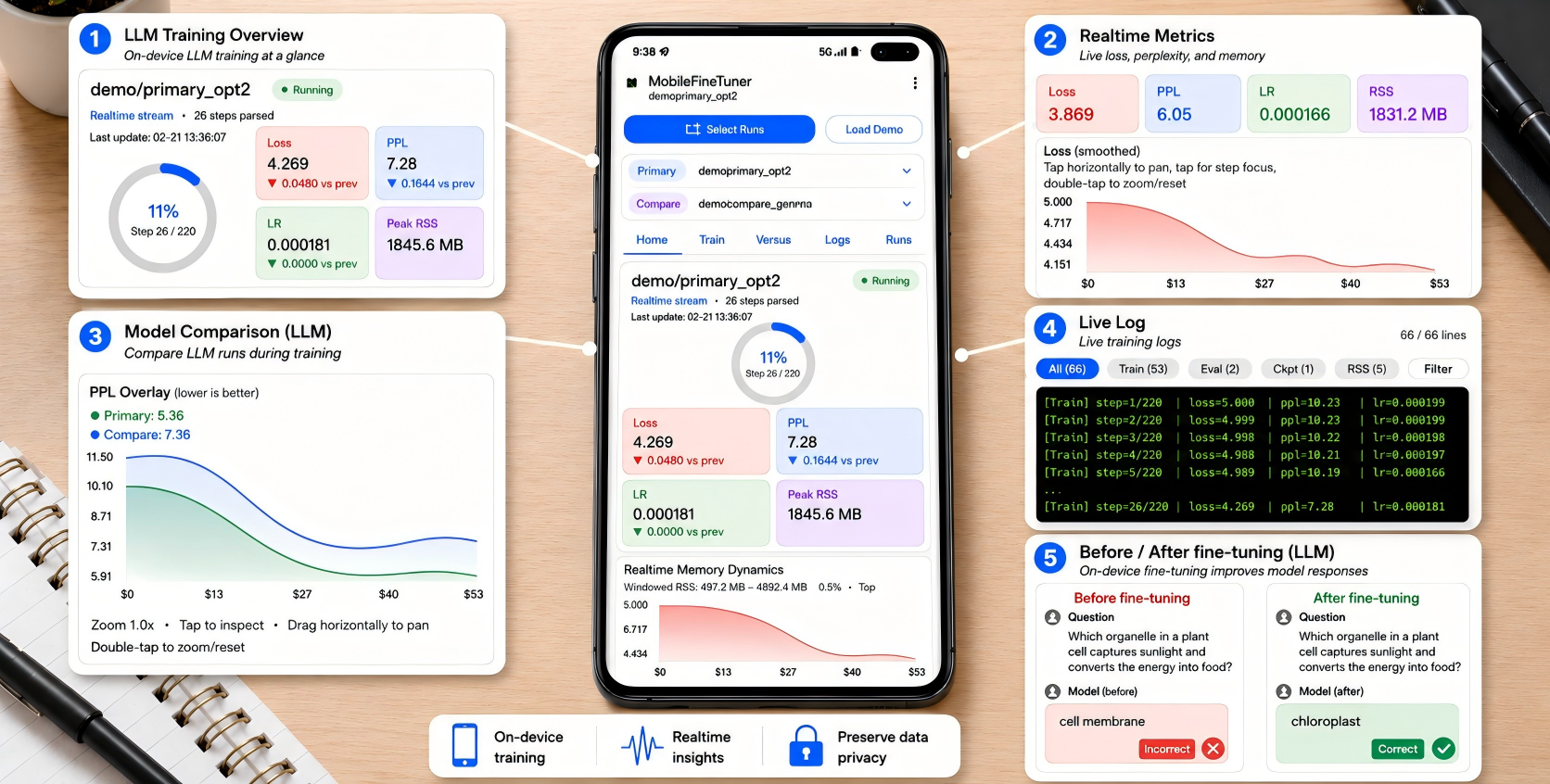} 
\vspace{-4mm}
\caption{MobileFineTuner realizes real-system LLM fine-tuning on a commodity mobile phone.}
\label{application}
\vspace{-4mm}
\end{figure}

The above infrastructure gap has shaped how existing private-domain LLM fine-tuning studies are implemented and evaluated. Many private-domain LLM fine-tuning studies avoid commodity mobile phones and instead rely on two implementation workarounds. First, some studies evaluate their methods in simulation, where clients are represented by server-side processes rather than real phones~\cite{10447454,wang2024flora,10.1145/3723877,cho-etal-2024-heterogeneous,wagner2024personalized}. However, simulation depends on manually specified system assumptions, misses real mobile-system bottlenecks, and cannot demonstrate deployability as an actual personal AI application. Second, other studies implement their methods on edge devices such as NVIDIA developer boards, Raspberry Pis, or PCs~\cite{chen2025memoryefficientsplitfederatedlearning,yuan2025flexiblepersonalizedsplitfederated,li2025mobillmenablingllmfinetuning,yang2025paemobillmprivacyawareefficient}. These devices are useful for controlled experiments, but they are not the primary devices through which users carry, sense, and interact with personal AI in daily life. Unlike mobile phones, they may not be continuously carried by users, nor deeply integrated with everyday mobile applications. The contrast between real mobile-application needs and current implementation practice raises the key question addressed in this paper:

\begin{center}
\emph{How can we build a unified fine-tuning framework that is efficient enough for resource-constrained mobile phones, scalable across models and tasks, and usable for building real mobile embedded AI applications?}
\end{center}

\begin{figure*}[t]
\centering
\includegraphics[width=0.9\textwidth]{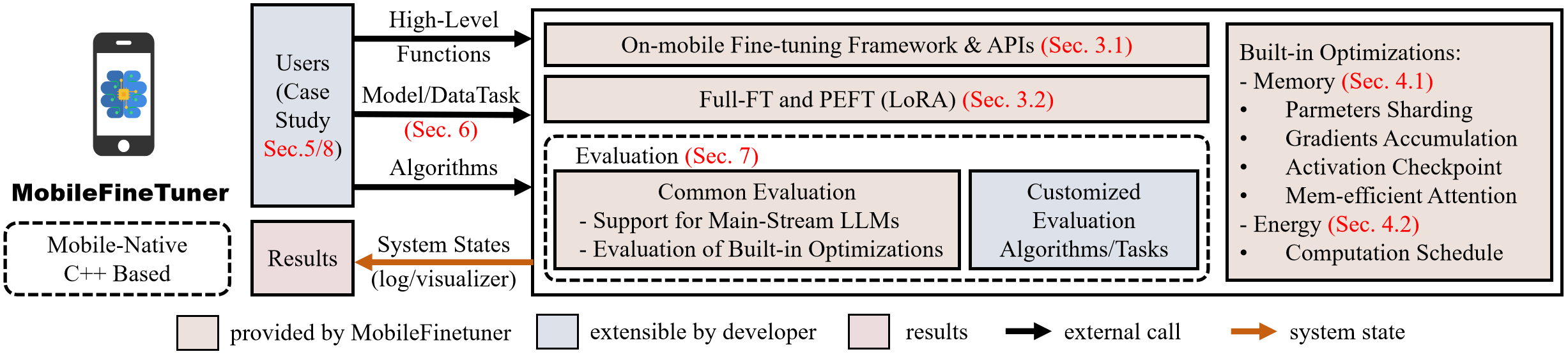} 
\vspace{-3mm}
\caption{MobileFineTuner overview.}
\label{overview}
\vspace{-4mm}
\end{figure*}

In this paper, we present \textbf{MobileFineTuner}, a mobile-native open-source framework for on-device LLM fine-tuning on commodity mobile phones\footnote{MobileFineTuner focuses on enabling practical single-device LLM fine-tuning on mobile phones. This serves as a foundational step toward future extensions to multi-device collaborative learning scenarios, such as federated fine-tuning.}. MobileFineTuner is designed as reusable infrastructure for embedded AI and sensing-system researchers who need to adapt LLMs locally over private user data. Instead of relying on Python runtimes or external machine-learning frameworks, MobileFineTuner is implemented in C++ and provides an end-to-end training stack directly executable on mobile devices. It supports both full-parameter fine-tuning (Full-FT) and parameter-efficient fine-tuning (PEFT, e.g. LoRA) and provide high-level APIs for reuse. It also supports loading and exporting standard model formats, enabling compatibility with existing LLM ecosystems. With this design, MobileFineTuner allows researchers to prototype local adaptation algorithms, evaluate real-device training behavior, and integrate fine-tuning into mobile applications without rebuilding the entire training pipeline.

MobileFineTuner is built around three design goals. First, it aims for \textbf{efficiency}. Its mobile-native C++ runtime avoids the overhead of Python-based execution and provides fine-grained control over tensor operations, automatic differentiation, parameter updates, and memory management. Second, it aims for \textbf{scalability}. MobileFineTuner supports multiple LLM families and both Full-FT and PEFT, allowing researchers to evaluate different model sizes, adaptation strategies, and deployment settings on real mobile hardware. Third, it aims for \textbf{usability}. It provides high-level training APIs, model loading/export support and runtime metric logging. It is also equiped with a training visualizer, so that researchers and application developers can monitor loss, accuracy or perplexity, memory usage, power behavior, and training progress during mobile-side fine-tuning.

To make LLM fine-tuning feasible under mobile resource constraints, MobileFineTuner integrates a resource-aware runtime including several built-in system-level optimization methods. For \textbf{memory efficiency}, it implements a ZeRO-inspired parameter sharding mechanism that offloads inactive parameter segments to disk, gradient accumulation that enables large effective batch sizes under small memory budgets, activation checkpointing that trades recomputation for reduced activation storage, and a memory-efficient attention operator that avoids materializing full quadratic attention matrices. For \textbf{energy efficiency}, it provides an energy-aware computation scheduler that monitors battery state and dynamically adjusts training frequency to reduce power pressure during sustained fine-tuning. Our experiments show that these optimization mechanisms are critical for training on resource-constrained devices: several model-device configurations that fail with out-of-memory errors without optimization can complete training successfully after enabling MobileFineTuner's optimizations.

Furthermore, we demonstrate MobileFineTuner through a real-world embedded AI application, a private campus health agent. In this case study, personal health and behavior data remain on the user's phone and are used to locally adapt an LLM into a personalized health assistant. The agent observes individual trends such as sleep duration, activity level, walking distance, heart rate, and screen-time usage, and then answers user questions based on the user's own historical patterns. This case study illustrates the practical value of MobileFineTuner by showing how a privacy-preserving mobile application can improve personalization through local LLM fine-tuning. It also provides an end-to-end example showing that MobileFineTuner can be directly integrated as the computation backend of mobile embedded AI applications, allowing developers to realize on-device LLM fine-tuning without building task-specific training pipelines from scratch.

In summary, this paper makes the following contributions:
\begin{itemize}
    \item \textbf{A mobile-native toolkit for on-device LLM fine-tuning.}
    We present MobileFineTuner, an open-source C++ framework that enables end-to-end Full-FT and PEFT of LLMs directly on commodity mobile phones, providing reusable infrastructure for embedded AI and sensing-system researchers. Our code is at: \url{https://github.com/Edge-Intelligence-Lab/MobileFineTuner}.

    \item \textbf{A resource-aware training runtime for constrained mobile hardware.}
    MobileFineTuner integrates parameter sharding, gradient accumulation, activation checkpointing, memory-efficient attention, and energy-aware scheduling to make LLM fine-tuning feasible under mobile memory and energy constraints.

    \item \textbf{A practical programming interface and observability stack.}
    MobileFineTuner provides high-level APIs, model loading/export support, runtime metric collection, and a training visualizer, allowing researchers to prototype, monitor, and compare on-device fine-tuning workloads without building device-specific training pipelines from scratch.

    \item \textbf{A real-world health-agent case study.}
    We demonstrate MobileFineTuner in a private campus health-agent application, where a mobile LLM is locally adapted using personal health and behavioral sensing data. The case study shows that on-device fine-tuning can improve personalized response quality while keeping raw user data on the phone.
\end{itemize}

\textbf{Roadmap:}  The roadmap of this paper is shown in Fig.~\ref{overview}. 
Section~2 discusses the motivation and related work. 
Section~3 introduces MobileFineTuner, including its framework architecture, programming interfaces, Full-FT and PEFT workflows, and comparison with existing LLM implementation frameworks. 
Section~4 presents the resource-aware training runtime for constrained mobile hardware, including built-in memory and energy optimizations. 
Section~5 describes our private campus health-agent case study and explains how MobileFineTuner is integrated as the on-device fine-tuning backend. 
Section~6 provides the implementation details, including framework compilation, programming interface, supported models, and the training visualizer. 
Section~7 evaluates MobileFineTuner on common fine-tuning workloads, including correctness validation, resource-aware runtime evaluation, and comparison with a Termux-based pipeline. 
Section~8 reports the application-level results of the private campus health-agent case study.

\section{Motivation and Related Work}

\begin{table*}[t]
\centering
\caption{Recent methods on leveraging private-domain data for LLM fine-tuning.}
\vspace{-2mm}
\label{tab:ondevice_llm}
\resizebox{\linewidth}{!}{
\begin{tabular}{lcccc}
\toprule
\textbf{Method} & \textbf{On-device Fine-tuning Target} & \textbf{Based Framework} & \textbf{Implementation Setup} & \textbf{Open Source} \\
\midrule
FedIT \cite{10447454}        & $\surd$ & Transformers and Huggingface PEFT  & Simulation                            & $\surd$ \\
FLoRA \cite{wang2024flora}        & $\surd$ & PyTorch and Transformers           & Simulation                            & $\surd$ \\
HeLoRA \cite{10.1145/3723877}       & $\surd$ & PyTorch and Flower                 & Simulation                            & $\surd$ \\
HetLoRA \cite{cho-etal-2024-heterogeneous}      & $\surd$ & Not mentioned                      & Simulation                            & $\times$ \\
PCFT-LLM  \cite{wagner2024personalized}    & $\surd$ & PyTorch and Transformers           & Simulation                            & $\surd$ \\
ME-SFL \cite{chen2025memoryefficientsplitfederatedlearning} & $\surd$ & Not mentioned                      & NVIDIA developer boards, SoCs and PCs & $\times$ \\
FlexP-SFL \cite{yuan2025flexiblepersonalizedsplitfederated}    & $\surd$ & PyTorch                            & NVIDIA developer boards and RasberryPi & $\surd$ \\
MobiLLM \cite{li2025mobillmenablingllmfinetuning}      & $\surd$ & PyTorch and Transformers           & NVIDIA developer boards and PCs       & $\times$ \\
PAE-MobiLLM  \cite{yang2025paemobillmprivacyawareefficient} & $\surd$ & PyTorch and Transformers           & NVIDIA developer boards and PCs       & $\times$ \\
PocketLLM \cite{peng-etal-2024-pocketllm}     & $\surd$ & PyTorch and Transformers by Termux \textcolor{red}{(runtime overhead; limited app integration)} & \textcolor{blue}{Mobile Phones}                         & $\times$ \\
XPerT \cite{10.1145/3711875.3729132}         & $\surd$ & ONNX Runtime Python API \textcolor{red}{(server-assisted; partial mobile pipeline)}           & \textcolor{blue}{Mobile Phones}                         & $\times$ \\
SP-LLM \cite{10.1145/3649329.3655665}       & $\surd$ & Not mentioned                      & Simulation                            & $\times$ \\
Crayon \cite{bang-etal-2024-crayon}       & $\surd$ & PyTorch and Huggingface PEFT       & Simulation                            & $\times$ \\
\bottomrule
\end{tabular}
}
\end{table*}

In this section, we review prior work related to LLM fine-tuning from two perspectives. First, we discuss methods that leverage private-domain data for on-device LLM fine-tuning, examine the implementation setups they rely on, and analyze why they do not typically adopt mobile phones as target platforms. Second, we review recent efforts that attempt to enable LLM fine-tuning on mobile phones, analyzing their implementation strategies and inherent limitations.

\vspace{-3mm}
\subsection{Leveraging Private-Domain Data for on-device LLM Fine-Tuning}

To leverage private-domain data for LLM fine-tuning, recent works have proposed several approaches that allow data on end devices to participate in the fine-tuning process while preserving privacy by ensuring the data never leaves the device. We summarize these works in Tab.~\ref{tab:ondevice_llm}, highlighting the frameworks, implementation setups, and open-source availability. The comparison reveals a notable gap: although all of these methods aim at enabling on-device LLM fine-tuning, among the 13 surveyed works, 7 works (over half) rely on simulation-based experiments, 4 works employ NVIDIA developer boards, SoCs, Raspberry Pis or PCs, while only 2 works (PocketLLM and XPerT) actually evaluate their methods on mobile phones. This raises a critical question: given that mobile phones are the most ubiquitous end devices in daily life, generating massive amounts of human-generated data and serving as the primary target platform for most end-side applications, \textbf{why are mobile phones rarely used as the actual implementation platform?}

To answer this question, we first examine the implementation frameworks on which these works are based. Except for two studies that utilize mobile phones (which will be discussed in Section 2.2), the remaining works rely on mature frameworks such as PyTorch \cite{NEURIPS2019_bdbca288}, Hugging Face Transformers \cite{DBLP:journals/corr/abs-1910-03771}, Hugging Face PEFT \cite{peft}, and Flower \cite{beutel2020flower}. A common characteristic of these frameworks is that they are all Python-based. These frameworks can be conveniently deployed on operating systems that support Python, such as Linux and Windows, and they have streamlined the LLM fine-tuning process, allowing researchers and developers to readily reuse packages for algorithmic development with minimal coding effort.

However, mobile phones, such as those running Android, do not natively support Python \cite{python3docs}, which means the aforementioned frameworks cannot be directly deployed. Currently, there is no unified framework for LLM fine-tuning on mobile phones. As a result, implementing real-world LLM fine-tuning on mobile phones requires significant effort from researchers to develop thousands of lines of custom code. This is why many studies opt for simulations or use devices running Linux or Windows, rather than mobile phones. Consequently, a significant portion of mobile phone scenarios has been largely overlooked in practical implementations. This gap motivated us to develop MobileFineTuner, a unified open-source framework for LLM fine-tuning on mobile phones, providing researchers and developers with a flexible platform for experimentation and further development.

\subsection{Limitations of Existing Efforts on LLM Fine-Tuning on Mobile Phones}

Although some recent efforts such as PocketLLM and XPerT have attempted to enable LLM fine-tuning on mobile phones, they still face significant limitations:

\begin{itemize}

\item \textbf{PocketLLM} \cite{peng-etal-2024-pocketllm} implements on Termux, a terminal emulator and Linux-like user-space environment for Android. In this setup, the fine-tuning pipeline is executed through a separate Termux environment and Python runtime which introduces additional runtime overhead and reduces efficiency under the constrained resources of mobile phones. Moreover, this pipeline is difficult to directly integrate into real-world mobile applications, because it depends on an external terminal environment rather than a native computation backend that can be packaged with the app.
    
\item \textbf{XPerT} \cite{10.1145/3711875.3729132} implements through an ONNX-based training workflow. However, this workflow does not perform the full fine-tuning pipeline locally on mobile phones. In XPerT, the user's local data must be sent to a server-side Python runtime, where Python-based toolkits process the data and model to generate ONNX training artifacts. These artifacts are then sent back to the phone, where the Android application performs only the final parameter updates through the ONNX Runtime Java API. Therefore, XPerT still depends on the server during the fine-tuning pipeline, exposes local user data outside the device, and requires substantial offline re-engineering when switching across models, tasks, or applications.

\end{itemize}

Additionally, the implementation processes of both PocketLLM and XPerT have not been open-sourced, hindering other researchers from easily reusing or extending them. As a result, the absence of a unified open-source framework to support LLM fine-tuning on mobile phones creates a critical gap, which MobileFineTuner is designed to address.

\begin{figure}[t]
\centering
\includegraphics[width=\linewidth]{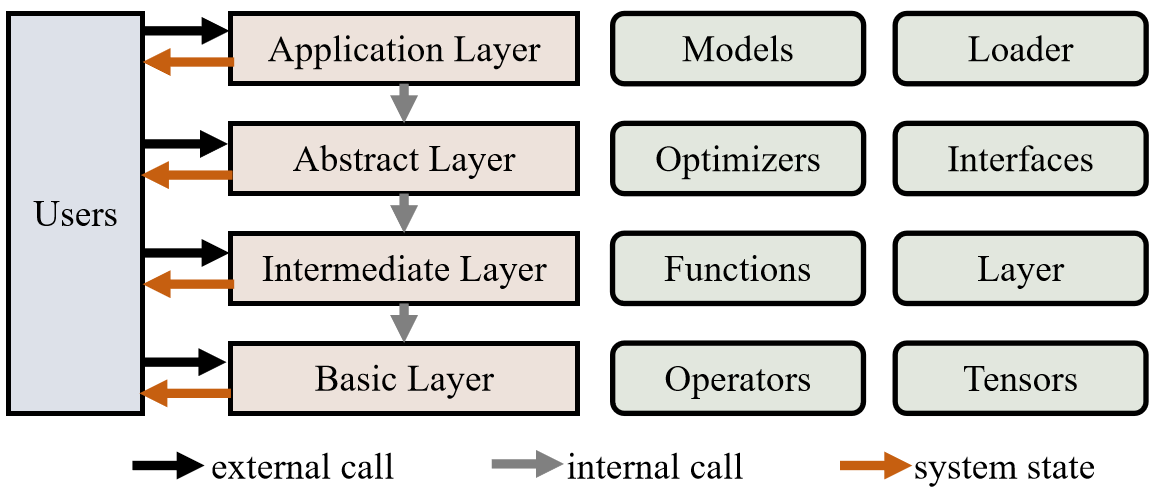} 
\caption{Four-layer hierarchical architecture of MobileFineTuner.}
\label{arch}
\vspace{-4mm}
\end{figure}

\section{MobileFineTuner}
In this section, we first introduce the four-layer architecture of MobileFineTuner, its functionalities, and corresponding API interfaces in Section 3.1. Then, we describe its usage process for full-parameter fine-tuning (Full-FT) and parameter-efficient fine-tuning (PEFT) in Section~3.2. Moreover, we compare our MobileFineTuner with mature frameworks for LLM implementation in Section~3.3.

\begin{table*}[t]
\centering
\caption{Comparison of MobileFineTuner with LLM Implementation Frameworks (Opt. refers to Optimization; "$\surd$" indicates support and "$\times$" indicates no support)}
\vspace{-2mm}
\label{tab:framework_comparison}
\resizebox{0.8\linewidth}{!}{
\begin{tabular}{lcccccccc}
\toprule
Framework & \makecell{LLM \\ Support} & \makecell{Inference \\ on Server} & \makecell{Inference \\ on Mobile} & \makecell{Fine-Tuning \\ on Server} & \makecell{Fine-Tuning \\ on Mobile} & \makecell{PEFT \\ Support} & \makecell{Built-in \\ Memory Opt.} & \makecell{Built-in \\ Energy Opt.} \\
\midrule
PyTorch & $\surd$ & $\surd$ & $\times$ & $\surd$ & $\textcolor{red}{\times}$ & $\times$ & $\times$ & $\times$ \\
Hugging Face Transformers & $\surd$ & $\surd$ & $\times$ & $\surd$ & $\textcolor{red}{\times}$ & $\surd$ & $\times$ & $\times$ \\
Hugging Face PEFT & $\surd$ & $\surd$ & $\times$ & $\surd$ & $\textcolor{red}{\times}$ & $\surd$ & $\times$ & $\times$ \\
Flower & $\surd$ & $\surd$ & $\times$ & $\surd$ & $\textcolor{red}{\times}$ & $\times$ & $\times$ & $\times$ \\
Llama.cpp & $\surd$ & $\surd$ & $\surd$ & $\times$ & $\textcolor{red}{\times}$ & $\times$ & $\surd$ & $\times$ \\
\textcolor{blue}{\textbf{MobileFineTuner (Ours)}} & \textcolor{blue}{\textbf{$\surd$}} & \textcolor{blue}{\textbf{$\surd$}} & \textcolor{blue}{\textbf{$\surd$}} & \textcolor{blue}{\textbf{$\surd$}} & \textcolor{red}{\textbf{$\surd$}} & \textcolor{blue}{\textbf{$\surd$}} & \textcolor{blue}{\textbf{$\surd$}} & \textcolor{blue}{\textbf{$\surd$}} \\
\bottomrule
\end{tabular}
}
\vspace{-2mm}
\end{table*}

\subsection{Framework Details and APIs}
To achieve high efficiency and fine-grained control suitable for mobile phones, MobileFineTuner is implemented entirely in C++. The framework employs a four-layer hierarchical architecture, which separates low-level computational primitives from high-level application logic. This design allows higher-level modules to automatically invoke lower-layer functionalities, while also enabling researchers and developers to directly call specific APIs for algorithm prototyping, experimental exploration, or system-level customization. The functionality and interfaces of each layer are as follows:

\begin{itemize}
    \item \textbf{Basic Layer}: 
    The Basic Layer provides the computational foundation of MobileFineTuner. It implements the core tensor abstraction, operator execution, automatic differentiation, data-type management, and device-level runtime support. This layer hides low-level memory and execution details from upper modules, allowing the framework to support training and inference through a unified computation interface.

    \item \textbf{Intermediate Layer}: 
    The Intermediate Layer builds reusable neural-network components on top of the Basic Layer. It provides common Transformer building blocks, including embedding layers, attention modules, feed-forward networks, and PEFT components such as LoRA adapters. These components expose modular interfaces, enabling developers to reuse or replace individual layers when implementing new model architectures or fine-tuning methods.

    \item \textbf{Abstract Layer}: 
    The Abstract Layer defines high-level modeling and optimization abstractions. It standardizes modules, layers, trainable parameters, optimizers, and update rules, thereby bridging low-level tensor computation and complete model training. With this layer, MobileFineTuner provides a PyTorch-like programming style while remaining independent of Python-based runtime environments.

    \item \textbf{Application Layer}: 
    The Application Layer constructs complete LLM fine-tuning and inference pipelines. It integrates model definition, pretrained parameter loading, tokenizer/model compatibility support, training-loop execution, LoRA fine-tuning, and model or adapter export. This layer is exposed to users and mobile applications as the main interface, allowing researchers and developers to perform end-to-end on-device fine-tuning without manually managing low-level operators or memory states.
\end{itemize}

\begin{lstlisting}[language=C++, caption={Simplified Full-FT Usage Code}, label={lst:full-ft}, float=t]
void train(DataLoader& data_loader) {
    for (int epoch = 0; epoch < config_.max_epochs; ++epoch) {
        float total_loss = 0.0f;
        for (int step = 0; step < config_.steps_per_epoch; ++step) {
            // Load a batch of data
            auto [input_ids, targets] = data_loader.get_batch(config_.batch_size);
            // Forward pass
            auto logits = model_->forward(input_ids);
            auto loss = model_->compute_loss(logits, targets);
            // Backward pass
            optimizer_->zero_grad();    // Reset gradients
            loss->backward();           // Automatic differentiation
            optimizer_->step();         // Update model parameters
            total_loss += loss->item<float>();
        }
    }
}
\end{lstlisting}

\subsection{Usage Process for Full-FT and PEFT}

In terms of fine-tuning capabilities, MobileFineTuner supports both Full-FT and PEFT. Full-FT enables research on novel fine-tuning paradigms, such as secondary development for implementing federated split fine-tuning. For PEFT, we support LoRA fine-tuning, which is currently the most widely used parameter-efficient method for LLM fine-tuning. Users can invoke these functionalities via high-level APIs, in a manner similar to PyTorch. The usage processes for Full-FT and PEFT (LoRA) are illustrated below.

\begin{itemize}
    \item Full-FT process: During the Full-FT process, users can invoke MobileFineTuner through a concise high-level API, as shown in Listing~\ref{lst:full-ft}. The process begins by defining a DataLoader that provides training batches, initializing the model and optimizer objects, and then executing the \textcolor{gray}{\textit{train()}} function. Within each training iteration, the framework sequentially performs the forward pass \textcolor{gray}{\textit{model\_$\rightarrow$forward()}}, loss computation \textcolor{gray}{\textit{model\_$\rightarrow$compute\_loss()}}, backward propagation \textcolor{gray}{\textit{loss$\rightarrow$ backward()}}, and \textcolor{gray}{\textit{optimizer\_$\rightarrow$step()}} to update the parameters. This design enables researchers to conduct on-device LLM fine-tuning using a PyTorch-like interface.

    \item PEFT (LoRA) process: MobileFineTuner defines a modular LoRA-based architecture that replaces standard layers within the self-attention module and constructs corresponding components, including \textcolor{gray}{\textit{LoRALinear}}, \textcolor{gray}{\textit{LoRAAttention}}, and \textcolor{gray}{\textit{LoRATransformerBlock}}. Users can conveniently perform LoRA fine-tuning by specifying a configuration object \textcolor{gray}{\textit{LoRAFinetuneConfig}}, loading pretrained LLM parameters, and invoking the unified \textcolor{gray}{\textit{LoRAFinetune}} interface to enable PEFT.
    
\end{itemize}

MobileFineTuner supports both input and output in downloadable model formats from Hugging Face, such as ".bin" and ".safetensor". This allows researchers and developers to directly download pretrained LLMs without the need for format conversion. Furthermore, the fine-tuned model is also output in ".bin" or ".safetensor" format, enabling direct integration with libraries such as PyTorch and Hugging Face Transformers. This ensures compatibility with existing frameworks and facilitates seamless integration for joint use.

\subsection{Comparison with Mature Frameworks for LLM Implementation}

In the context of LLM implementation, widely recognized frameworks include PyTorch, Hugging Face Transformers, Hugging Face PEFT, Flower, and Llama.cpp. These frameworks have gained popularity due to their efficiency, scalability, multi-model support, and ease of use for development and research. 
Their key characteristics and technical details are summarized as follows:

\begin{itemize}
     
\item \textbf{PyTorch} \cite{NEURIPS2019_bdbca288}: A Python-based deep learning framework that provides flexible tensor computation and automatic differentiation. PyTorch supports general model training and inference but does not natively provide mobile phone deployment or built-in system-level optimizations.

\item \textbf{Hugging Face Transformers} \cite{DBLP:journals/corr/abs-1910-03771}: Built on top of PyTorch, this Python library specializes in transformer-based models, offering pre-trained LLMs and streamlined pipelines for both inference and fine-tuning. It simplifies model handling and supports multiple LLM architectures, but it is limited to environments that support Python, such as Linux or Windows servers.

\item \textbf{Hugging Face PEFT} \cite{peft}: A Python-based library focused on PEFT for LLMs, such as LoRA and adapters. While it reduces the computational cost of fine-tuning and enables flexible algorithmic experimentation, it is designed for server-side execution and does not support mobile devices directly.

\item \textbf{Flower}\footnote{Flower does not currently support LLM fine-tuning on mobile phones. It allows Python-based LLM fine-tuning on servers and supports federated learning of traditional models on mobile devices. The mobile SDK is still under development and the updated version has not yet been released.} \cite{beutel2020flower}: A Python-based federated learning framework that orchestrates federated model training across multiple clients. Flower enables federated LLM training on server and edge devices that support Python, but it does not natively provide on-device LLM fine-tuning for mobile phones or built-in optimizations.

\item \textbf{Llama.cpp} \cite{ggmlorg_llama_cpp}: A C++ framework designed for efficient inference of LLMs on edge devices such as mobile phones. It integrates quantization and other device-side optimizations, achieving high efficiency without relying on Python. However, Llama.cpp focuses exclusively on inference and does not support LLM fine-tuning.

\end{itemize}

The features of these mature frameworks are summarized in Tab.~\ref{tab:framework_comparison}. It is evident that none of them support LLM fine-tuning directly on mobile phones. In contrast, MobileFineTuner not only enables fine-tuning on mobile devices but also integrates the functionalities provided by existing frameworks, including support for multiple LLM architectures, inference on both servers and mobile phones, server-side fine-tuning, PEFT support, and built-in system optimizations such as memory optimization, and energy optimization. By combining these features into a streamlined package, MobileFineTuner allows researchers and developers to implement LLMs with minimal overhead and reduced development effort, facilitating both experimentation and practical deployment.

\section{Resource-aware training runtime}

MobileFineTuner identifies two central challenges for LLM fine-tuning on mobile phones: \textbf{memory limitations (RAM)} and \textbf{energy constraints}. On the memory side, a “rule-of-thumb” estimate suggests that 16 GB of RAM is required per 1 billion parameters for FP16 training \cite{McKeag2025}, with the requirement doubling for FP32 training . In contrast, typical mobile phones in 2025 are equipped with only 4-16 GB of RAM \cite{AndroidAuthority2025}, making memory capacity a major bottleneck for LLM fine-tuning on mobile phones. On the energy side, continuous forward and backward passes over large models impose sustained computational loads, resulting in significant power consumption and accelerated battery drain. Since mobile devices typically rely on relatively small battery packs (e.g., ~3000–5000 mAh) and must simultaneously support regular user activities, performing LLM fine-tuning on mobile phones without energy optimization would severely degrade the overall user experience.

In this section, we propose a resource-aware training runtime including several built-in optimizations specifically designed for fine-tuning LLMs on mobile phones. Section~4.1 presents our memory optimization techniques, covering parameters sharding, gradients accumulation, activation checkpoint and memory-efficient attention, while Section~4.2 introduces our energy optimization strategy.

\subsection{Memory Optimization}

\subsubsection{ZeRO-inspired parameters sharding}

Considering the limited RAM on mobile phones and the relatively large disk storage available (often in the hundreds of GBs), it is crucial to offload some of the fine-tuning process to disk to alleviate the pressure on RAM. Therefore, MobileFineTuner implements a parameter sharding mechanism to optimize RAM usage during the fine-tuning process. 

\begin{figure}[t]
\centering
\includegraphics[width=0.8\linewidth]{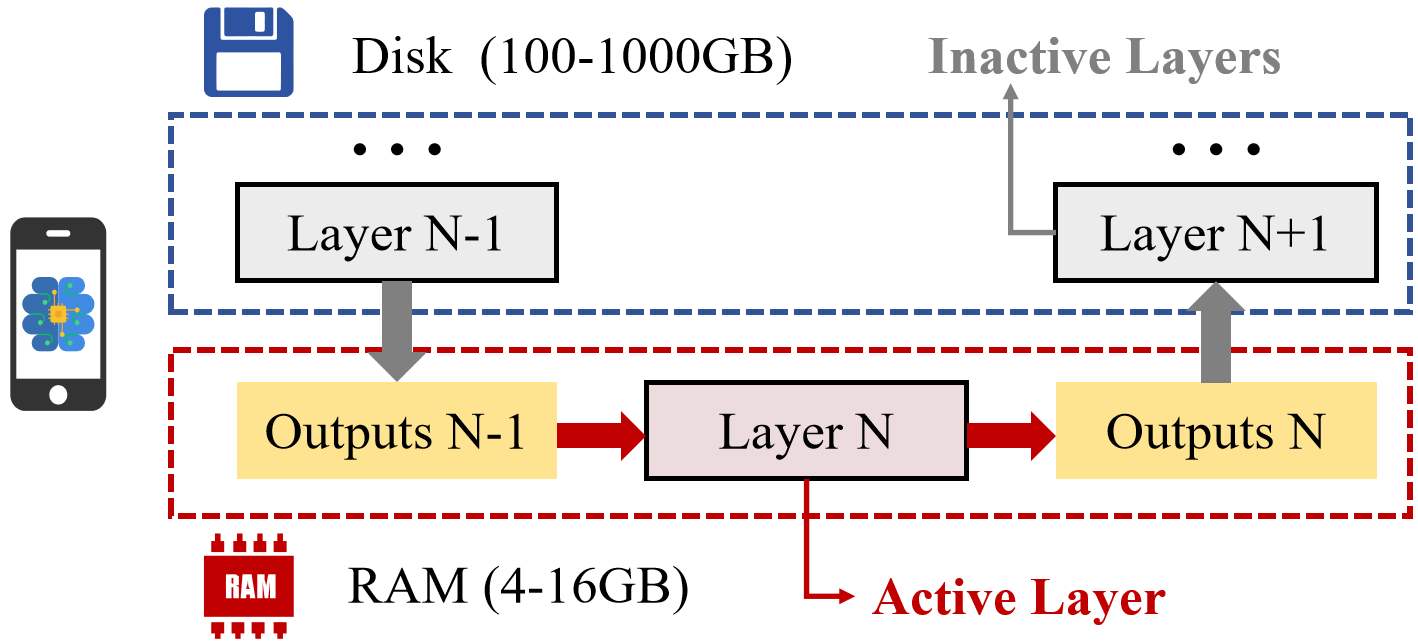}
\caption{ZeRO-inspired parameters sharding.}
\label{memory_parameters}
\end{figure}

The core design is inspired by the ZeRO-3 (Zero Redundancy Optimizer) principle \cite{10.5555/3433701.3433727}, but is re-engineered for single-device execution on mobile phones. Instead of maintaining a complete copy of all parameters in RAM, MobileFineTuner partitions the model parameters into multiple contiguous segments and loads only the active segment required for the current forward or backward pass, as shown in Fig.~\ref{memory_parameters}. Once a segment becomes inactive, it is promptly offloaded to disk storage to release RAM resources. 

Furthermore, we have developed a real-time memory management module to dynamically clear memory redundancy caused by reading or offloading parameter shards\footnote{Python has an automatic memory management system, whereas MobileFineTuner based on C++ requires real-time monitoring of memory usage and release to prevent memory leaks.}. During computation, MobileFineTuner also maintains an efficient mapping table that tracks the physical location and state of each parameter shard, ensuring seamless parameter swapping and uninterrupted model training without redundant recomputation.

\subsubsection{Gradients accumulation.}  

During fine-tuning, both forward activations and backward gradients consume a significant amount of memory. While smaller batch sizes reduce memory usage, they also result in gradient updates derived from fewer samples, leading to higher noise and greater fluctuations during training. Conversely, larger batch sizes lead to more stable model updates but increase memory usage~\cite{NEURIPS2024_15ba84c1}. 

To address this, MobileFineTuner incorporates gradient accumulation, which effectively balances memory usage and training stability. The approach works by breaking a single large-batch update into multiple micro-batches for forward and backward propagations. Gradients are accumulated, and a single optimizer update is performed afterward, achieving training stability similar to that of a large batch with the memory requirements of a micro-batch.

\subsubsection{Activation Checkpoint.}

\begin{figure}[t]
\centering
\includegraphics[width=0.9
\linewidth]{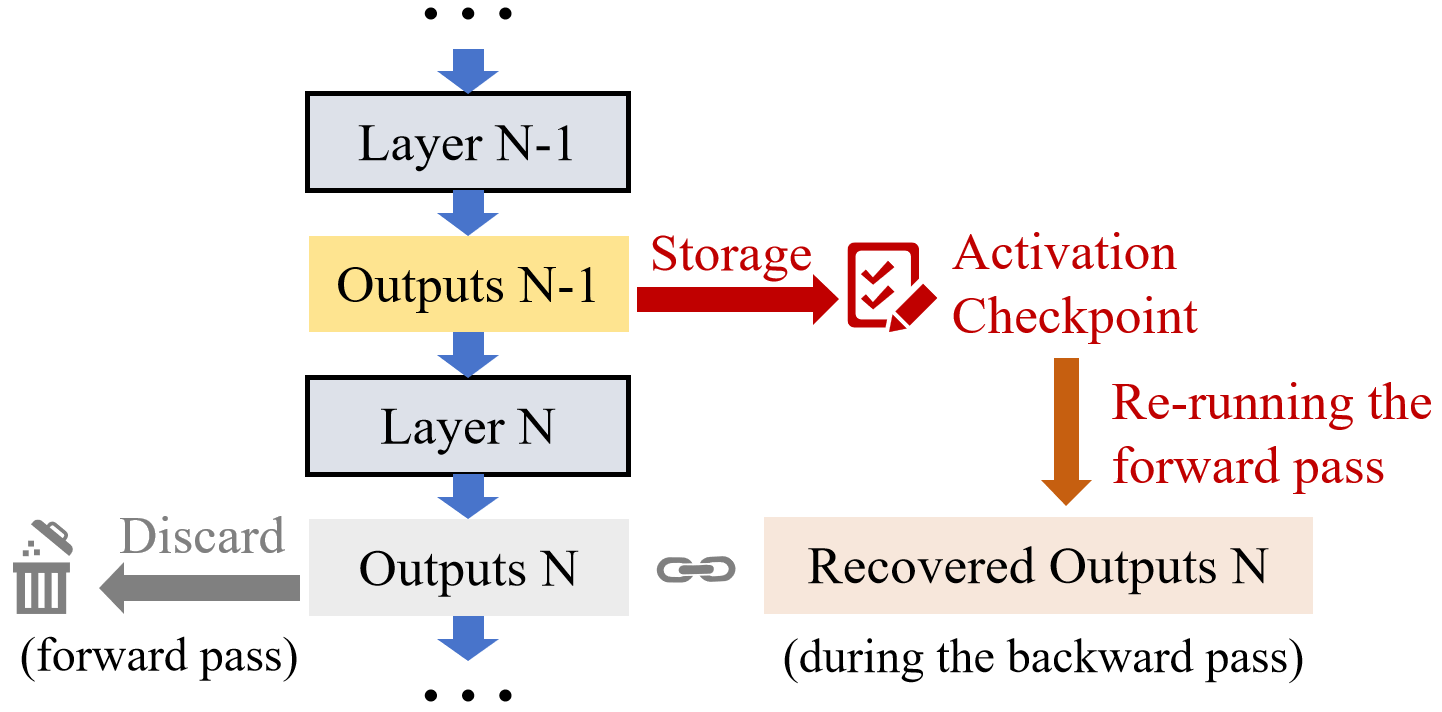}
\caption{Activation Checkpointing Mechanism.}
\label{memory_checkpoints}
\vspace{-2mm}
\end{figure}

Activation checkpointing is widely adopted in large Transformer model training to reduce activation memory by storing only selected activations and recomputing the discarded intermediate states during backpropagation~\cite{korthikanti2023reducing}. MobileFineTuner first supports activation checkpointing mechanism on mobile fine-tuning settings. As shown in Fig.\ref{memory_checkpoints}, the core principle of activation checkpointing is that not all intermediate activations (outputs of each layer during the forward pass) need to be stored in memory for backpropagation during the backward pass. Instead, checkpoints are created at strategic points, saving only a subset of activations during the forward pass while discarding others that are not immediately needed. When the backward pass is executed, the activations that were discarded can be recomputed by re-running the forward pass from the checkpointed layers. This selective retention of activations significantly reduces memory usage, as only a fraction of the activations need to be stored in memory at any given time.

\subsubsection{Memory-efficient Attention}

MobileFineTuner also supports an optional memory-efficient attention optimization to reduce the memory pressure caused by Transformer attention. In standard self-attention, the attention score matrix and attention probability matrix are explicitly materialized with the shape of $[B, H, S, S]$, where $B$, $H$, and $S$ denote the batch size, number of attention heads, and sequence length, respectively. These intermediate tensors grow quadratically with the sequence length and can become a major source of peak RSS during fine-tuning. Prior studies have shown that self-attention does not necessarily require storing the full quadratic attention matrix, and that exact attention can be implemented in a memory-efficient manner by avoiding unnecessary materialization of attention intermediates~\cite{rabe2021selfattention,dao2022flashattention}.

MobileFineTuner provides a configurable memory-efficient attention path, which avoids constructing the full attention matrix and instead computes attention in a row-streaming manner. For each query position, the operator computes only the corresponding row of attention scores against all keys, applies the causal or padding mask, performs a numerically stable softmax through row-wise max normalization, and directly accumulates the weighted sum over value vectors. Once the output of the current query position is generated, the temporary row buffer is released and reused for the next query position.

During backpropagation, this optimized path does not store the full attention probability matrix from the forward pass. Instead, it recomputes the local row-wise softmax statistics from $Q$, $K$, and $V$, and then accumulates gradients for the query, key, and value tensors. This design trades additional computation for lower memory usage, reducing the attention intermediate memory from $O(BHS^2)$ to row-level temporary storage. Different from GPU-oriented fused attention kernels such as FlashAttention~\cite{dao2022flashattention}, our implementation is designed as a pure C++ row-streaming attention operator for mobile-side training.

\subsection{Energy Optimization}
\label{sec:energy}

\begin{figure}[t]
\centering
\includegraphics[width=0.95
\linewidth]{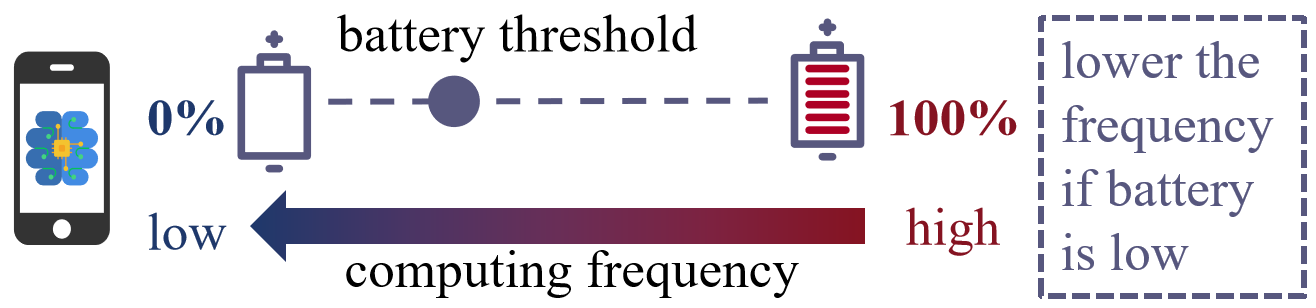}
\caption{Energy-aware computation scheduling.}
\label{energy_optimizer}
\vspace{-2mm}
\end{figure}

MobileFineTuner proposes an energy-aware dynamic computation scheduling method to optimize energy consumption during fine-tuning of LLMs on mobile phones. Given the typical usage patterns of smartphones, this method aims to reduce the computation frequency of LLM fine-tuning under conditions of low battery, thereby alleviating the computational load on the phone and ensuring a better user experience. 

We have integrated a \textit{PowerMonitor} that tracks the mobile phone's battery percentage during fine-tuning. The \textit{PowerMonitor} performs checks every $K$ fine-tuning steps. As shown in Fig. \ref{energy_optimizer}, when the battery percentage drops below a user-defined threshold $\mu$, the computation frequency is dynamically adjusted, reducing it by a percentage $\rho$. Both $K$, $\mu$ and $\rho$ are configurable hyperparameters, allowing the system to be adapted to various use cases. The frequency adjustment is implemented by introducing a sleep delay at each fine-tuning step, ensuring that the computational process is dynamically managed to balance energy efficiency with system performance.

\section{Case Study: Private Campus Health Agent}
\label{sec:health-agent}

\begin{figure}[t]
\centering
\includegraphics[width=0.99
\linewidth]{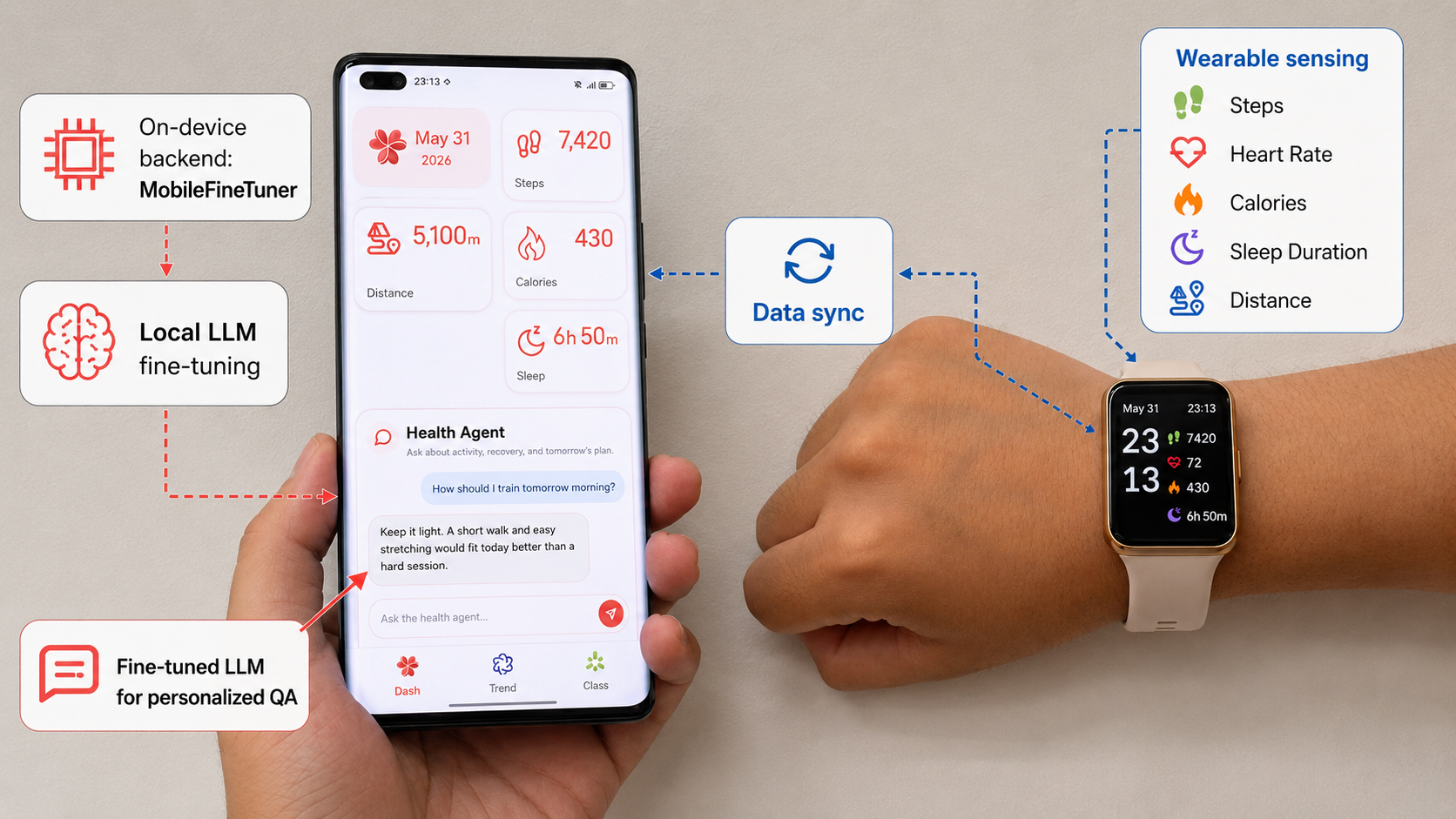}
\caption{Real-world Private Campus Health Agent with MobileFineTuner as Backend.}
\label{case}
\vspace{-4mm}
\end{figure}

To demonstrate how MobileFineTuner supports real mobile embedded AI applications, we build a private campus health agent on top of MobileFineTuner, as shown in Fig.~\ref{case}. The real-world test is based on a campus wearable-sensing scenario, where students use smartwatches to record daily health and activity information, and a mobile agent provides personalized feedback based on each student's own historical records. In this section, we first introduce our three-month real-world application test in campus with 28 student participants. We then describe how the local health records are converted into personalized QA pairs for local LLM fine-tuning. Finally, we describe how the health-agent application uses MobileFineTuner as its on-device fine-tuning backend.

\subsection{Campus Health Sensing Scenario}

The case study is conducted in a university campus setting. We invite 28 student particiants to use our campus health agent app for 3 months. Each participant uses a Huawei smartwatch to collect daily health and activity records, including steps, calories, walking distance, heart rate, and sleep duration. These records are synchronized to the student's mobile phone, which serves as both the local data host and the computation platform for the health agent.

Our application is a private health assistant that serves an individual student. Given the student's historical records, the private agent is expected to answer questions about daily status, recent trends, abnormal changes, and personalized suggestions using the local-based LLM. For example, it is able to explain whether the student's recent activity level is lower than usual, whether the student's sleep pattern deviates from their own historical baseline, or what behavioral adjustment may be useful based on recent health trends. Different from existing cloud-centric health agents, our application performs both inference and fine-tuning on the mobile phone. Raw wearable records are not uploaded to a server. Instead, the phone locally processes these records, constructs fine-tuning samples, invokes MobileFineTuner to adapt the LLM on the device, and stores the personalized adapter for subsequent health-agent inference.

\subsection{Local QA Construction}
\label{sec:local-qa-construction}

The records collected by smartwatches are numerical and time-series signals, while the target health agent interacts with users through question answering. Therefore, to fine-tune the agent, we need to convert each student's local health records into instruction-response QA pairs. To preserve the health records locally, this conversion is performed through a template-based local construction pipeline. We first use GPT-5.5 to generate generic QA templates for health-agent interactions. These templates only define the linguistic structure of questions and answers, with abstract slots for user-specific values, trends, and baselines. They do not contain any personal records. The mobile application then fills these slots locally using health statistics computed from the raw records. 

The generated QA pairs cover five categories: \textbf{Activity Summary}, \textbf{Goal Adjustment}, \textbf{Habit Coaching}, \textbf{Metric Insight}, and \textbf{Plan Recommendation}. These categories cover both descriptive queries over local records and reasoning-oriented queries that require comparison with the student's historical baseline. For each student, the mobile application locally generates 8,000 personalized QA pairs from the student's wearable records and derived health statistics, and uses them for local LLM fine-tuning on the student's phone. We anonymize the derived QA pairs and release them as the \textbf{Campus Health Question Answering Dataset (CHQA)}, which contains 28 anonymized IDs with 8,000 QA pairs per ID and 224,000 QA pairs in total. QA samples are shown in Appendix~\ref{appE}.

\subsection{MobileFineTuner as the Application Backend}
\label{sec:health-agent-backend}

We implement the private campus health agent by directly using MobileFineTuner as the on-device fine-tuning backend. The mobile application manages smartwatch synchronization, local record storage, health-statistics computation, QA construction, and user interaction, while MobileFineTuner performs local model adaptation, including model loading, LoRA fine-tuning, resource-aware execution, and local adapter storage. After the personalized QA pairs are generated on the phone, the application passes the local dataset and fine-tuning configuration to MobileFineTuner. MobileFineTuner then fine-tunes the selected LLM locally and stores the adapted model or LoRA adapter on the phone for health-agent inference. We use Qwen2.5-0.5B as the base model. We report the detailed parameters and experimental results in Sec.~\ref{case_results_sec}.

\begin{table}[t]
\centering
\caption{Device Specifications}
\resizebox{\linewidth}{!}{
\begin{tabular}{cccc}
\toprule
\textbf{Device Name} & \textbf{OS} & \textbf{CPU / SoC} & \textbf{RAM Size} \\
\midrule

Huawei P50 Pro 
& Android 11.0 
& Kirin 9000
& 8 GB \\

Huawei nova 9 Pro 
& HarmonyOS 2.0 
& Snapdragon 778G 4G 
& 8 GB  \\

iQOO 15 
& Android 16 
& Snapdragon 8 Elite Gen 5 
& 16 GB+16 GB\\

MacBook Air 2023 
& macOS Sequoia 15.6.1 
& Apple M2 
& 16 GB+32 GB \\
\bottomrule
\end{tabular}
}
\label{devices}
\end{table}

\section{Implementation}
In this section, we provide the implementation details of MobileFineTuner for our common evaluation, including device setup, models support, and data tasks.

\subsection{Device Setup}
\subsubsection{Framework compilation}
MobileFineTuner supports all mobile devices running Android 9+ (API level 28+). The framework offers two compilation methods\footnote{The first one is recommended for faster performance.}: the first involves server-side NDK compilation followed by uploading the compiled project to the mobile device via ADB (Android Debug Bridge\footnote{ADB is a versatile tool that enables direct communication with the device's backend, allowing operations via shell-like commands, thus simplifying the process for users.}); the second method places the source code directly on the mobile device and compiles it on mobile phone via ADB. Once the compilation is complete, users can upload the model and dataset, configure the necessary runtime parameters, and execute the compiled binary to initiate the fine-tuning process. This entire workflow eliminates the need for additional development efforts from the user.

\subsubsection{Metrics observer} MobileFineTuner includes a built-in metrics observer that logs key statistics at each step, including step number, training loss, testing loss, testing perplexity (PPL) / accuracy, RSS (Resident Set Size), and power consumption. Below is an explanation of the observers for RSS and power consumption:
\begin{itemize}
 \item RSS is a crucial memory metric that indicates the actual amount of RAM used by a process. It is calculated by summing the private memory used by the process and the memory occupied by shared libraries. RSS represents the total physical memory consumed by a process, including its code, data, heap, and stack. MobileFineTuner retrieves RSS information by calling the \textcolor{gray}{\textit{dumpsys procstats}} command, which provides memory usage statistics for running processes.

\item Power consumption is measured using the built-in \textit{PowerMonitor}, which interacts with Android's \textcolor{gray}{\textit{BatteryStatsService}}. Our  \textit{PowerMonitor} acquires power consumption data by reading the Android kernel's power configuration files, \textcolor{gray}{\textit{power\_profile.xml}}.
 \end{itemize}

\subsubsection{Device specifications for our common evaluation.}
For our common evaluation, we select a diverse set of mainstream commercial smartphones, together with one laptop platform, as listed in Tab.~\ref{devices}. The smartphone devices cover different vendors, operating systems, processor platforms, and memory capacities, allowing us to evaluate the practicality and robustness of MobileFineTuner under heterogeneous mobile hardware conditions. The MacBook Air is additionally included to examine the cross-platform adaptability of MobileFineTuner beyond Android smartphones.

\subsection{Models}
Unlike inference, fine-tuning requires significantly more memory and computational resources—often several times more than inference. Consequently, large models with extensive parameters are impractical on mobile devices, as they would exceed the capabilities of most mobile phones.

Thus, we have focused on models specifically designed for deployment on resource-constrained devices. The models selected for our evaluation include: GPT2-small-124M, GPT2-medium-355M \cite{radford2019language}, Qwen2.5-0.5B \cite{qwen2.5}, Gemma3-270M, and Gemma3-1B \cite{gemma_2025}. The fine-tuning processes for these models have already been integrated as examples within MobileFineTuner, enabling users to run them directly without additional development. For other models, we plan to gradually update the examples. For user-defined models, users can also make adjustments as needed, requiring only minor modifications to the high-level functions (such as adjusting the tokenizer to suit their specific model).

\begin{table*}[t]
\centering
\caption{PEFT (LoRA) metrics with Seq 128. (Initial Loss/Acc/PPL $\rightarrow$ Best Final Loss/Acc/PPL)}
\renewcommand{\arraystretch}{1.15}
\resizebox{\linewidth}{!}{%
\begin{tabular}{llccccccccc}
\toprule
\multirow{2}{*}{Task}
& \multirow{2}{*}{Model} 
& \multicolumn{2}{c}{Loss} 
& \multicolumn{2}{c}{Accuracy (\%)} 
& \multicolumn{2}{c}{PPL}
& \multicolumn{3}{c}{System Metrics} \\
\cmidrule(lr){3-4}\cmidrule(lr){5-6}\cmidrule(lr){7-8}\cmidrule(lr){9-11}
& 
& MobileFineTuner & PyTorch 
& MobileFineTuner & PyTorch 
& MobileFineTuner & PyTorch
& Time (h) & Energy (kJ) & Peak RSS (MB) \\
\midrule

\multirow{5}{*}{MMLU}
& GPT2-124M 
& $2.232 \rightarrow 1.531$ 
& $1.617 \rightarrow 1.532$ 
& $21.70 \rightarrow 26.07$ 
& $24.54 \rightarrow 26.54$ 
& $15.37 \rightarrow 4.37$ 
& $5.04 \rightarrow 4.63$ 
& 36.33 & 90.15 & 1246.47 \\

& GPT2-355M 
& $1.708 \rightarrow 1.690$ 
& $1.361 \rightarrow 1.257$ 
& $24.61 \rightarrow 26.44$ 
& $23.95 \rightarrow 26.31$ 
& $6.16 \rightarrow 4.25$ 
& $3.90 \rightarrow 3.51$ 
& 97.91 & 250.59 & 3310.55 \\

& Qwen2.5-0.5B 
& $0.870 \rightarrow 0.658$ 
& $0.901 \rightarrow 0.506$ 
& $44.25 \rightarrow 47.35$ 
& $45.93 \rightarrow 47.50$ 
& $2.39 \rightarrow 1.93$ 
& $2.46 \rightarrow 1.66$ 
& 25.94 & 75.88 & 2887.74 \\

& Gemma3-270M 
& $1.781 \rightarrow 1.651$ 
& $1.333 \rightarrow 1.303$ 
& $24.60 \rightarrow 26.84$ 
& $24.32 \rightarrow 26.89$ 
& $5.68 \rightarrow 4.32$ 
& $4.79 \rightarrow 3.68$ 
& 49.39 & 118.09 & 4041.93 \\

& Gemma3-1B 
& $9.697 \rightarrow 1.361$ 
& $1.459 \rightarrow 1.422$ 
& $24.60 \rightarrow 27.51$ 
& $24.69 \rightarrow 27.59$ 
& $8.23 \rightarrow 4.02$ 
& $4.30 \rightarrow 4.14$ 
& 142.76 & 437.98 & 8209.24 \\

\midrule

\multirow{5}{*}{PIQA}
& GPT2-124M 
& $3.203 \rightarrow 3.098$ 
& $3.162 \rightarrow 2.832$ 
& $62.62 \rightarrow 63.66$ 
& $63.00 \rightarrow 64.04$ 
& $28.26 \rightarrow 24.56$ 
& $26.70 \rightarrow 22.59$ 
& 22.64 & 55.79 & 1241.93 \\

& GPT2-355M 
& $2.933 \rightarrow 2.906$ 
& $3.132 \rightarrow 2.920$ 
& $67.57 \rightarrow 68.34$ 
& $68.12 \rightarrow 69.37$ 
& $20.90 \rightarrow 18.20$ 
& $19.67 \rightarrow 16.65$ 
& 61.24 & 158.00 & 3343.67 \\

& Qwen2.5-0.5B 
& $2.677 \rightarrow 2.355$ 
& $2.694 \rightarrow 2.325$ 
& $70.08 \rightarrow 70.84$ 
& $71.27 \rightarrow 71.44$ 
& $16.97 \rightarrow 16.01$ 
& $15.51 \rightarrow 14.03$ 
& 39.30 & 114.34 & 2887.35 \\

& Gemma3-270M 
& $3.727 \rightarrow 3.214$ 
& $2.791 \rightarrow 2.676$ 
& $60.77 \rightarrow 68.81$ 
& $66.21 \rightarrow 69.42$ 
& $41.55 \rightarrow 24.88$ 
& $16.29 \rightarrow 14.52$ 
& 61.88 & 144.14 & 4042.02 \\

& Gemma3-1B 
& $3.412 \rightarrow 3.260$ 
& $2.401 \rightarrow 2.364$ 
& $66.38 \rightarrow 76.99$ 
& $74.37 \rightarrow 77.53$ 
& $30.33 \rightarrow 26.05$ 
& $11.03 \rightarrow 10.63$ 
& 178.00 & 535.92 & 8726.88 \\

\midrule

\multirow{5}{*}{ARC-C}
& GPT2-124M 
& $3.654 \rightarrow 3.188$ 
& $3.527 \rightarrow 3.120$ 
& $21.07 \rightarrow 22.74$ 
& $21.07 \rightarrow 22.74$ 
& $51.20 \rightarrow 40.67$ 
& $50.25 \rightarrow 42.30$ 
& 4.53 & 13.25 & 1230.14 \\

& GPT2-355M 
& $3.391 \rightarrow 2.904$ 
& $3.056 \rightarrow 2.925$ 
& $26.09 \rightarrow 28.76$ 
& $26.09 \rightarrow 26.76$ 
& $36.80 \rightarrow 27.04$ 
& $30.29 \rightarrow 26.73$ 
& 12.24 & 44.63 & 3334.55 \\

& Qwen2.5-0.5B 
& $2.804 \rightarrow 2.311$ 
& $2.506 \rightarrow 2.273$ 
& $35.79 \rightarrow 38.79$ 
& $38.13 \rightarrow 38.80$ 
& $18.39 \rightarrow 15.45$ 
& $13.24 \rightarrow 11.87$ 
& 7.84 & 22.31 & 2887.33 \\

& Gemma3-270M 
& $4.697 \rightarrow 3.743$ 
& $3.224 \rightarrow 3.109$ 
& $25.08 \rightarrow 27.42$ 
& $25.75 \rightarrow 28.43$ 
& $109.62 \rightarrow 42.22$ 
& $25.14 \rightarrow 22.39$ 
& 12.33 & 28.83 & 3914.85 \\

& Gemma3-1B 
& $4.644 \rightarrow 3.232$ 
& $2.521 \rightarrow 2.191$ 
& $25.08 \rightarrow 41.44$ 
& $40.13 \rightarrow 42.81$ 
& $103.96 \rightarrow 25.33$ 
& $12.44 \rightarrow 8.94$ 
& 35.52 & 103.49 & 8941.61 \\

\midrule

\multirow{5}{*}{ARC-E}
& GPT2-124M 
& $3.075 \rightarrow 2.748$ 
& $2.797 \rightarrow 2.692$ 
& $42.28 \rightarrow 46.84$ 
& $43.33 \rightarrow 46.14$ 
& $42.46 \rightarrow 27.73$ 
& $31.73 \rightarrow 23.12$ 
& 9.05 & 22.45 & 1231.36 \\

& GPT2-355M 
& $2.651 \rightarrow 2.455$ 
& $2.585 \rightarrow 2.314$ 
& $45.26 \rightarrow 52.11$ 
& $51.23 \rightarrow 53.33$ 
& $29.79 \rightarrow 17.02$ 
& $18.01 \rightarrow 13.79$ 
& 24.48 & 62.37 & 3334.63 \\

& Qwen2.5-0.5B 
& $1.973 \rightarrow 1.917$ 
& $2.526 \rightarrow 1.651$ 
& $67.72 \rightarrow 71.05$ 
& $69.82 \rightarrow 71.40$ 
& $10.86 \rightarrow 7.62$ 
& $7.62 \rightarrow 6.76$ 
& 15.61 & 45.74 & 2887.61 \\

& Gemma3-270M 
& $3.907 \rightarrow 3.523$ 
& $2.547 \rightarrow 2.287$ 
& $42.63 \rightarrow 62.81$ 
& $52.63 \rightarrow 62.81$ 
& $49.75 \rightarrow 33.89$ 
& $12.77 \rightarrow 9.84$ 
& 24.65 & 58.49 & 3899.98 \\

& Gemma3-1B 
& $4.257 \rightarrow 2.241$ 
& $2.371 \rightarrow 1.573$ 
& $50.00 \rightarrow 75.02$ 
& $72.46 \rightarrow 77.89$ 
& $70.60 \rightarrow 10.16$ 
& $10.71 \rightarrow 4.82$ 
& 76.95 & 198.66 & 8962.99 \\

\bottomrule
\end{tabular}%
}
\label{tab:peft_metrics_all_seq128}
\vspace{-2mm}
\end{table*}

\begin{figure}[t]
\centering
\includegraphics[width=\linewidth]{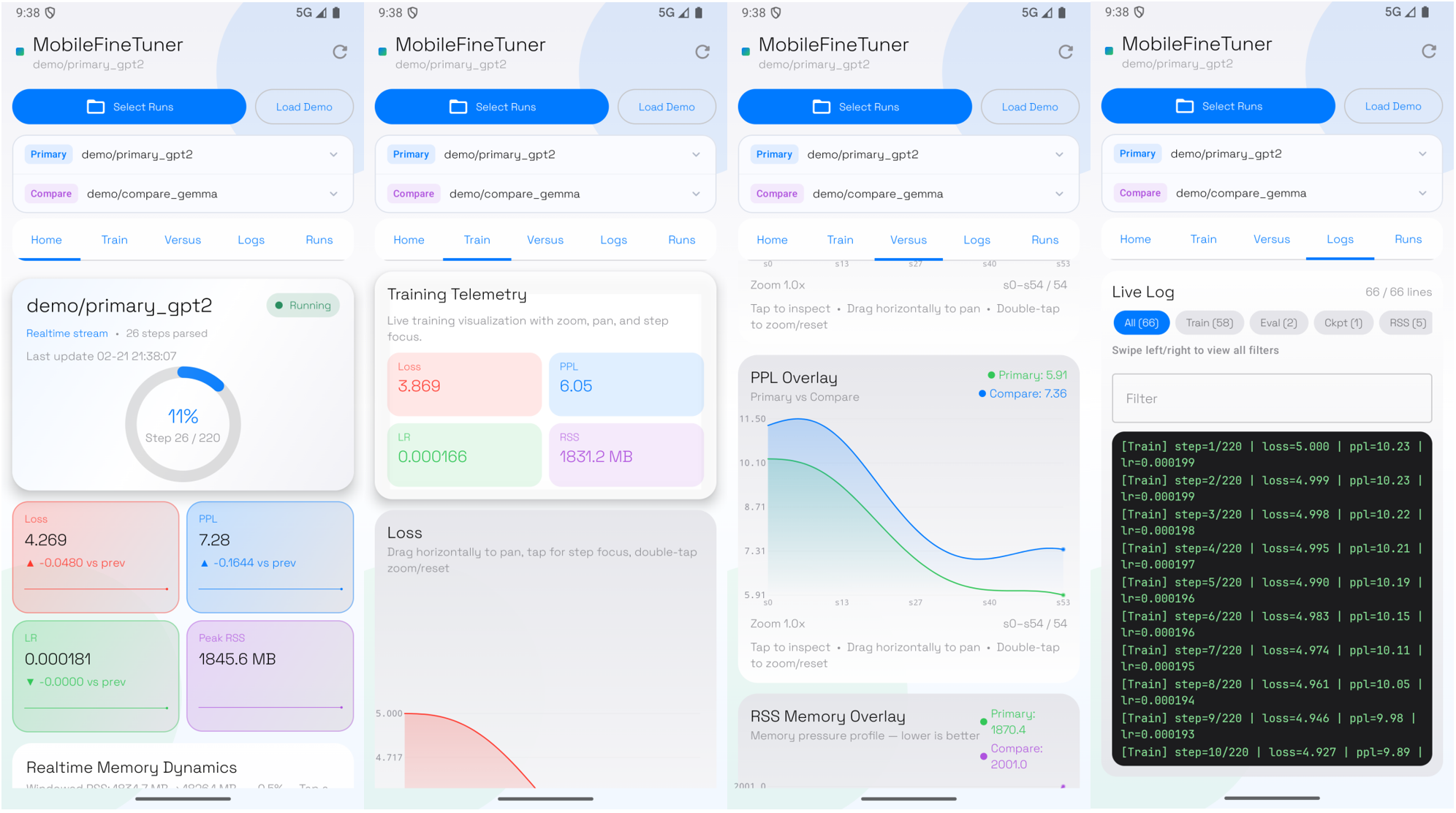}
\caption{MobileFineTuner training visualizer.}
\label{fig:visualizer}
\vspace{-4mm}
\end{figure}

\subsection{Datasets}

For our common evaluation, we validate the effectiveness of MobileFineTuner on six widely used language modeling and reasoning benchmarks: WikiText-2, MMLU, ARC-Challenge, ARC-Easy, HellaSwag, and PIQA. These datasets cover two representative categories of mobile-side LLM fine-tuning tasks: text generation and text-based reasoning/question answering.

\begin{itemize}
    \item \textbf{Text Generation Task.} 
    We use WikiText-2~\cite{merity2016pointer} for the text generation task. We fine-tune the models on the training corpus and report perplexity (PPL), which measures the model's ability to predict the next token in a sequence.

    \item \textbf{Text-based Reasoning and Question Answering Tasks.} 
    To evaluate the reasoning and question answering capability of MobileFineTuner, we use five multiple-choice benchmarks: MMLU~\cite{hendrycks2021measuring}, ARC-Challenge (ARC-C)~\cite{clark2018think}, ARC-Easy (ARC-E)~\cite{clark2018think}, HellaSwag~\cite{zellers2019hellaswag}, and PIQA~\cite{bisk2020piqa}. 
All evaluations are conducted in a zero-shot setting.
For accuracy evaluation, we adopt letter-token classification accuracy for multiple-choice tasks, and the final accuracy is computed as the fraction of examples for which the predicted letter matches the ground-truth answer. This follows the common likelihood-based multiple-choice evaluation protocol used for autoregressive language models~\cite{brown2020language,wang2024answerc}.
\end{itemize}

\subsection{Training Visualizer}
\label{sec:visualizer}

Although MobileFineTuner executes fine-tuning as a backend training process on mobile devices, directly monitoring such a process through command-line logs is inconvenient for users and makes it difficult to inspect the training status in real time. To improve usability and observability, we develop a lightweight training visualizer, as shown in Fig.~\ref{fig:visualizer}. The visualizer is decoupled from the training engine. It provides a user interface for tracking the runtime behavior of MobileFineTuner without interrupting the backend training process. It reads the metrics and logs generated during fine-tuning and presents them in an interactive dashboard, including the current training progress, loss, perplexity, learning rate, peak RSS, and live training logs. We present the screen record during the fine-tuning process in Appendix~\ref{appD}.

\begin{figure}[t]
    \centering
    \begin{minipage}{0.48\linewidth}
        \centering
        \includegraphics[width=\linewidth]{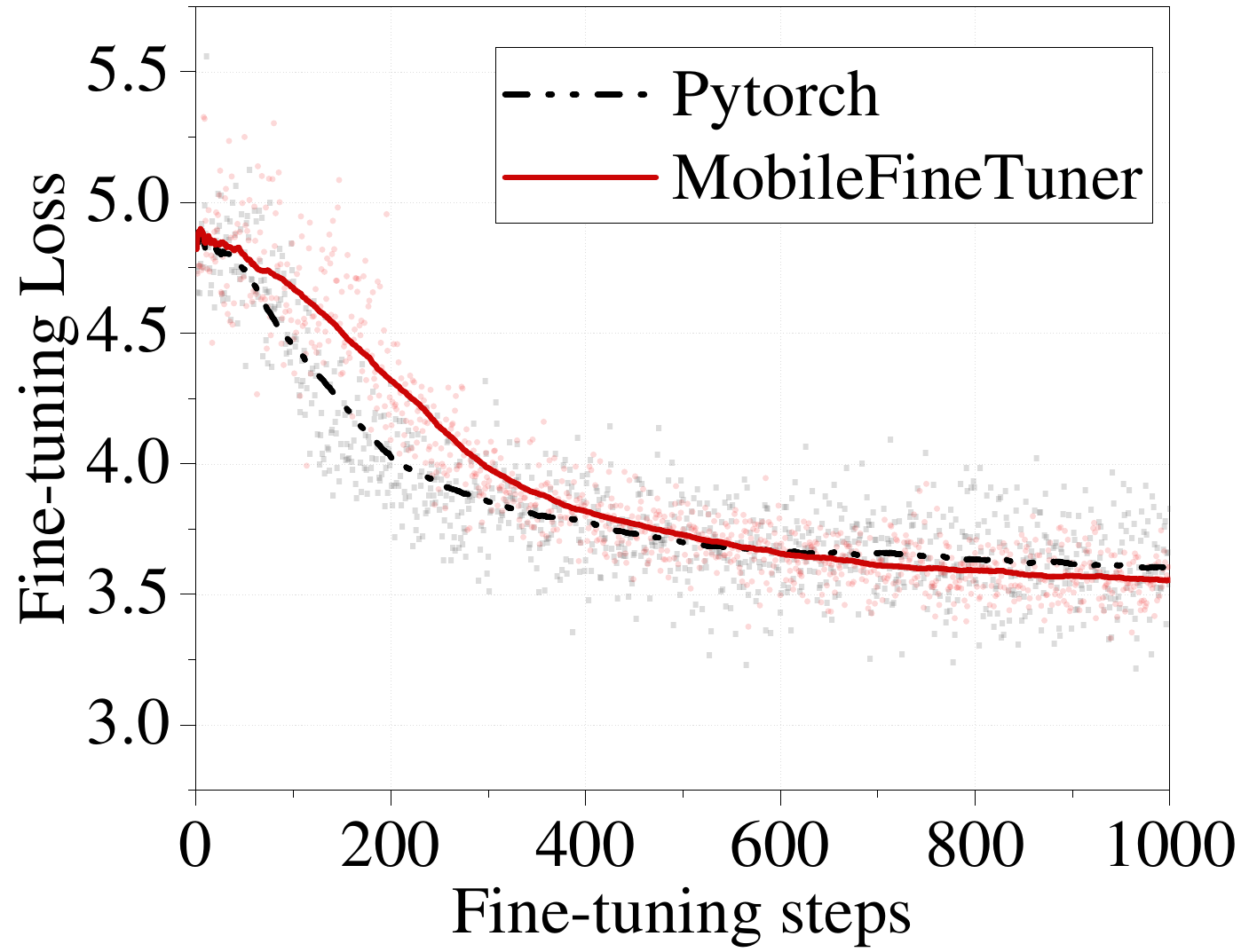}
        \small (a) Fine-tuning loss
    \end{minipage}
    \hfill
    \begin{minipage}{0.49\linewidth}
        \centering
        \includegraphics[width=\linewidth]{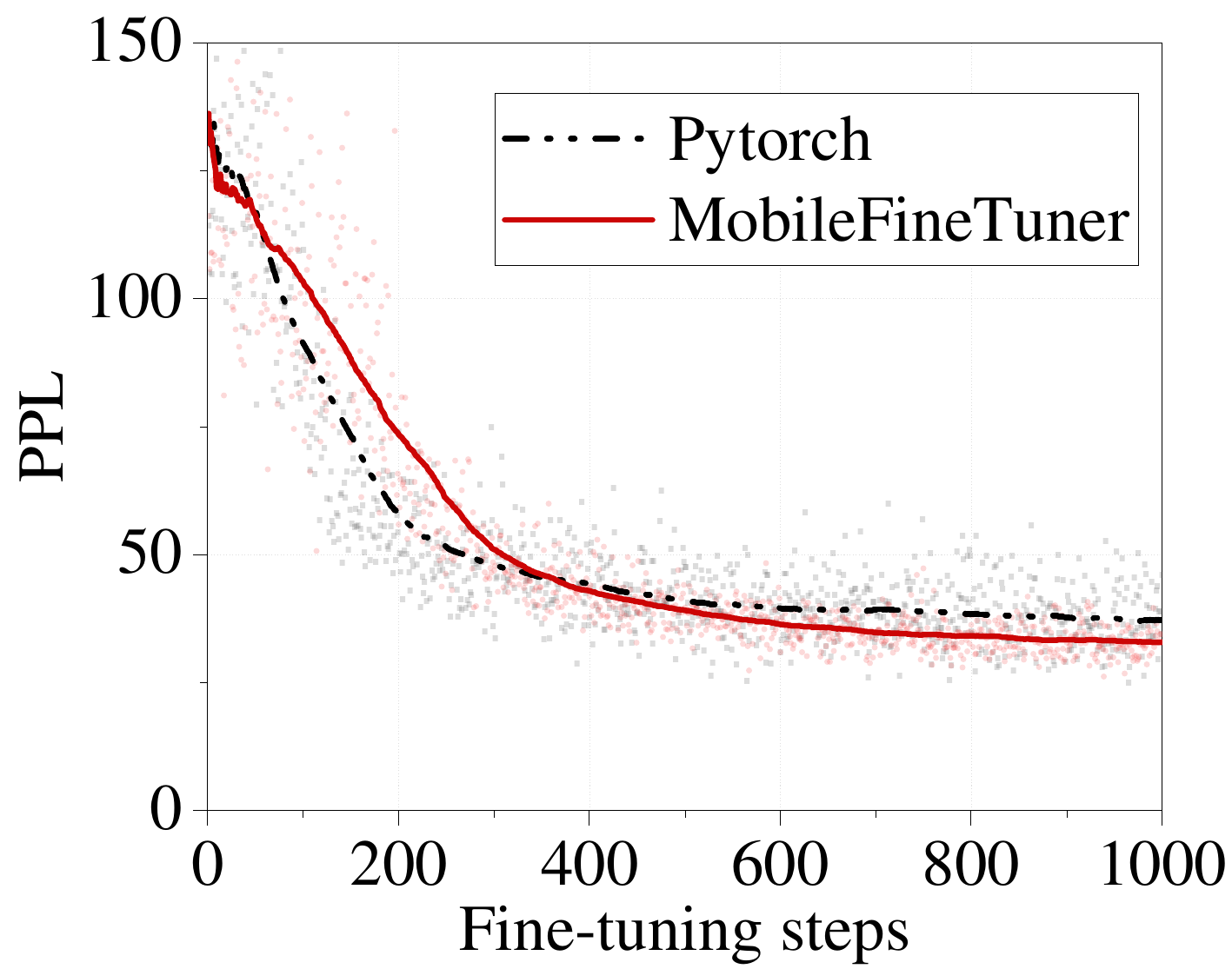}
        \small (b) Perplexity
    \end{minipage}
\vspace{-2mm}
   \caption{Loss of Full-FT on GPT2-124M@WikiText-2.}
\label{Full-FT}
\vspace{-2mm}
\end{figure}

\section{Framework Evaluation}
In this section, we present the common framework evaluation of MobileFineTuner. The evaluation consists of three parts. First, we validate the correctness of MobileFineTuner by comparing its Full-FT and PEFT (LoRA) training behavior with server-side Pytorch on mainstream LLMs. Second, we evaluate the effectiveness of the resource-aware training runtime. Third, we compare MobileFineTuner with a Termux-based fine-tuning pipeline to quantify the performance benefits of its mobile-native design.

\subsection{Correctness Evaluation}

\subsubsection{Full-FT}
\

\noindent \textbf{Settings.}
To evaluate the feasibility of MobileFineTuner’s Full-FT capability, we evaluate MobileFineTuner's Full-FT capability on GPT2-124M using the WikiText-2 dataset. All experiments are conducted on an iQOO 15 with sequence length 128, batch size 8, and learning rate $1\times10^{-5}$.

\noindent \textbf{Results.}
Fig.~\ref{Full-FT} compares the loss and PPL trajectories of MobileFineTuner and a PyTorch implementation. PyTorch runs in a server-side Python environment, while MobileFineTuner performs Full-FT directly on the phone. MobileFineTuner shows steadily decreasing loss and PPL curves that closely follow the PyTorch results, validating the correctness of its Full-FT implementation. It is important to note that this comparison with PyTorch is not intended to demonstrate performance superiority.  Instead, the alignment between the two results serves to validate the correctness and reliability of our Full-FT implementation.

\begin{table*}[t]
\centering
\caption{Runtime testing accuracy and PPL with Seq 128. (M) indicates MobileFineTuner. (P) indicates PyTorch.}
\renewcommand{\arraystretch}{1.15}
\resizebox{\linewidth}{!}{%
\begin{tabular}{llccccccccc}
\toprule
\multirow{2}{*}{Task}
& \multirow{2}{*}{Model} 
& \multicolumn{3}{c}{30\% Progress} 
& \multicolumn{3}{c}{60\% Progress} 
& \multicolumn{3}{c}{90\% Progress} \\
\cmidrule(lr){3-5}\cmidrule(lr){6-8}\cmidrule(lr){9-11}
& 
& Step (M/P) & Accuracy (M/P) & PPL (M/P) 
& Step (M/P) & Accuracy (M/P) & PPL (M/P) 
& Step (M/P) & Accuracy (M/P) & PPL (M/P) \\
\midrule

\multirow{5}{*}{MMLU}
& GPT2-124M 
& 330/360 & 24.74/24.62 & 4.29/4.28 
& 660/660 & 24.94/26.36 & 4.20/4.23 
& 990/990 & 25.28/25.57 & 4.16/4.19 \\

& GPT2-355M 
& 360/360 & 25.07/25.63 & 4.28/4.39 
& 690/660 & 25.41/25.72 & 4.24/4.20 
& 1020/990 & 25.92/25.80 & 4.16/4.09 \\

& Qwen2.5-0.5B 
& 390/390 & 43.01/45.39 & 7.75/8.35 
& 630/690 & 43.36/45.69 & 3.37/3.32 
& 1050/1050 & 44.21/47.19 & 1.88/1.87 \\

& Gemma3-270M 
& 270/300 & 23.66/25.25 & 4.29/4.21 
& 600/630 & 23.99/25.45 & 4.08/3.96 
& 870/990 & 25.90/26.79 & 4.06/2.85 \\

& Gemma3-1B 
& 270/300 & 23.80/24.18 & 4.18/4.31 
& 510/630 & 24.73/24.80 & 4.06/3.73 
& 810/1020 & 25.46/27.53 & 4.02/2.53 \\

\midrule

\multirow{5}{*}{PIQA}
& GPT2-124M 
& 690/540 & 62.73/63.44 & 24.60/23.28 
& 1140/1140 & 63.33/63.44 & 24.24/22.28 
& 1800/1770 & 63.33/63.49 & 24.09/22.60 \\

& GPT2-355M 
& 630/510 & 67.90/68.61 & 18.65/17.20 
& 1140/1140 & 68.12/68.77 & 18.20/16.72 
& 1200/1650 & 68.34/68.93 & 18.15/16.49 \\

& Qwen2.5-0.5B 
& 660/630 & 70.24/70.51 & 15.40/13.92 
& 1170/1110 & 70.40/70.57 & 15.11/13.68 
& 1170/1800 & 70.40/70.73 & 14.40/13.57 \\

& Gemma3-270M 
& 570/690 & 65.94/67.57 & 20.49/20.57 
& 1260/1290 & 66.27/68.93 & 17.99/17.96 
& 1830/1860 & 66.54/69.15 & 16.29/15.76 \\

& Gemma3-1B 
& 690/540 & 71.55/76.44 & 15.23/15.35 
& 1200/1110 & 73.12/76.71 & 11.14/10.99 
& 1800/1830 & 73.18/76.88 & 10.89/10.97 \\

\midrule

\multirow{5}{*}{ARC-C}
& GPT2-124M 
& 30/30 & 21.07/21.07 & 51.20/50.25 
& 90/90 & 22.41/22.41 & 41.53/42.63 
& 120/120 & 22.74/22.74 & 40.67/42.30 \\

& GPT2-355M 
& 30/30 & 26.09/26.09 & 36.80/30.29 
& 90/90 & 28.43/26.09 & 27.04/26.84 
& 120/120 & 28.76/26.76 & 26.61/26.73 \\

& Qwen2.5-0.5B 
& 30/30 & 35.79/38.13 & 18.39/13.24 
& 90/90 & 36.12/38.46 & 15.45/11.87 
& 120/120 & 36.79/38.80 & 15.26/11.76 \\

& Gemma3-270M 
& 30/30 & 25.08/25.75 & 109.60/22.39 
& 90/90 & 26.42/26.76 & 42.23/16.73 
& 120/120 & 27.42/27.42 & 23.59/12.63 \\

& Gemma3-1B 
& 30/30 & 25.08/40.13 & 103.96/12.44 
& 90/90 & 29.43/42.14 & 36.43/10.16 
& 120/120 & 31.44/42.81 & 25.34/8.94 \\

\midrule

\multirow{5}{*}{ARC-E}
& GPT2-124M 
& 90/90 & 45.79/45.26 & 30.33/25.03 
& 180/180 & 46.14/45.96 & 28.04/23.47 
& 240/240 & 46.32/46.14 & 27.49/22.96 \\

& GPT2-355M 
& 90/90 & 50.35/52.81 & 18.61/14.96 
& 180/180 & 50.88/52.81 & 17.16/14.21 
& 240/240 & 51.75/53.33 & 16.92/13.85 \\

& Qwen2.5-0.5B 
& 90/90 & 69.82/69.82 & 8.43/6.83 
& 180/180 & 70.70/69.30 & 7.70/6.72 
& 240/240 & 71.05/70.18 & 7.62/6.66 \\

& Gemma3-270M 
& 90/90 & 45.61/61.75 & 49.74/20.91 
& 180/180 & 50.70/61.93 & 40.51/13.04 
& 240/240 & 52.81/62.28 & 36.51/11.31 \\

& Gemma3-1B 
& 90/90 & 45.44/77.02 & 154.53/10.71 
& 180/180 & 58.07/77.72 & 94.49/6.04 
& 240/240 & 69.47/77.89 & 66.79/5.21 \\

\bottomrule
\end{tabular}%
}
\label{tab:progress_matched_all_seq128}
\vspace{-2mm}
\end{table*}

\subsubsection{PEFT (LoRA)}
\

\noindent \textbf{Settings.}
We evaluate MobileFineTuner's PEFT capability on five models and six datasets with sequence lengths of 128 and 256, resulting in 60 tasks. For all tasks, we use batch size 8, LoRA rank 8, $\alpha=32$, dropout 0.1, and learning rate $2\times10^{-4}$. MobileFineTuner and the PyTorch baseline use the same fine-tuning configuration. MobileFineTuner enables its built-in memory optimizations during mobile-side execution. The experiments are conducted on iQOO~15.

\noindent \textbf{Results-1 Final Performance and System Metrics.}
Tab.~\ref{tab:peft_metrics_all_seq128} reports representative results on MMLU, PIQA, ARC-C, and ARC-E with sequence length 128, while the remaining results are provided in Appendix~\ref{sec:appendixA}. We compare MobileFineTuner with PyTorch in terms of final loss, accuracy, and PPL under the same configuration. Across different models and datasets, MobileFineTuner achieves results close to the server-side PyTorch baseline, indicating that it can reproduce standard LoRA fine-tuning behavior directly on mobile phones. We also report system metrics, including fine-tuning time, energy consumption, and peak RSS, to characterize the execution cost of mobile-side PEFT.

\noindent \textbf{Results-2 Runtime Testing Performance.}
To further examine the fine-tuning dynamics, we report runtime testing loss, PPL, and accuracy throughout the training process. We compare the loss curves of MobileFineTuner and PyTorch, and evaluate testing PPL and accuracy at 30\%, 60\%, and 90\% of the total fine-tuning steps for convergence. As shown in Tab.~\ref{tab:progress_matched_all_seq128}, MobileFineTuner maintains testing performance close to PyTorch during training. The loss, PPL, and accuracy evolve in similar trends at different training stages, indicating that MobileFineTuner not only reaches comparable final performance but also reproduces the runtime fine-tuning behavior of the server-side baseline. The remaining results are provided in Appendix~\ref{appB}. We also provide the runtime testing loss curves in Appendix~\ref{appC}.

\subsection{Resource-aware Training Runtime}
\subsubsection{Memory Optimization}

\noindent \textbf{Settings.}
We evaluate MobileFineTuner's built-in memory optimizations on WikiText-2 using GPT2-small, GPT2-medium, Gemma3-270M, and Qwen2.5-0.5B with PEFT. We set the batch size to 8, sequence length to 256, LoRA rank to 8, LoRA $\alpha$ to 32, and learning rate to $2\times10^{-4}$.

\begin{figure}[t]
\centering
\includegraphics[width=0.95\linewidth]{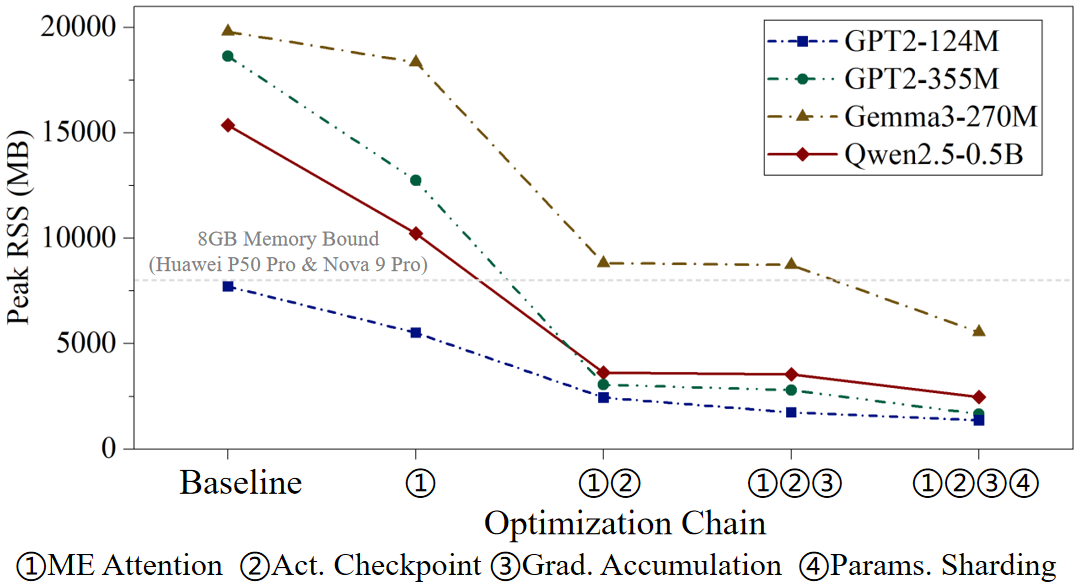}
\vspace{-2mm}
\caption{Peak RSS with Optimization Chains.}
\label{chain}
\vspace{-4mm}

\end{figure}

\noindent \textbf{Results-1 Optimization Chains.}
Fig.~\ref{chain} shows the peak RSS under different optimization chains. We enable four optimization components: \ding{172} memory-efficient attention, \ding{173} activation checkpointing, \ding{174} gradient accumulation, and \ding{175} parameter sharding. These components reduce different sources of memory pressure can therefore be composed as an optimization chain. The results show that memory optimization is essential for practical on-device PEFT. Without optimization, several models fail due to out-of-memory errors, especially on Huawei P50 Pro and Huawei Nova 9 Pro, each with 8~GB RAM. After enabling the full optimization chain, all evaluated devices can complete fine-tuning. Table~\ref{tab:min_exec_config_all} summarizes the minimum optimization configuration required for each model-device pair to complete fine-tuning.

\noindent \textbf{Results-2 Ablation Study on Gradient Accumulation.}
We further examine whether gradient accumulation affects fine-tuning performance, since it changes the execution granularity by splitting a batch into multiple micro-batches. We vary the accumulation steps among 2, 4, and 8, corresponding to micro-batch sizes of 4, 2, and 1 under a total batch size of 8, denoted as \texttt{b4a2}, \texttt{b2a4}, and \texttt{b1a8}. As shown in Table~\ref{grad_list}, the convergence steps, final loss, and PPL remain nearly unchanged across different settings, indicating that gradient accumulation reduces memory pressure without compromising fine-tuning accuracy.

\subsubsection{Energy Optimization}

\noindent \textbf{Settings.}
We evaluate the energy-aware scheduling mechanism using PEFT on Qwen2.5-0.5B with the WikiText-2 dataset. Experiments are conducted on a Huawei Nova 9 Pro with batch size 8, sequence length 128, LoRA rank 8, LoRA $\alpha=16$, and learning rate $2\times10^{-4}$. We set $K=1$, $\mu=60\%$, and $\rho=50\%$, and record the wall-clock time at each step.

\begin{figure}[t]
\centering
\includegraphics[width=0.8\linewidth]{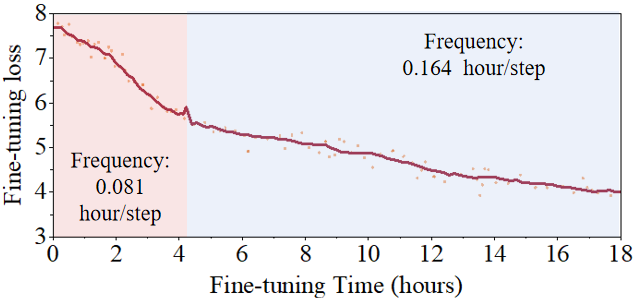}
\caption{Loss with Computation Scheduling on Qwen2.5-0.5B@WikiTest-2.}
\vspace{-2mm}
\label{schedule}

\end{figure}

\noindent \textbf{Results.}
As shown in Fig.~\ref{schedule}, the battery level drops below the 60\% threshold at step 53, approximately 4 hours into fine-tuning. MobileFineTuner then reduces the computation frequency, increasing the per-step interval from 0.081 to 0.164 hours.

\begin{table}[t]
\centering
\caption{Minimum optimization configuration required to complete on-device fine-tuning. ``Any'' means the task can be completed without any optmizations.}
\label{tab:min_exec_config_all}
\renewcommand{\arraystretch}{1.15}
\setlength{\tabcolsep}{6pt}
\resizebox{\linewidth}{!}{%
\begin{tabular}{lcccc}
\toprule
\textbf{Device} 
& \textbf{GPT2-124M} 
& \textbf{GPT2-355M} 
& \textbf{Qwen2.5-0.5B} 
& \textbf{Gemma3-270M} \\
\midrule
Huawei P50 Pro     & \ding{172} & \ding{172}\ding{173} & \ding{172}\ding{173} & \ding{172}\ding{173}\ding{174}\ding{175} \\
Huawei Nova 9 Pro  & \ding{172} & \ding{172}\ding{173} & \ding{172}\ding{173} & \ding{172}\ding{173}\ding{174}\ding{175} \\
iQOO 15            & Any        & Any                  & Any                  & Any \\
MacBook Air 2023   & Any        & Any                  & Any                  & Any \\
\bottomrule
\end{tabular}%
}
\vspace{2pt}
\begin{center}
\footnotesize
\ding{172}ME Attention \ \ding{173}Act. Checkpoint \
\ding{174}Grad. Accumulation \ \ding{175}Params. Sharding
\end{center}
\vspace{-4mm}
\end{table}

\subsection{Comparison with Termux Pipeline}

\noindent \textbf{Settings.}
We compare MobileFineTuner with the Termux-based pipeline to evaluate the benefit of its mobile-native design. The Termux baseline runs PyTorch inside the Termux emulator on mobile phone, while MobileFineTuner executes the same PEFT task through its native C++ runtime. Both methods fine-tune Qwen2.5-0.5B on QNLI using batch size 8, sequence length 128, LoRA rank 8, LoRA $\alpha=16$, and learning rate $2\times10^{-4}$ on an iQOO phone.

\noindent \textbf{Results.}
As shown in Table~\ref{tab:termux_mobilefinetuner}, MobileFineTuner achieves lower step time and lower peak RSS than the Termux-based pipeline, demonstrating better execution and memory efficiency. More importantly, Termux depends on an external terminal environment and Python runtime, making it difficult to integrate into normal mobile applications. In contrast, MobileFineTuner can be directly embedded into Android applications as a local training backend, making it more suitable for real-world on-device LLM fine-tuning.

\begin{table}[t]
\centering
\caption{Fine-tuning Metrics on Different Gradient Accumulation Steps on Gemma3-270M@WikiText-2.}
\vspace{-2mm}
\resizebox{0.8\linewidth}{!}{%
\begin{tabular}{cccc}
\toprule
Methods & Convergence Steps & Final Loss & Final PPL \\
\midrule
b4a2 & 1219 & 3.37 & 29.27 \\
b2a4 & 1218 & 4.08 & 28.91 \\
b1a8 & 1218 & 4.28 & 26.79 \\
\bottomrule
\end{tabular}%
}
\vspace{-2mm}
\label{grad_list}
\end{table}

\section{Results of Campus Health Agent}
\label{case_results_sec}

In this section, we report the configuration and application-level results of the campus health-agent case study. The health-agent application synchronizes wearable records from the smartwatch to the user's phone, where the local LLM answers user questions based on the user's historical health records. Our goal is to evaluate whether continuous on-device fine-tuning can make the local agent better understand each individual user and generate more personalized responses.

For each user, the application constructs personalized QA pairs by filling the generated health-agent templates with local health statistics. These QA pairs are generated from the user's own wearable records and used to fine-tune the local LLM on the user's phone. In our deployment, fine-tuning is performed locally at night, when the phone is less likely to interfere with normal user interaction. As the user continues to use the application over the three-month period, the local adapter is progressively updated with the user's accumulated records. We use Qwen2.5-0.5B as the base model and perform LoRA fine-tuning with sequence length 256, LoRA rank 8, LoRA $\alpha=16$, and learning rate $2\times10^{-4}$. The fine-tuned model evaluated in this section is the personalized local model obtained after the three-month deployment.

\noindent \textbf{Evaluation metric.}
We use GPT-5.5 as an external judge to evaluate the quality of health-agent responses on a 0--5 scale. Scores of 0--1 indicate mostly unusable answers that are incorrect, generic, ungrounded, hallucinated, or fail to answer the question. Scores of 2--3 indicate partially useful answers with some relevant content but limited grounding, weak specificity, or incomplete phrasing. Scores of 4--5 indicate strong answers that are grounded in the user's records, directly answer the question, and provide safe and useful suggestions. For each category, we evaluate the responses generated by each user's personalized model and report the average judge score across all users.

\begin{figure}[t]
\centering
\includegraphics[width=\linewidth]{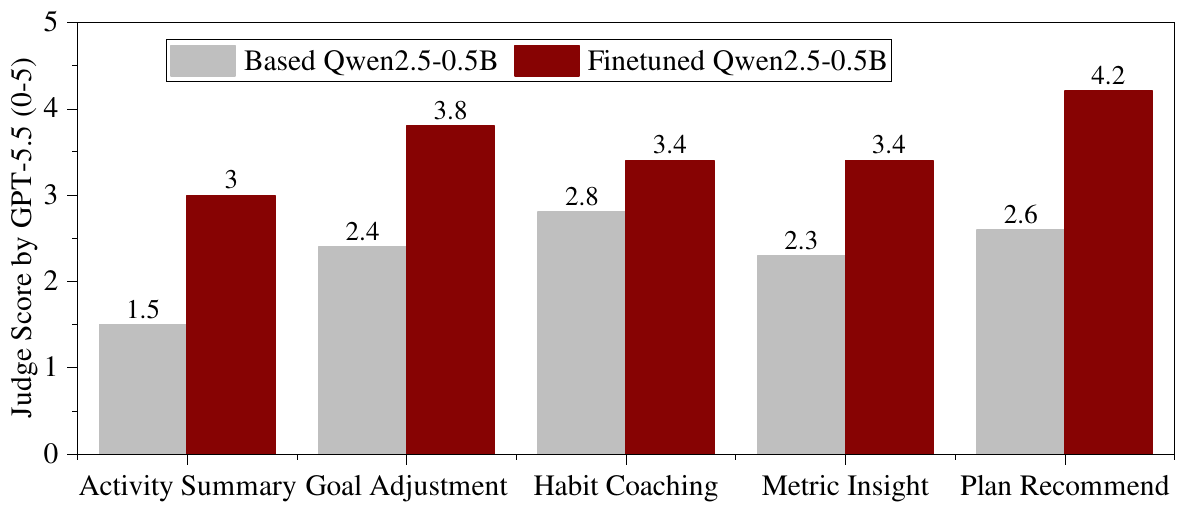}
\caption{LLM Judge Score of Agent Output.}
\label{case_results}
\vspace{-2mm}
\end{figure}

\noindent \textbf{Results.}
As shown in Fig.~\ref{case_results}, the fine-tuned local LLM consistently achieves higher judge scores than the base Qwen2.5-0.5B model across all five health-agent categories. The improvement is particularly clear in \textit{Goal Adjustment} and \textit{Plan Recommendation}, where the agent needs to connect the user's recent records with their historical baseline and provide actionable suggestions. These results show that after three months of local use and nightly fine-tuning, MobileFineTuner enables each user's local LLM to better leverage personalized health records, improving response quality through on-device adaptation while keeping raw wearable data on the phone.

\section{Ethical Statement}
The case study involves wearable sensing data collected from student volunteers and was approved by an institutional review board. All participants provided informed consent. We remove direct identifiers, replace participant identities with random IDs, and use only anonymized or derived records for constructing the CHQA dataset.

\begin{table}[t]
\centering
\caption{Comparison with Termux Pipeline on Qwen2.5-0.5B@PIQA.}
\vspace{-2mm}
\resizebox{0.8\linewidth}{!}{%
\begin{tabular}{lcc}
\toprule
Methods & Average Step Time (s) & Peak RSS (MB) \\
\midrule
Termux + PyTorch & 489.16 & 3313.69 \\
MobileFineTuner  & 107.36 & 2395.76 \\
\bottomrule
\end{tabular}%
}
\label{tab:termux_mobilefinetuner}
\vspace{-2mm}
\end{table}

\section{Conclusion}
In this paper, we presented MobileFineTuner, a mobile-native open-source framework for on-device LLM fine-tuning on commodity mobile phones. MobileFineTuner provides new insights into the behavior of large-model training under mobile hardware constraints and establishes a foundation for future research on resource-aware, privacy-preserving on-device learning. Looking forward, we plan to extend MobileFineTuner to support heterogeneous mobile platforms and hardware accelerators, and to explore integrating federated and collaborative LLM fine-tuning across mobile devices. We believe MobileFineTuner opens a promising new direction for mobile systems research by enabling large-scale learning capabilities to move closer to user data and end devices.

\bibliographystyle{ACM-Reference-Format}
\bibliography{example_paper}

\appendix

\onecolumn

\clearpage

\section{Final Performance and System Metrics}

\label{sec:appendixA}

\begin{table*}[h]
\centering
\caption{PEFT (LoRA) metrics on WikiText-2 with Seq 128. (Initial Loss/Acc/PPL $\rightarrow$ Best Final Loss/Acc/PPL)}
\renewcommand{\arraystretch}{1.15}
\resizebox{\textwidth}{!}{%
\begin{tabular}{lccccccc}
\toprule
\multirow{2}{*}{Model} 
& \multicolumn{2}{c}{Loss} 
& \multicolumn{2}{c}{PPL}
& \multicolumn{3}{c}{System Metrics} \\
\cmidrule(lr){2-3}\cmidrule(lr){4-5}\cmidrule(lr){6-8}
& MobileFineTuner & PyTorch 
& MobileFineTuner & PyTorch
& Time (h) & Energy (kJ) & Peak RSS (MB) \\
\midrule
GPT2-124M 
& $4.259 \rightarrow 3.461$ 
& $3.855 \rightarrow 3.072$ 
& $69.92 \rightarrow 35.87$ 
& $50.31 \rightarrow 29.35$ 
& 40.96 & 114.48 & 1229.54 \\

GPT2-355M 
& $3.817 \rightarrow 3.569$ 
& $3.627 \rightarrow 3.428$ 
& $49.34 \rightarrow 26.83$ 
& $36.14 \rightarrow 21.74$ 
& 94.92 & 257.10 & 3301.83 \\

Qwen2.5-0.5B 
& $3.532 \rightarrow 2.646$ 
& $3.304 \rightarrow 2.405$ 
& $33.56 \rightarrow 22.33$ 
& $24.81 \rightarrow 17.97$ 
& 79.61 & 220.24 & 2887.23 \\

Gemma3-270M 
& $5.613 \rightarrow 4.075$ 
& $3.914 \rightarrow 2.792$ 
& $109.63 \rightarrow 28.91$ 
& $31.06 \rightarrow 20.75$ 
& 247.59 & 707.28 & 4014.10 \\

Gemma3-1B 
& $5.544 \rightarrow 3.563$ 
& $3.377 \rightarrow 2.439$ 
& $138.76 \rightarrow 28.61$ 
& $20.91 \rightarrow 13.82$ 
& 490.65 & 1232.15 & 6773.37 \\
\bottomrule
\end{tabular}%
}
\label{tab:peft_metrics_wikitext_seq128}
\end{table*}

\begin{table*}[h]
\centering
\caption{PEFT (LoRA) metrics on WikiText-2 with Seq 256. (Initial Loss/Acc/PPL $\rightarrow$ Best Final Loss/Acc/PPL)}
\renewcommand{\arraystretch}{1.15}
\resizebox{\textwidth}{!}{%
\begin{tabular}{lccccccc}
\toprule
\multirow{2}{*}{Model} 
& \multicolumn{2}{c}{Loss} 
& \multicolumn{2}{c}{PPL}
& \multicolumn{3}{c}{System Metrics} \\
\cmidrule(lr){2-3}\cmidrule(lr){4-5}\cmidrule(lr){6-8}
& MobileFineTuner & PyTorch 
& MobileFineTuner & PyTorch
& Time (h) & Energy (kJ) & Peak RSS (MB) \\
\midrule
GPT2-124M 
& $4.378 \rightarrow 3.397$ 
& $4.008 \rightarrow 2.818$ 
& $56.91 \rightarrow 31.67$ 
& $40.20 \rightarrow 25.49$ 
& 63.94 & 212.47 & 1452.20 \\

GPT2-355M 
& $3.981 \rightarrow 3.150$ 
& $3.634 \rightarrow 2.920$ 
& $40.61 \rightarrow 25.08$ 
& $29.40 \rightarrow 18.75$ 
& 173.17 & 503.53 & 3783.41 \\

Qwen2.5-0.5B 
& $2.980 \rightarrow 2.854$ 
& $2.731 \rightarrow 1.851$ 
& $23.99 \rightarrow 17.68$ 
& $18.95 \rightarrow 14.82$ 
& 144.56 & 427.86 & 3437.00 \\

Gemma3-270M 
& $4.356 \rightarrow 3.002$ 
& $3.205 \rightarrow 2.739$ 
& $65.33 \rightarrow 22.98$ 
& $23.18 \rightarrow 17.45$ 
& 314.14 & 919.75 & 4707.23 \\
\bottomrule
\end{tabular}%
}
\label{tab:peft_metrics_wikitext_seq256}
\end{table*}

\begin{table*}[h]
\centering
\caption{PEFT (LoRA) metrics on MMLU with Seq 256. (Initial Loss/Acc/PPL $\rightarrow$ Best Final Loss/Acc/PPL)}
\renewcommand{\arraystretch}{1.15}
\resizebox{0.99\textwidth}{!}{%
\begin{tabular}{lccccccccc}
\toprule
\multirow{2}{*}{Model} 
& \multicolumn{2}{c}{Loss} 
& \multicolumn{2}{c}{Accuracy (\%)} 
& \multicolumn{2}{c}{PPL}
& \multicolumn{3}{c}{System Metrics} \\
\cmidrule(lr){2-3}\cmidrule(lr){4-5}\cmidrule(lr){6-7}\cmidrule(lr){8-10}
& MobileFineTuner & PyTorch 
& MobileFineTuner & PyTorch 
& MobileFineTuner & PyTorch
& Time (h) & Energy (kJ) & Peak RSS (MB) \\
\midrule
GPT2-124M 
& $2.231 \rightarrow 1.353$ 
& $1.533 \rightarrow 1.340$ 
& $21.35 \rightarrow 27.06$ 
& $24.87 \rightarrow 27.36$ 
& $16.31 \rightarrow 4.24$ 
& $4.63 \rightarrow 3.82$ 
& 36.46 & 90.37 & 1484.86 \\

GPT2-355M 
& $1.707 \rightarrow 1.308$ 
& $1.256 \rightarrow 0.807$ 
& $24.78 \rightarrow 26.12$ 
& $23.87 \rightarrow 30.07$ 
& $6.10 \rightarrow 4.57$ 
& $3.51 \rightarrow 2.24$ 
& 98.90 & 253.36 & 3817.18 \\

Qwen2.5-0.5B 
& $0.883 \rightarrow 0.850$ 
& $1.135 \rightarrow 0.506$ 
& $44.72 \rightarrow 45.29$ 
& $46.68 \rightarrow 48.24$ 
& $2.42 \rightarrow 2.34$ 
& $3.11 \rightarrow 1.66$ 
& 40.54 & 118.09 & 3164.37 \\

Gemma3-270M 
& $1.652 \rightarrow 1.461$ 
& $1.346 \rightarrow 0.974$ 
& $24.45 \rightarrow 27.10$ 
& $23.96 \rightarrow 28.29$ 
& $5.76 \rightarrow 4.06$ 
& $3.84 \rightarrow 2.65$ 
& 98.93 & 227.86 & 4658.07 \\
\bottomrule
\end{tabular}%
}
\label{tab:peft_metrics_mmlu_seq256}
\end{table*}

\begin{table*}[h]
\centering
\caption{PEFT (LoRA) metrics on PIQA with Seq 256. (Initial Loss/Acc/PPL $\rightarrow$ Best Final Loss/Acc/PPL)}
\renewcommand{\arraystretch}{1.15}
\resizebox{0.99\textwidth}{!}{%
\begin{tabular}{lccccccccc}
\toprule
\multirow{2}{*}{Model} 
& \multicolumn{2}{c}{Loss} 
& \multicolumn{2}{c}{Accuracy (\%)} 
& \multicolumn{2}{c}{PPL}
& \multicolumn{3}{c}{System Metrics} \\
\cmidrule(lr){2-3}\cmidrule(lr){4-5}\cmidrule(lr){6-7}\cmidrule(lr){8-10}
& MobileFineTuner & PyTorch 
& MobileFineTuner & PyTorch 
& MobileFineTuner & PyTorch
& Time (h) & Energy (kJ) & Peak RSS (MB) \\
\midrule
GPT2-124M 
& $3.203 \rightarrow 3.131$ 
& $3.162 \rightarrow 2.865$ 
& $62.62 \rightarrow 63.49$ 
& $63.00 \rightarrow 64.09$ 
& $27.57 \rightarrow 24.12$ 
& $26.16 \rightarrow 22.32$ 
& 36.41 & 89.26 & 1479.26 \\

GPT2-355M 
& $3.522 \rightarrow 2.933$ 
& $2.920 \rightarrow 2.762$ 
& $67.57 \rightarrow 68.50$ 
& $68.12 \rightarrow 69.26$ 
& $20.39 \rightarrow 18.08$ 
& $19.28 \rightarrow 16.82$ 
& 94.69 & 253.92 & 3834.39 \\

Qwen2.5-0.5B 
& $2.735 \rightarrow 2.356$ 
& $2.520 \rightarrow 2.322$ 
& $69.97 \rightarrow 70.62$ 
& $71.33 \rightarrow 71.44$ 
& $16.44 \rightarrow 14.96$ 
& $15.15 \rightarrow 14.19$ 
& 52.98 & 191.82 & 3223.19 \\

Gemma3-270M 
& $3.727 \rightarrow 2.974$ 
& $2.673 \rightarrow 2.663$ 
& $60.56 \rightarrow 68.17$ 
& $66.70 \rightarrow 69.15$ 
& $41.55 \rightarrow 19.57$ 
& $14.49 \rightarrow 14.34$ 
& 85.19 & 245.04 & 4835.58 \\
\bottomrule
\end{tabular}%
}
\label{tab:peft_metrics_piqa_seq256}
\end{table*}

\begin{table*}[h]
\centering
\caption{PEFT (LoRA) metrics on HellaSwag with Seq 128. (Initial Loss/Acc/PPL $\rightarrow$ Best Final Loss/Acc/PPL)}
\renewcommand{\arraystretch}{1.15}
\resizebox{0.99\textwidth}{!}{%
\begin{tabular}{lccccccccc}
\toprule
\multirow{2}{*}{Model} 
& \multicolumn{2}{c}{Loss} 
& \multicolumn{2}{c}{Accuracy (\%)} 
& \multicolumn{2}{c}{PPL}
& \multicolumn{3}{c}{System Metrics} \\
\cmidrule(lr){2-3}\cmidrule(lr){4-5}\cmidrule(lr){6-7}\cmidrule(lr){8-10}
& MobileFineTuner & PyTorch 
& MobileFineTuner & PyTorch 
& MobileFineTuner & PyTorch
& Time (h) & Energy (kJ) & Peak RSS (MB) \\
\midrule
GPT2-124M 
& $3.168 \rightarrow 2.777$ 
& $2.987 \rightarrow 2.966$ 
& $29.90 \rightarrow 30.87$ 
& $30.04 \rightarrow 30.75$ 
& $26.92 \rightarrow 19.90$ 
& $23.19 \rightarrow 20.00$ 
& 45.33 & 112.54 & 1240.80 \\

GPT2-355M 
& $3.066 \rightarrow 2.764$ 
& $2.663 \rightarrow 2.431$ 
& $36.32 \rightarrow 37.40$ 
& $37.11 \rightarrow 37.86$ 
& $24.01 \rightarrow 16.09$ 
& $17.02 \rightarrow 16.03$ 
& 122.49 & 316.01 & 3343.12 \\

Qwen2.5-0.5B 
& $3.051 \rightarrow 2.389$ 
& $3.186 \rightarrow 2.386$ 
& $48.04 \rightarrow 48.54$ 
& $48.45 \rightarrow 48.45$ 
& $18.89 \rightarrow 16.45$ 
& $18.36 \rightarrow 12.36$ 
& 78.49 & 230.08 & 2887.45 \\

Gemma3-270M 
& $3.811 \rightarrow 3.544$ 
& $2.729 \rightarrow 2.582$ 
& $31.71 \rightarrow 38.77$ 
& $35.91 \rightarrow 39.40$ 
& $45.20 \rightarrow 34.61$ 
& $15.32 \rightarrow 13.22$ 
& 123.76 & 291.06 & 4042.04 \\

Gemma3-1B 
& $3.684 \rightarrow 3.081$ 
& $2.498 \rightarrow 2.235$ 
& $42.69 \rightarrow 56.05$ 
& $54.50 \rightarrow 56.60$ 
& $39.81 \rightarrow 21.78$ 
& $12.16 \rightarrow 9.34$ 
& 154.00 & 397.32 & 9146.13 \\
\bottomrule
\end{tabular}%
}
\label{tab:peft_metrics_hellaswag_seq128}
\end{table*}

\begin{table*}[h]
\centering
\caption{PEFT (LoRA) metrics on HellaSwag with Seq 256. (Initial Loss/Acc/PPL $\rightarrow$ Best Final Loss/Acc/PPL)}
\renewcommand{\arraystretch}{1.15}
\resizebox{0.99\textwidth}{!}{%
\begin{tabular}{lccccccccc}
\toprule
\multirow{2}{*}{Model} 
& \multicolumn{2}{c}{Loss} 
& \multicolumn{2}{c}{Accuracy (\%)} 
& \multicolumn{2}{c}{PPL}
& \multicolumn{3}{c}{System Metrics} \\
\cmidrule(lr){2-3}\cmidrule(lr){4-5}\cmidrule(lr){6-7}\cmidrule(lr){8-10}
& MobileFineTuner & PyTorch 
& MobileFineTuner & PyTorch 
& MobileFineTuner & PyTorch
& Time (h) & Energy (kJ) & Peak RSS (MB) \\
\midrule
GPT2-124M 
& $3.396 \rightarrow 3.168$ 
& $3.291 \rightarrow 2.987$ 
& $29.90 \rightarrow 30.71$ 
& $30.04 \rightarrow 30.40$ 
& $26.92 \rightarrow 19.19$ 
& $23.20 \rightarrow 22.97$ 
& 36.59 & 88.70 & 1506.03 \\

GPT2-355M 
& $3.066 \rightarrow 2.764$ 
& $3.049 \rightarrow 2.431$ 
& $36.32 \rightarrow 37.40$ 
& $37.11 \rightarrow 37.62$ 
& $24.01 \rightarrow 16.09$ 
& $16.95 \rightarrow 16.25$ 
& 99.04 & 251.70 & 3849.85 \\

Qwen2.5-0.5B 
& $2.506 \rightarrow 2.385$ 
& $2.481 \rightarrow 2.289$ 
& $48.03 \rightarrow 48.32$ 
& $48.43 \rightarrow 48.43$ 
& $18.78 \rightarrow 16.40$ 
& $12.27 \rightarrow 11.23$ 
& 61.45 & 179.63 & 3222.88 \\

Gemma3-270M 
& $3.811 \rightarrow 3.133$ 
& $2.755 \rightarrow 2.320$ 
& $31.71 \rightarrow 35.17$ 
& $36.04 \rightarrow 37.39$ 
& $45.20 \rightarrow 22.94$ 
& $15.72 \rightarrow 10.17$ 
& 99.17 & 230.63 & 4813.57 \\
\bottomrule
\end{tabular}%
}
\label{tab:peft_metrics_hellaswag_seq256}
\end{table*}

\begin{table*}[h]
\centering
\caption{PEFT (LoRA) metrics on ARC-Challenge with Seq 256. (Initial Loss/Acc/PPL $\rightarrow$ Best Final Loss/Acc/PPL)}
\renewcommand{\arraystretch}{1.15}
\resizebox{0.99\textwidth}{!}{%
\begin{tabular}{lccccccccc}
\toprule
\multirow{2}{*}{Model} 
& \multicolumn{2}{c}{Loss} 
& \multicolumn{2}{c}{Accuracy (\%)} 
& \multicolumn{2}{c}{PPL}
& \multicolumn{3}{c}{System Metrics} \\
\cmidrule(lr){2-3}\cmidrule(lr){4-5}\cmidrule(lr){6-7}\cmidrule(lr){8-10}
& MobileFineTuner & PyTorch 
& MobileFineTuner & PyTorch 
& MobileFineTuner & PyTorch
& Time (h) & Energy (kJ) & Peak RSS (MB) \\
\midrule
GPT2-124M 
& $3.654 \rightarrow 2.774$ 
& $3.527 \rightarrow 2.713$ 
& $21.07 \rightarrow 22.74$ 
& $21.07 \rightarrow 22.74$ 
& $50.99 \rightarrow 40.55$ 
& $49.99 \rightarrow 41.83$ 
& 9.72 & 21.62 & 1438.16 \\

GPT2-355M 
& $3.391 \rightarrow 2.423$ 
& $3.056 \rightarrow 2.451$ 
& $26.09 \rightarrow 29.43$ 
& $26.09 \rightarrow 26.76$ 
& $36.61 \rightarrow 26.97$ 
& $29.87 \rightarrow 26.40$ 
& 26.39 & 62.09 & 3758.36 \\

Qwen2.5-0.5B 
& $2.814 \rightarrow 2.612$ 
& $2.507 \rightarrow 1.982$ 
& $35.45 \rightarrow 37.46$ 
& $38.13 \rightarrow 38.80$ 
& $18.56 \rightarrow 16.31$ 
& $13.39 \rightarrow 11.87$ 
& 16.34 & 43.80 & 3115.82 \\

Gemma3-270M 
& $4.697 \rightarrow 2.907$ 
& $3.118 \rightarrow 3.030$ 
& $25.08 \rightarrow 27.42$ 
& $25.42 \rightarrow 28.76$ 
& $109.62 \rightarrow 18.30$ 
& $22.60 \rightarrow 20.69$ 
& 26.31 & 56.27 & 4749.05 \\
\bottomrule
\end{tabular}%
}
\label{tab:peft_metrics_arc_c_seq256}
\end{table*}

\begin{table*}[h]
\centering
\caption{PEFT (LoRA) metrics on ARC-Easy with Seq 256. (Initial Loss/Acc/PPL $\rightarrow$ Best Final Loss/Acc/PPL)}
\renewcommand{\arraystretch}{1.15}
\resizebox{0.99\textwidth}{!}{%
\begin{tabular}{lccccccccc}
\toprule
\multirow{2}{*}{Model} 
& \multicolumn{2}{c}{Loss} 
& \multicolumn{2}{c}{Accuracy (\%)} 
& \multicolumn{2}{c}{PPL}
& \multicolumn{3}{c}{System Metrics} \\
\cmidrule(lr){2-3}\cmidrule(lr){4-5}\cmidrule(lr){6-7}\cmidrule(lr){8-10}
& MobileFineTuner & PyTorch 
& MobileFineTuner & PyTorch 
& MobileFineTuner & PyTorch
& Time (h) & Energy (kJ) & Peak RSS (MB) \\
\midrule
GPT2-124M 
& $3.075 \rightarrow 2.748$ 
& $2.797 \rightarrow 2.692$ 
& $42.28 \rightarrow 46.84$ 
& $43.33 \rightarrow 46.14$ 
& $42.46 \rightarrow 27.73$ 
& $31.73 \rightarrow 23.12$ 
& 19.10 & 43.52 & 1457.67 \\

GPT2-355M 
& $2.651 \rightarrow 2.455$ 
& $2.585 \rightarrow 2.314$ 
& $45.26 \rightarrow 52.11$ 
& $51.23 \rightarrow 53.33$ 
& $29.79 \rightarrow 17.02$ 
& $18.01 \rightarrow 13.79$ 
& 49.38 & 126.96 & 3818.65 \\

Qwen2.5-0.5B 
& $1.897 \rightarrow 1.828$ 
& $2.527 \rightarrow 1.645$ 
& $67.72 \rightarrow 70.88$ 
& $69.82 \rightarrow 70.53$ 
& $10.64 \rightarrow 7.68$ 
& $7.64 \rightarrow 6.75$ 
& 30.65 & 90.37 & 3154.21 \\

Gemma3-270M 
& $3.907 \rightarrow 3.523$ 
& $2.547 \rightarrow 2.287$ 
& $42.63 \rightarrow 62.81$ 
& $52.63 \rightarrow 63.68$ 
& $49.75 \rightarrow 33.89$ 
& $12.78 \rightarrow 9.84$ 
& 49.39 & 114.76 & 4458.83 \\
\bottomrule
\end{tabular}%
}
\label{tab:peft_metrics_arc_e_seq256}
\end{table*}

\FloatBarrier

\clearpage

\section{Fine-tuning Runtime Testing Accuracy/PPL}
\label{appB}

\begin{table*}[h]
\centering
\caption{Runtime Testing PPL on WikiText-2 with Seq 128. (M) indicates MobileFineTuner. (P) means Pytorch.}
\renewcommand{\arraystretch}{1.15}
\resizebox{0.73\textwidth}{!}{%
\begin{tabular}{lcccccc}
\toprule
\multirow{2}{*}{Model} 
& \multicolumn{2}{c}{30\% Progress} 
& \multicolumn{2}{c}{60\% Progress} 
& \multicolumn{2}{c}{90\% Progress} \\
\cmidrule(lr){2-3}\cmidrule(lr){4-5}\cmidrule(lr){6-7}
& Step (M/P) & PPL (M/P) 
& Step (M/P) & PPL (M/P) 
& Step (M/P) & PPL (M/P) \\
\midrule
GPT2-124M 
& 720/720 & 38.37/30.88 
& 1410/1410 & 36.85/29.77 
& 2130/2130 & 36.01/29.40 \\

GPT2-355M 
& 720/720 & 28.47/22.67 
& 1440/1440 & 27.05/21.99 
& 1830/2130 & 26.83/21.77 \\

Qwen2.5-0.5B 
& 240/240 & 23.27/18.96 
& 480/480 & 22.55/18.13 
& 720/720 & 22.33/17.97 \\

Gemma3-270M 
& 480/750 & 33.48/22.66 
& 900/1470 & 30.70/21.44 
& 1350/2190 & 29.34/20.85 \\

Gemma3-1B 
& 540/750 & 33.39/14.65 
& 1020/1470 & 29.90/14.05 
& 1560/2190 & 28.82/13.86 \\
\bottomrule
\end{tabular}%
}
\label{tab:progress_matched_wikitext_seq128}
\end{table*}

\begin{table*}[h]
\centering
\caption{Runtime Testing PPL on WikiText-2 with Seq 256. (M) indicates MobileFineTuner. (P) means Pytorch.}
\renewcommand{\arraystretch}{1.15}
\resizebox{0.73\textwidth}{!}{%
\begin{tabular}{lcccccc}
\toprule
\multirow{2}{*}{Model} 
& \multicolumn{2}{c}{30\% Progress} 
& \multicolumn{2}{c}{60\% Progress} 
& \multicolumn{2}{c}{90\% Progress} \\
\cmidrule(lr){2-3}\cmidrule(lr){4-5}\cmidrule(lr){6-7}
& Step (M/P) & PPL (M/P) 
& Step (M/P) & PPL (M/P) 
& Step (M/P) & PPL (M/P) \\
\midrule
GPT2-124M 
& 360/360 & 33.29/26.88 
& 720/720 & 31.97/25.81 
& 1080/1080 & 31.74/25.52 \\

GPT2-355M 
& 360/360 & 26.22/19.53 
& 720/720 & 25.33/18.92 
& 1050/1050 & 25.09/18.79 \\

Qwen2.5-0.5B 
& 240/240 & 18.20/15.26 
& 480/480 & 17.71/14.91 
& 720/720 & 17.68/14.82 \\

Gemma3-270M 
& 420/390 & 25.33/18.40 
& 720/720 & 23.92/17.82 
& 1110/1110 & 23.04/17.48 \\
\bottomrule
\end{tabular}%
}
\label{tab:progress_matched_wikitext_seq256}
\end{table*}

\begin{table*}[h]
\centering
\caption{Runtime Testing Accuracy and PPL on MMLU with Seq 256. (M) indicates MobileFineTuner. (P) indicates PyTorch.}
\renewcommand{\arraystretch}{1.15}
\resizebox{0.99\textwidth}{!}{%
\begin{tabular}{lccccccccc}
\toprule
\multirow{2}{*}{Model} & \multicolumn{3}{c}{30\% Progress} & \multicolumn{3}{c}{60\% Progress} & \multicolumn{3}{c}{90\% Progress} \\
\cmidrule(lr){2-4}\cmidrule(lr){5-7}\cmidrule(lr){8-10}
& Step (M/P) & Accuracy (M/P) & PPL (M/P) & Step (M/P) & Accuracy (M/P) & PPL (M/P) & Step (M/P) & Accuracy (M/P) & PPL (M/P) \\
\midrule
GPT2-124M & 750/750 & 23.23/23.12 & 4.22/4.35 & 1200/1290 & 25.36/25.12 & 4.07/4.02 & 2340/2340 & 27.16/27.36 & 3.87/3.82 \\
GPT2-355M & 750/750 & 23.38/23.29 & 4.42/4.35 & 1290/1290 & 25.51/25.22 & 4.18/4.16 & 2340/2340 & 28.51/29.61 & 3.78/3.48 \\
Qwen2.5-0.5B & 750/750 & 43.60/45.87 & 2.66/2.66 & 1290/1290 & 45.63/46.52 & 1.95/1.96 & 2370/2370 & 46.90/47.51 & 1.66/1.15 \\
Gemma3-270M & 540/780 & 25.13/25.36 & 4.08/4.07 & 1320/1620 & 25.65/25.39 & 4.05/4.06 & 2340/2340 & 25.92/26.56 & 4.03/4.01 \\
\bottomrule
\end{tabular}%
}
\label{tab:progress_matched_mmlu_seq256}
\end{table*}

\begin{table*}[h]
\centering
\caption{Runtime Testing Accuracy and PPL on PIQA with Seq 256. (M) indicates MobileFineTuner. (P) indicates PyTorch.}
\renewcommand{\arraystretch}{1.15}
\resizebox{0.99\textwidth}{!}{%
\begin{tabular}{lccccccccc}
\toprule
\multirow{2}{*}{Model} & \multicolumn{3}{c}{30\% Progress} & \multicolumn{3}{c}{60\% Progress} & \multicolumn{3}{c}{90\% Progress} \\
\cmidrule(lr){2-4}\cmidrule(lr){5-7}\cmidrule(lr){8-10}
& Step (M/P) & Accuracy (M/P) & PPL (M/P) & Step (M/P) & Accuracy (M/P) & PPL (M/P) & Step (M/P) & Accuracy (M/P) & PPL (M/P) \\
\midrule
GPT2-124M & 600/540 & 62.73/63.44 & 24.28/22.96 & 1200/1140 & 63.11/63.55 & 23.87/22.17 & 1590/1680 & 63.17/63.55 & 23.76/21.95 \\
GPT2-355M & 690/660 & 68.06/68.61 & 18.14/16.85 & 1140/1200 & 68.16/68.66 & 17.87/16.37 & 1140/1620 & 68.23/68.93 & 17.77/16.22 \\
Qwen2.5-0.5B & 630/630 & 70.20/70.35 & 15.01/13.58 & 1050/1050 & 70.24/70.51 & 14.94/13.37 & 1800/1800 & 70.35/70.78 & 13.94/13.22 \\
Gemma3-270M & 510/540 & 65.45/67.90 & 18.69/18.89 & 1230/1290 & 66.43/68.66 & 18.26/17.89 & 1860/1830 & 67.49/69.15 & 14.99/15.01 \\
\bottomrule
\end{tabular}%
}
\label{tab:progress_matched_piqa_seq256}
\end{table*}

\begin{table*}[h]
\centering
\caption{Runtime Testing Accuracy and PPL on ARC-Challenge with Seq 256. (M) indicates MobileFineTuner. (P) indicates PyTorch.}
\renewcommand{\arraystretch}{1.15}
\resizebox{0.99\textwidth}{!}{%
\begin{tabular}{lccccccccc}
\toprule
\multirow{2}{*}{Model} & \multicolumn{3}{c}{30\% Progress} & \multicolumn{3}{c}{60\% Progress} & \multicolumn{3}{c}{90\% Progress} \\
\cmidrule(lr){2-4}\cmidrule(lr){5-7}\cmidrule(lr){8-10}
& Step (M/P) & Accuracy (M/P) & PPL (M/P) & Step (M/P) & Accuracy (M/P) & PPL (M/P) & Step (M/P) & Accuracy (M/P) & PPL (M/P) \\
\midrule
GPT2-124M & 30/30 & 21.07/21.07 & 50.99/49.99 & 90/90 & 22.07/22.41 & 41.35/42.26 & 120/120 & 22.74/22.74 & 40.55/41.83 \\
GPT2-355M & 30/30 & 26.09/26.09 & 36.61/29.87 & 90/90 & 26.63/26.09 & 29.43/26.49 & 120/120 & 28.43/26.76 & 26.97/26.40 \\
Qwen2.5-0.5B & 30/30 & 35.45/38.13 & 18.56/13.39 & 90/90 & 36.12/38.80 & 15.59/11.96 & 120/120 & 36.12/38.80 & 15.42/11.87 \\
Gemma3-270M & 30/30 & 25.08/25.42 & 109.60/22.60 & 90/90 & 26.76/27.09 & 33.70/9.95 & 120/120 & 27.42/28.09 & 18.30/8.76 \\
\bottomrule
\end{tabular}%
}
\label{tab:progress_matched_arc_c_seq256}
\end{table*}

\begin{table*}[h]
\centering
\caption{Runtime Testing Accuracy and PPL on ARC-Easy with Seq 256. (M) indicates MobileFineTuner. (P) indicates PyTorch.}
\renewcommand{\arraystretch}{1.15}
\resizebox{0.99\textwidth}{!}{%
\begin{tabular}{lccccccccc}
\toprule
\multirow{2}{*}{Model} & \multicolumn{3}{c}{30\% Progress} & \multicolumn{3}{c}{60\% Progress} & \multicolumn{3}{c}{90\% Progress} \\
\cmidrule(lr){2-4}\cmidrule(lr){5-7}\cmidrule(lr){8-10}
& Step (M/P) & Accuracy (M/P) & PPL (M/P) & Step (M/P) & Accuracy (M/P) & PPL (M/P) & Step (M/P) & Accuracy (M/P) & PPL (M/P) \\
\midrule
GPT2-124M & 90/90 & 45.79/45.26 & 30.33/25.02 & 180/180 & 46.14/45.96 & 28.04/23.47 & 240/240 & 46.32/46.14 & 27.49/22.96 \\
GPT2-355M & 90/90 & 50.35/52.63 & 18.61/14.96 & 180/180 & 50.88/52.81 & 17.16/14.21 & 240/240 & 51.75/53.33 & 16.92/13.85 \\
Qwen2.5-0.5B & 90/90 & 70.18/69.12 & 8.52/6.84 & 180/180 & 70.53/69.47 & 7.71/6.73 & 240/240 & 70.77/70.35 & 7.67/6.65 \\
Gemma3-270M & 90/90 & 45.61/59.30 & 36.51/20.14& 180/180 & 50.70/61.93 & 49.74/13.54 & 240/240 & 52.81/61.93 & 40.51/11.16 \\
\bottomrule
\end{tabular}%
}
\label{tab:progress_matched_arc_e_seq256}
\end{table*}

\clearpage

\FloatBarrier

\newcommand{\runtimecell}[2]{%
\begin{minipage}[t]{0.29\linewidth}
    \centering
    \includegraphics[width=\linewidth]{#1}
    \vspace{1pt}
    {\scriptsize #2\par}
\end{minipage}%
}

\newcommand{\runtimegroupfive}[7]{%
\begin{figure}[H]
    \centering
    \makebox[\linewidth][c]{%
        \runtimecell{#3}{GPT2-small}
        \hfill
        \runtimecell{#4}{GPT2-medium}
        \hfill
        \runtimecell{#5}{Qwen2.5-0.5B}
    }

    \vspace{0.7em}

    \makebox[\linewidth][c]{%
        \runtimecell{#6}{Gemma3-270M}
        \hspace{0.08\linewidth}
        \runtimecell{#7}{Gemma3-1B}
    }

    \caption{Runtime testing loss curves on #1 with Seq 128.}
    \label{#2}
\end{figure}
}

\newcommand{\runtimegroupfour}[6]{%
\begin{figure}[H]
    \centering
    \makebox[\linewidth][c]{%
        \runtimecell{#3}{GPT2-small}
        \hspace{0.08\linewidth}
        \runtimecell{#4}{GPT2-medium}
    }

    \vspace{0.7em}

    \makebox[\linewidth][c]{%
        \runtimecell{#5}{Qwen2.5-0.5B}
        \hspace{0.08\linewidth}
        \runtimecell{#6}{Gemma3-270M}
    }

    \caption{Runtime testing loss curves on #1 with Seq 256.}
    \label{#2}
\end{figure}
}

\section{Runtime Testing Loss Curves}
\label{appC}
\vspace{-4mm}

\runtimegroupfive
{ARC-Challenge}
{fig:runtime-loss-arcc-seq128}
{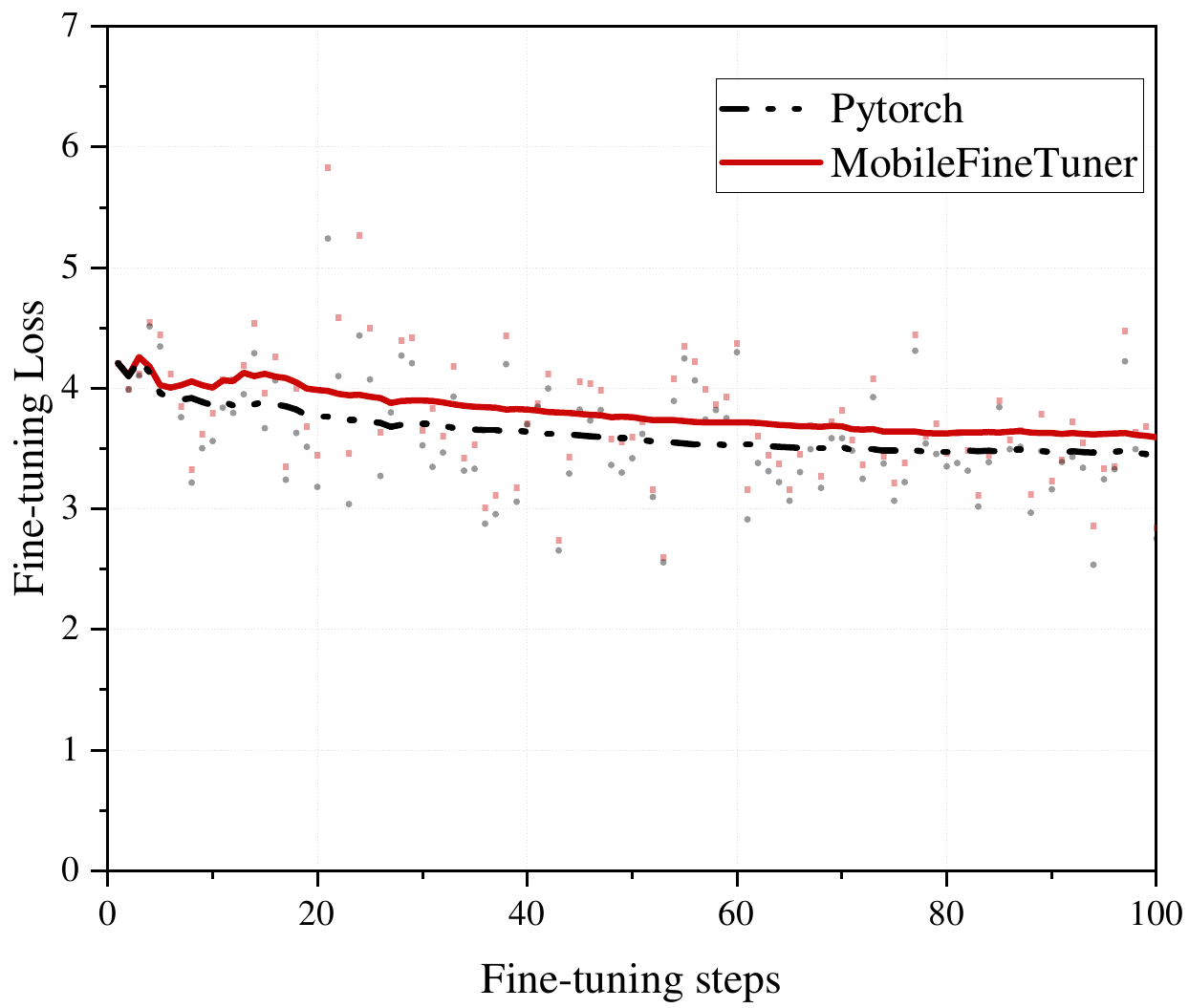}
{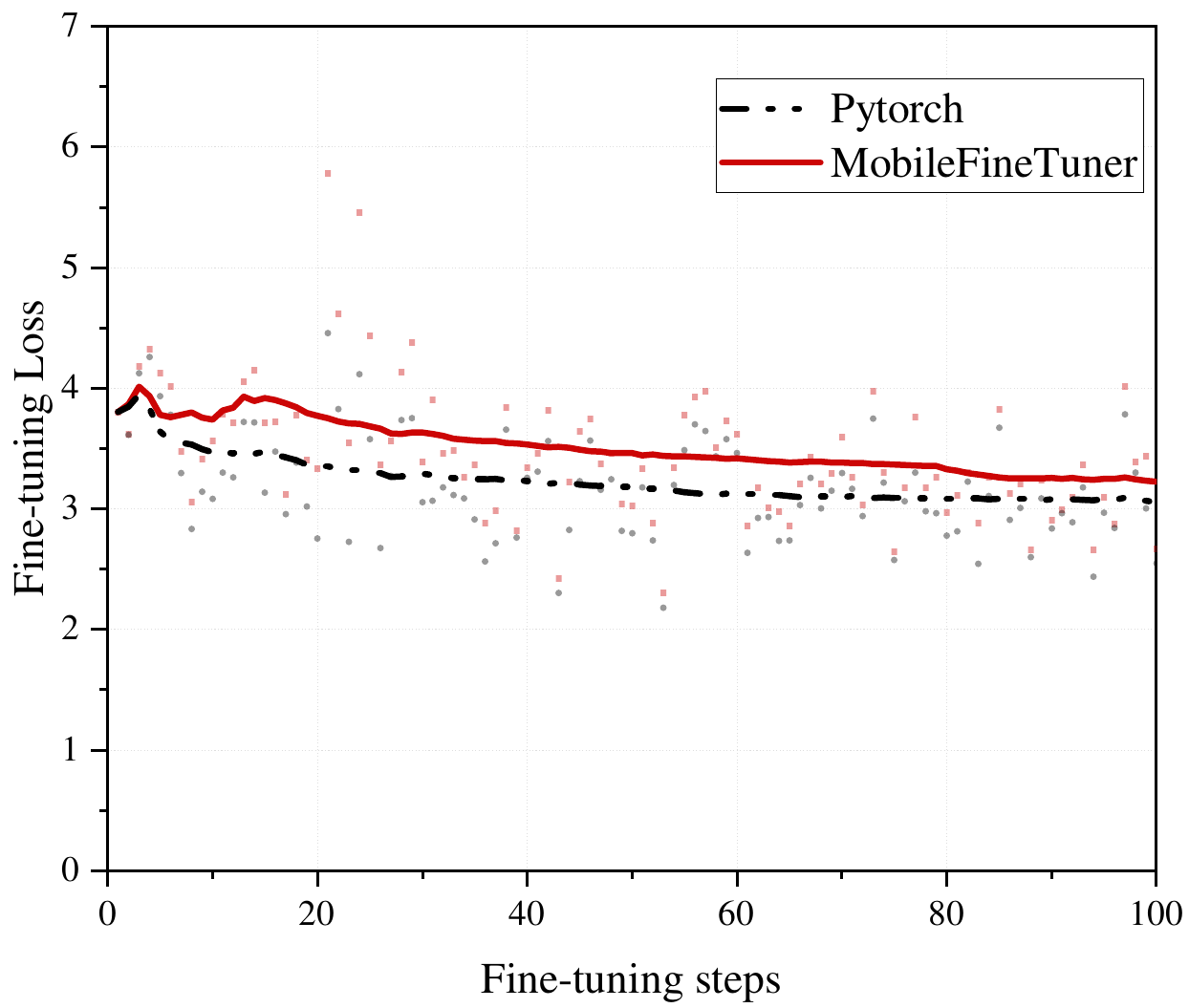}
{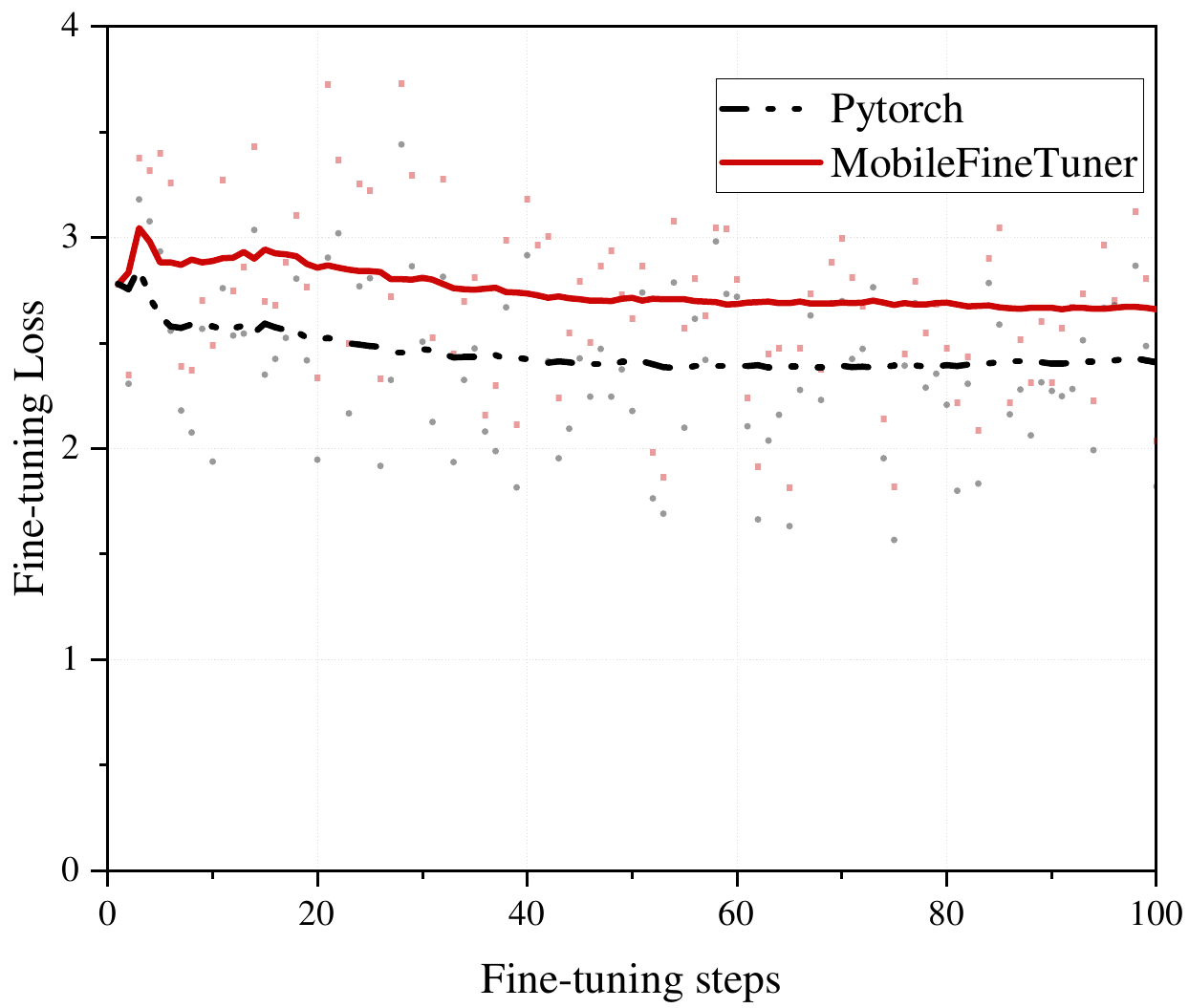}
{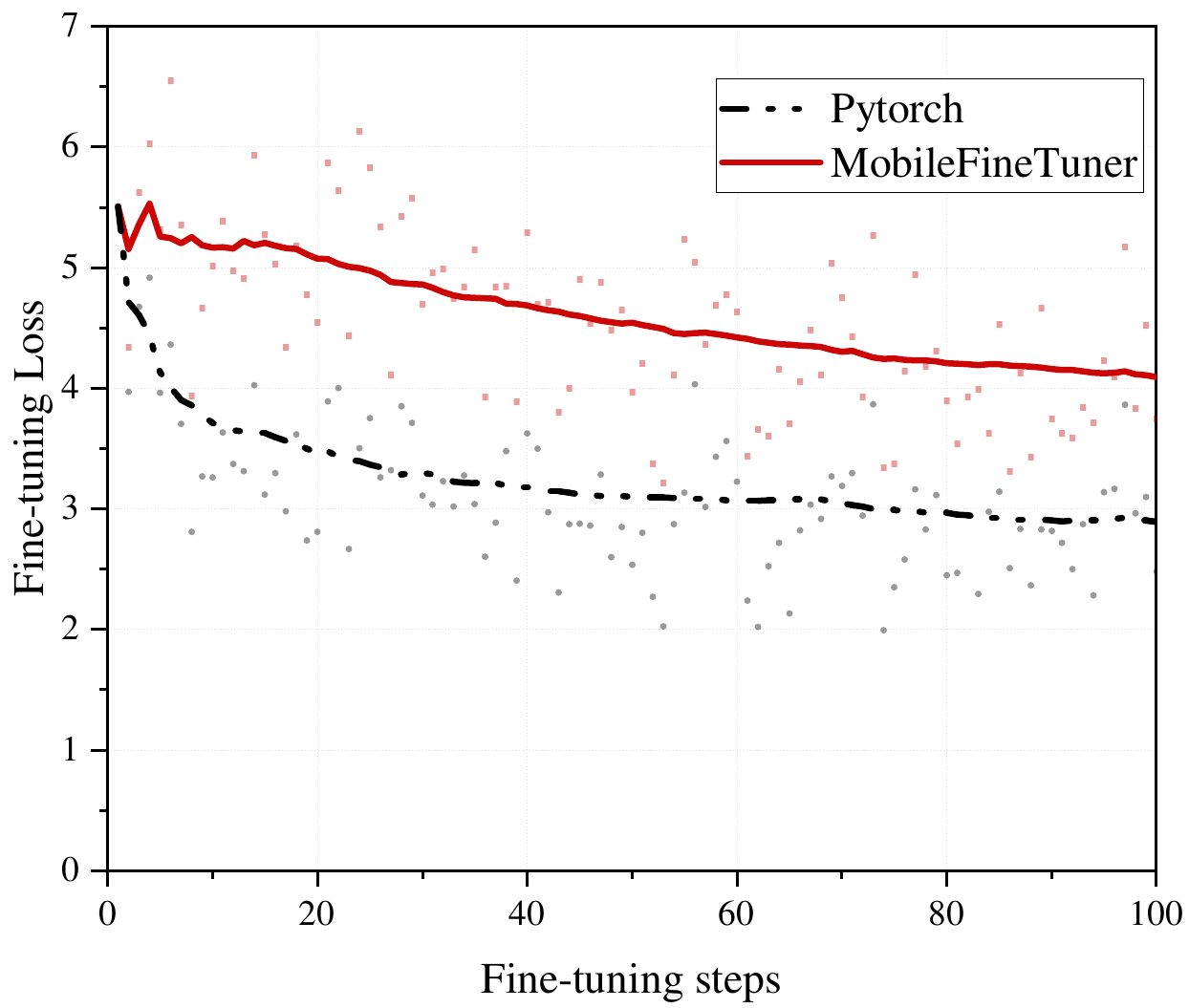}
{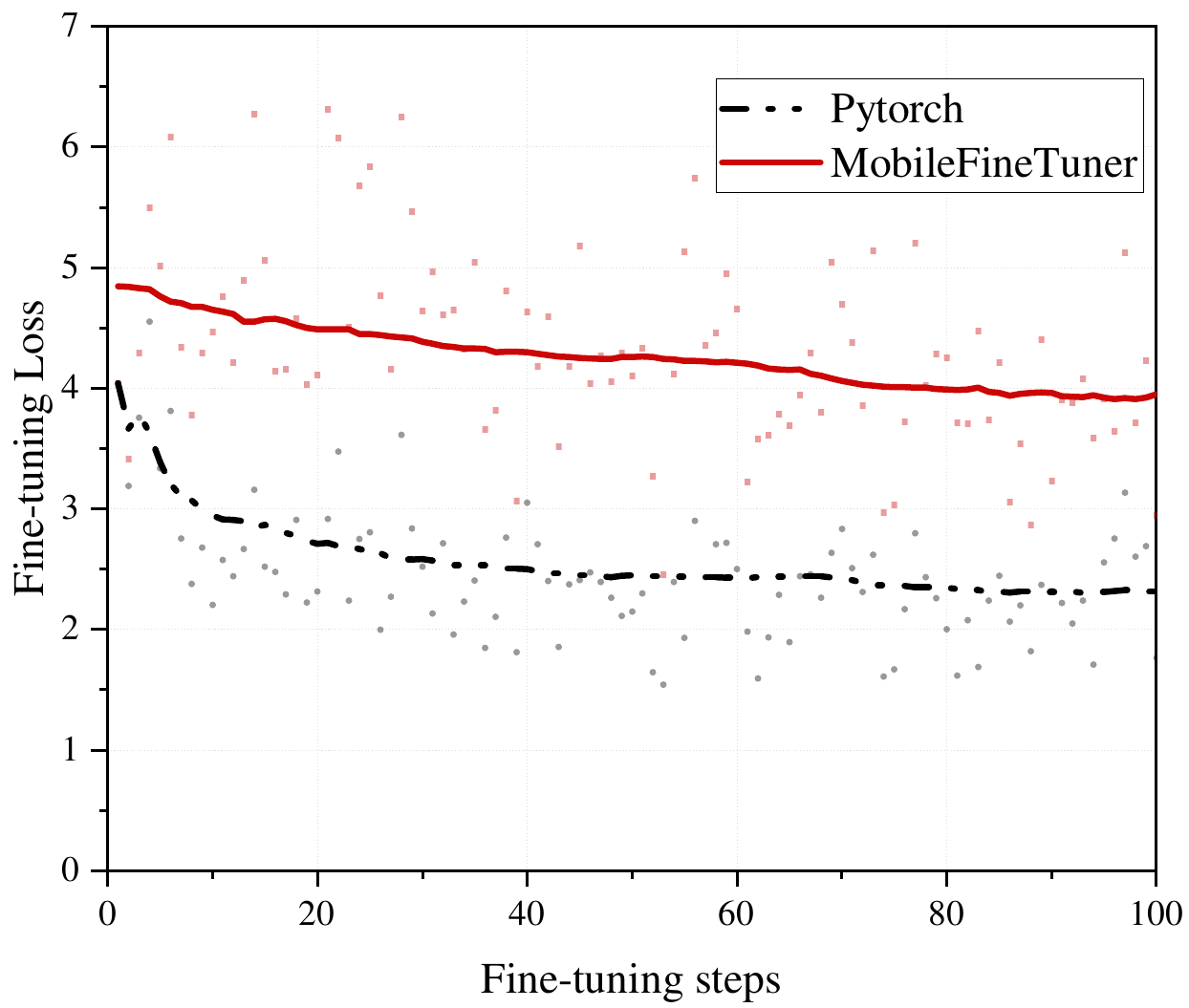}

\runtimegroupfour
{ARC-Challenge}
{fig:runtime-loss-arcc-seq256}
{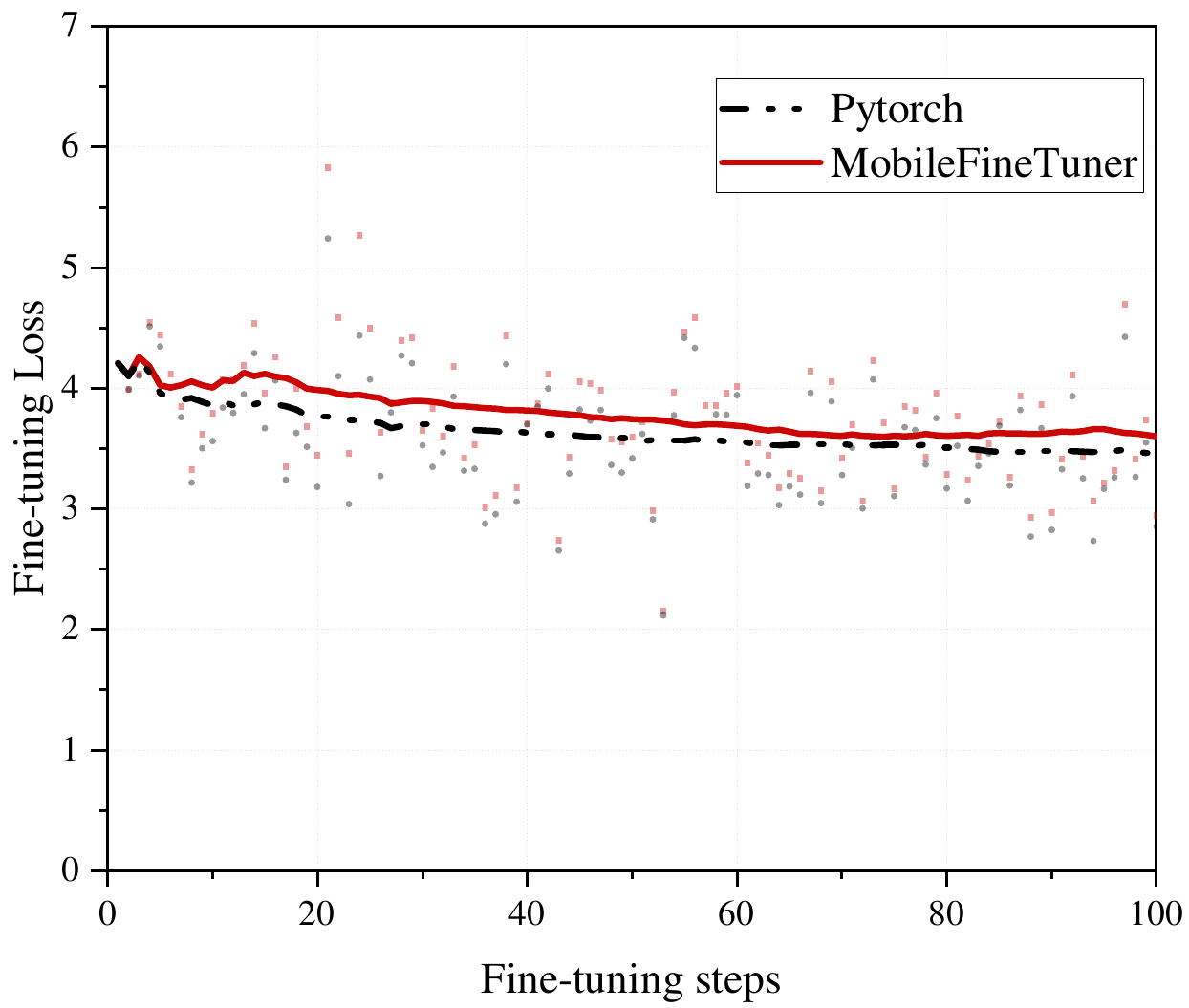}
{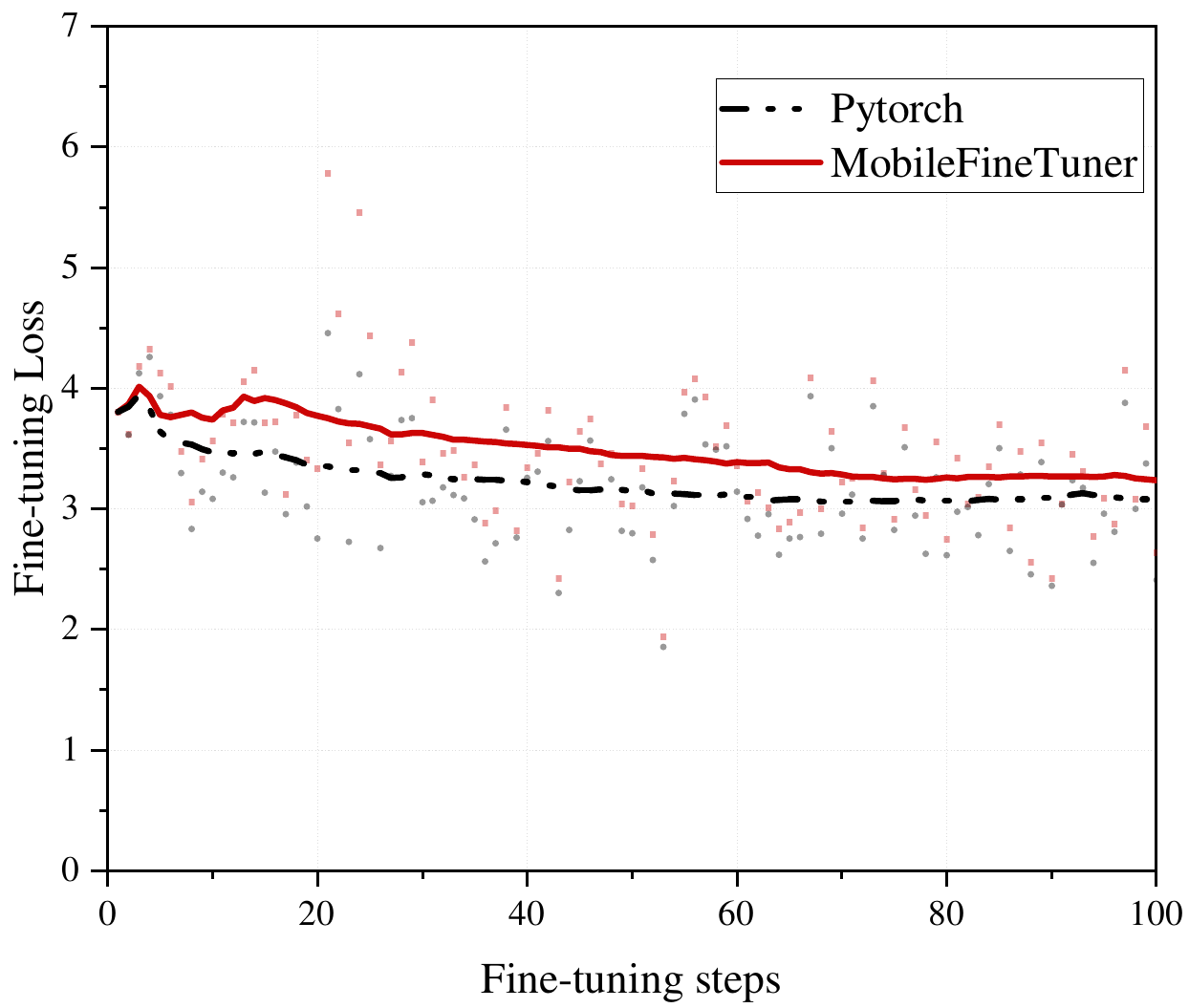}
{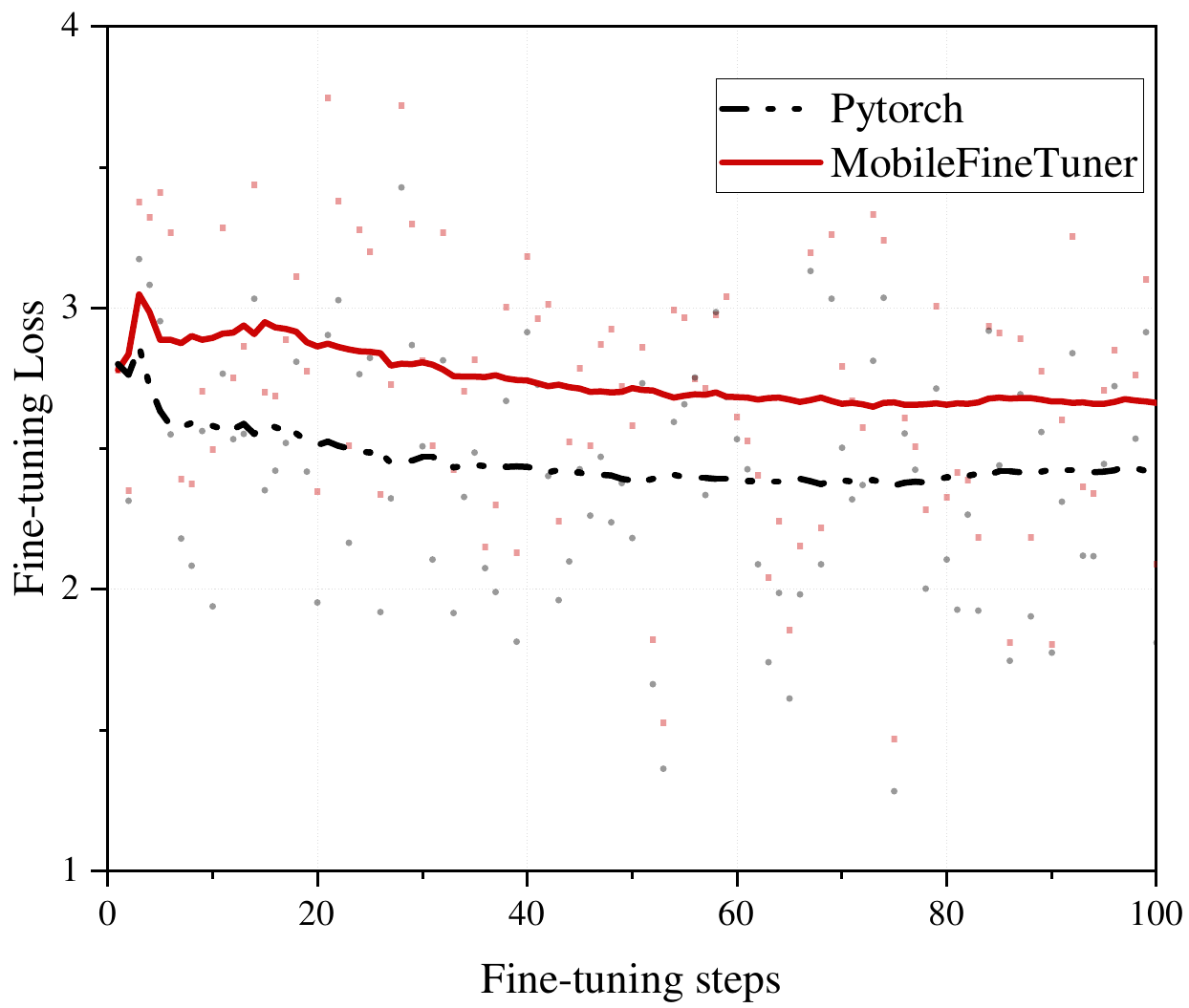}
{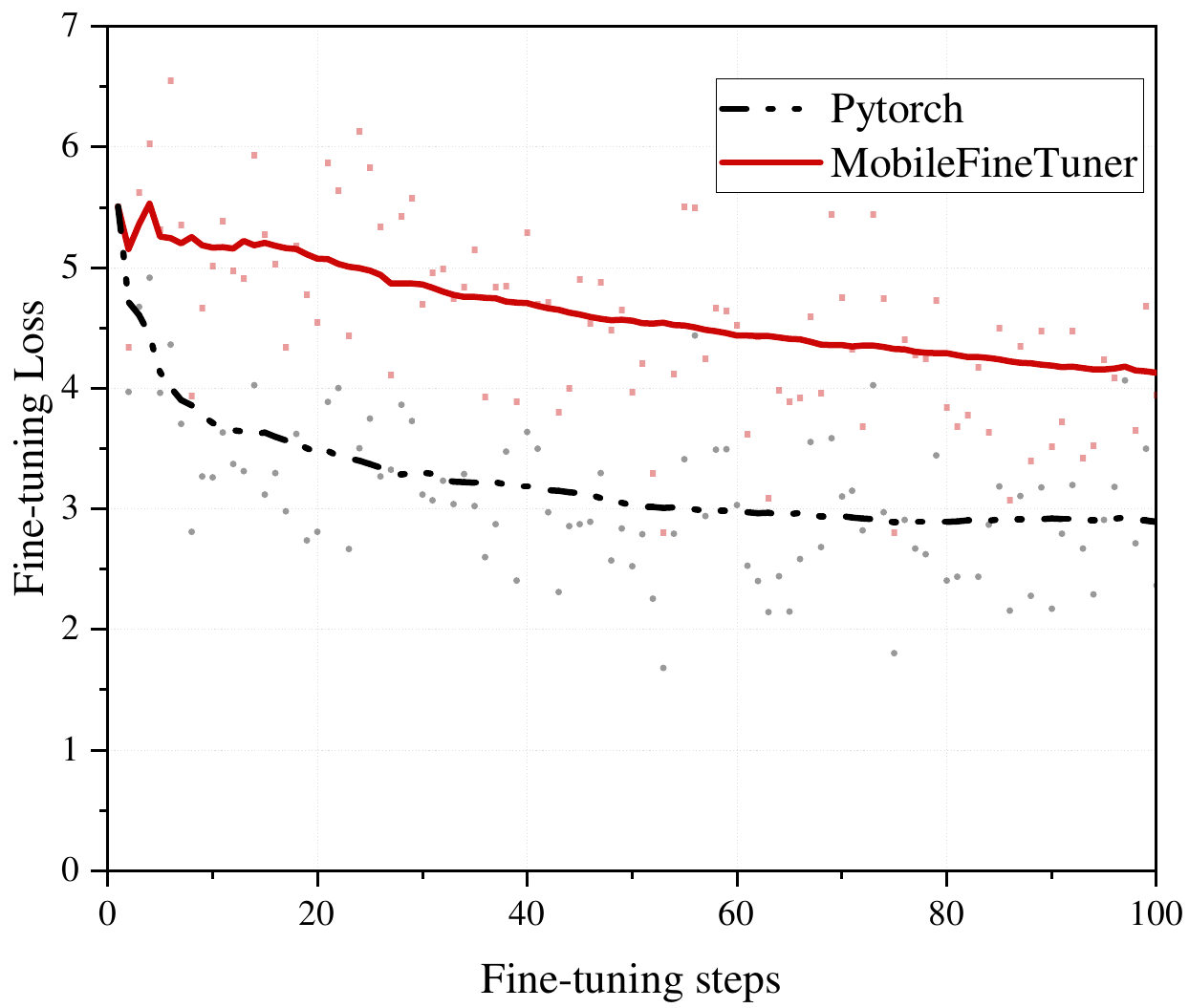}

\runtimegroupfive
{ARC-Easy}
{fig:runtime-loss-arce-seq128}
{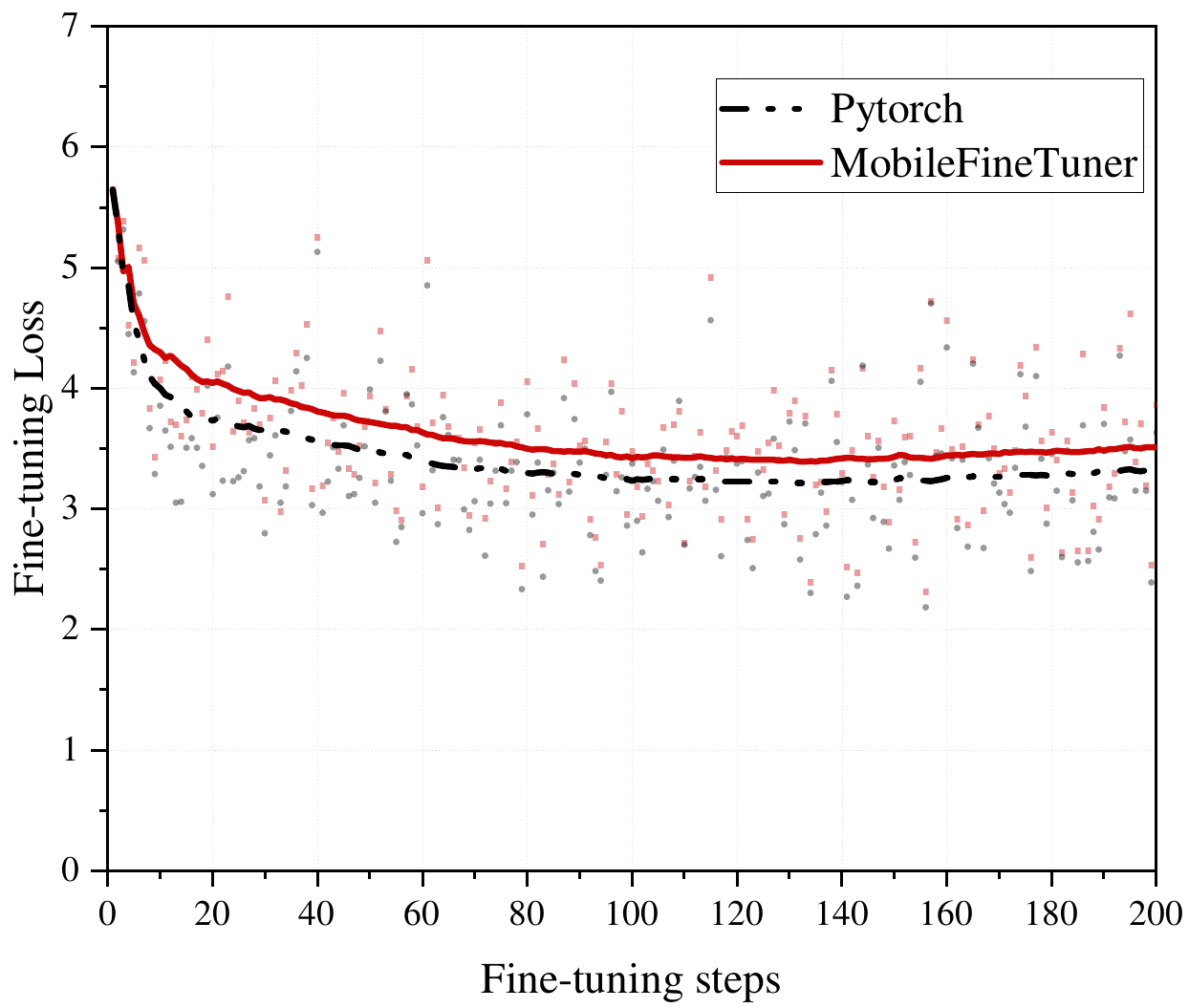}
{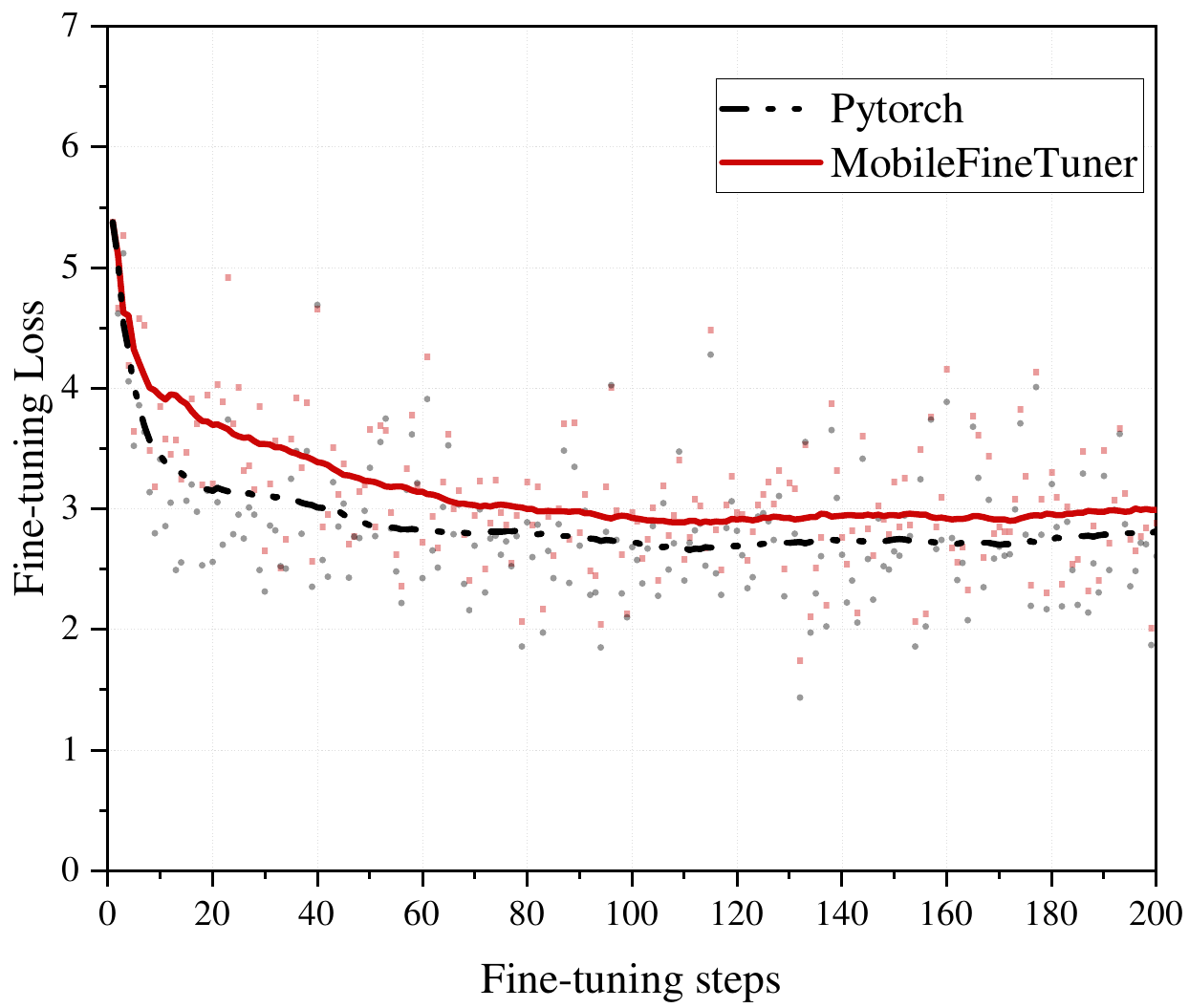}
{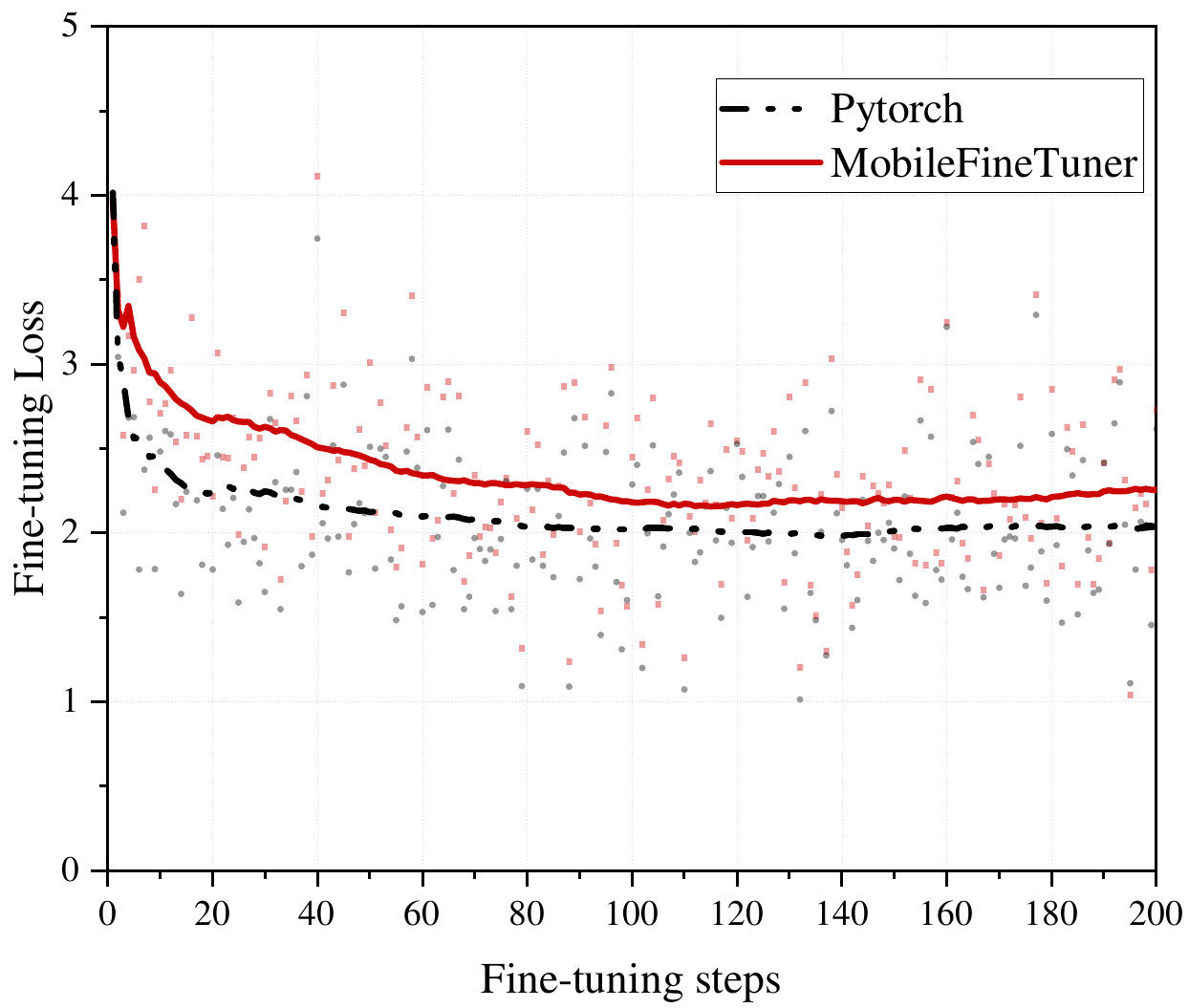}
{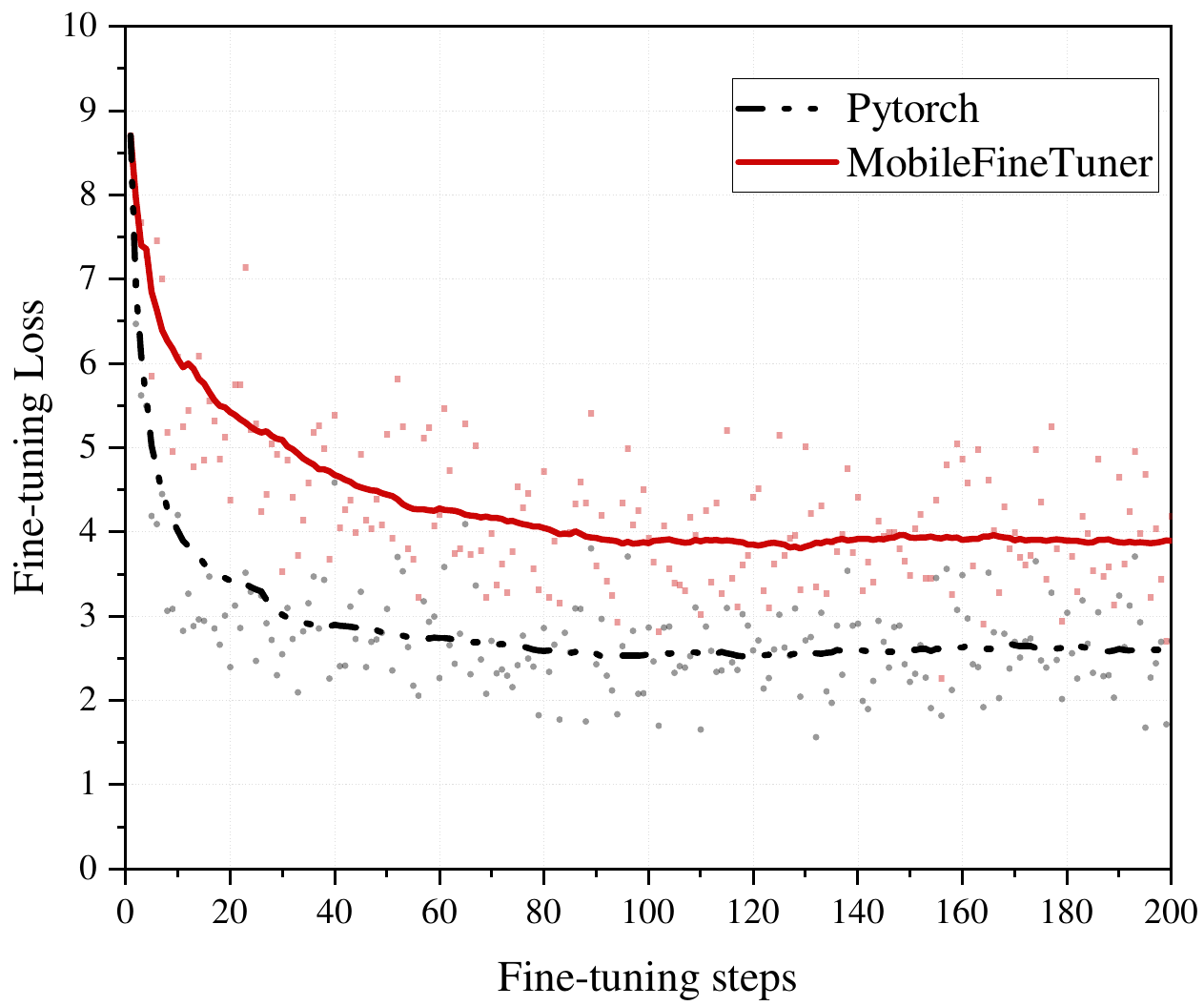}
{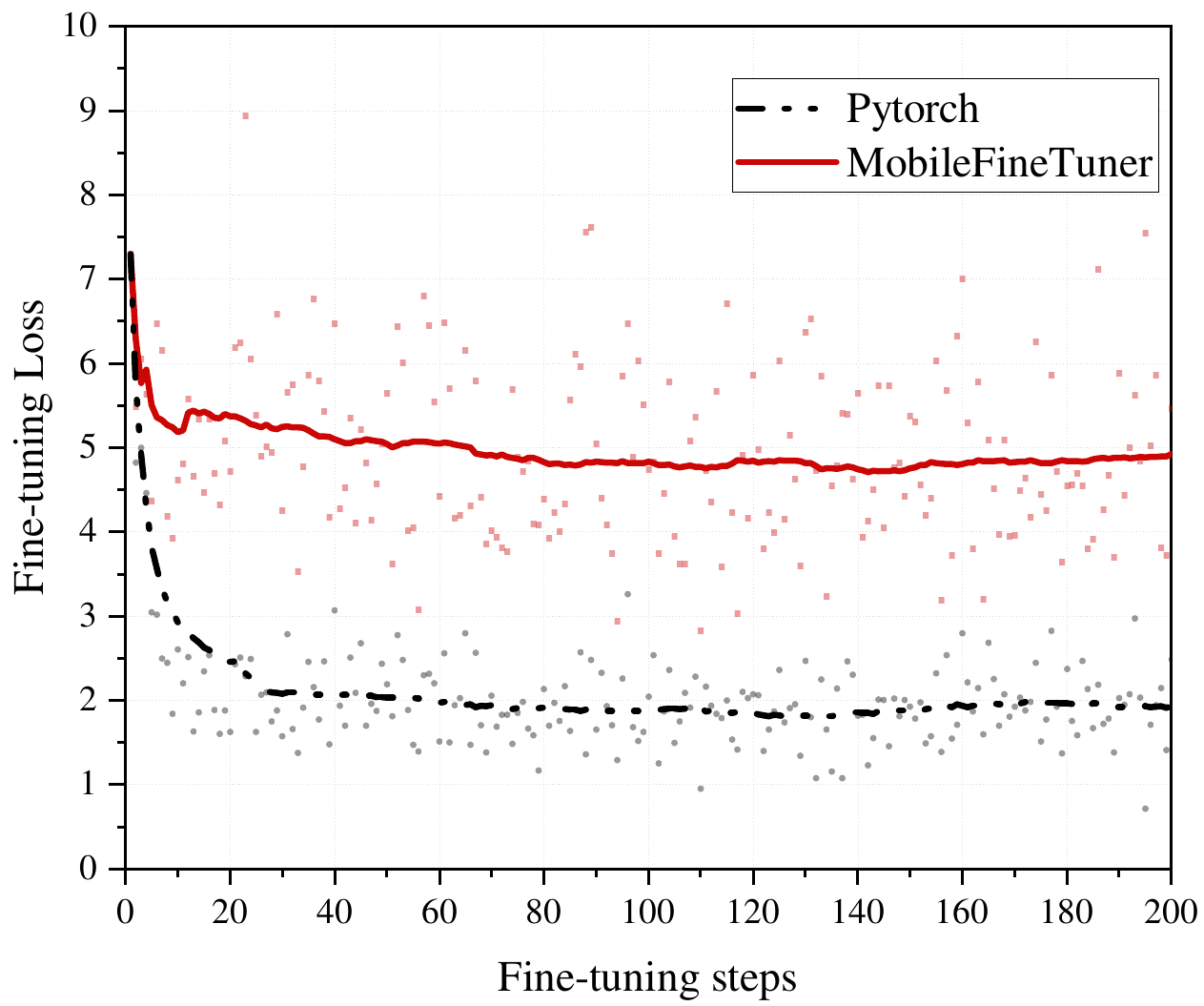}

\runtimegroupfour
{ARC-Easy}
{fig:runtime-loss-arce-seq256}
{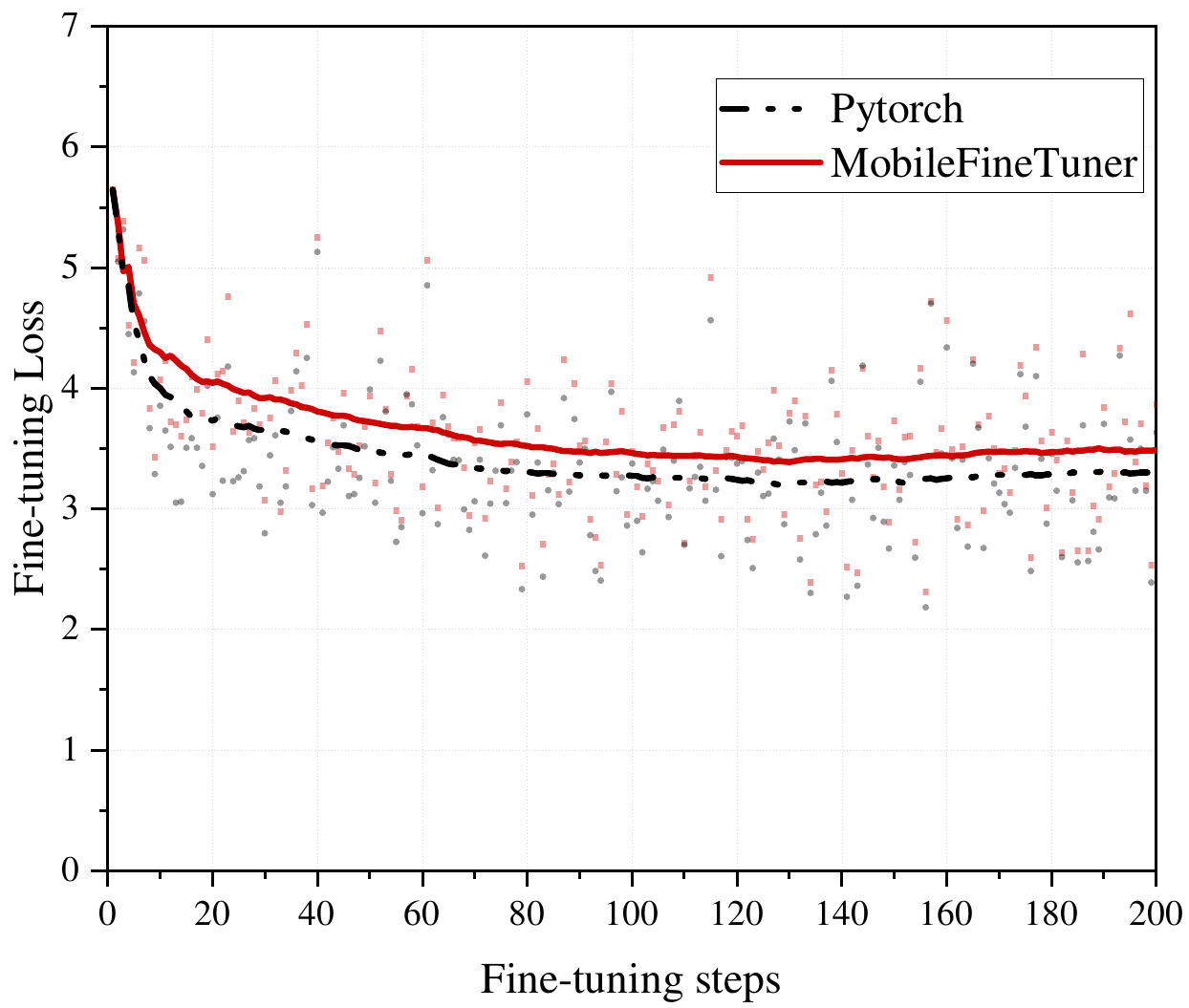}
{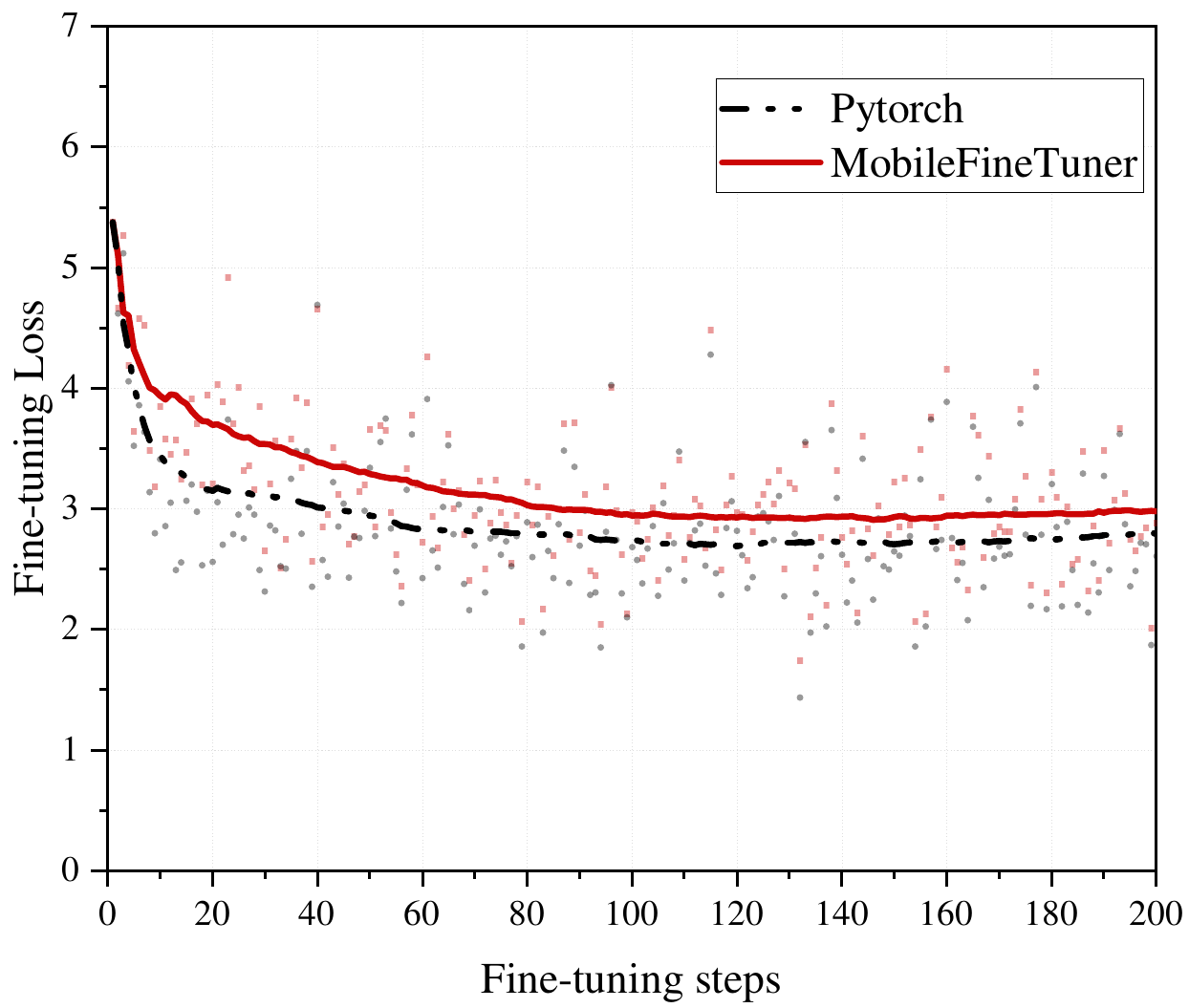}
{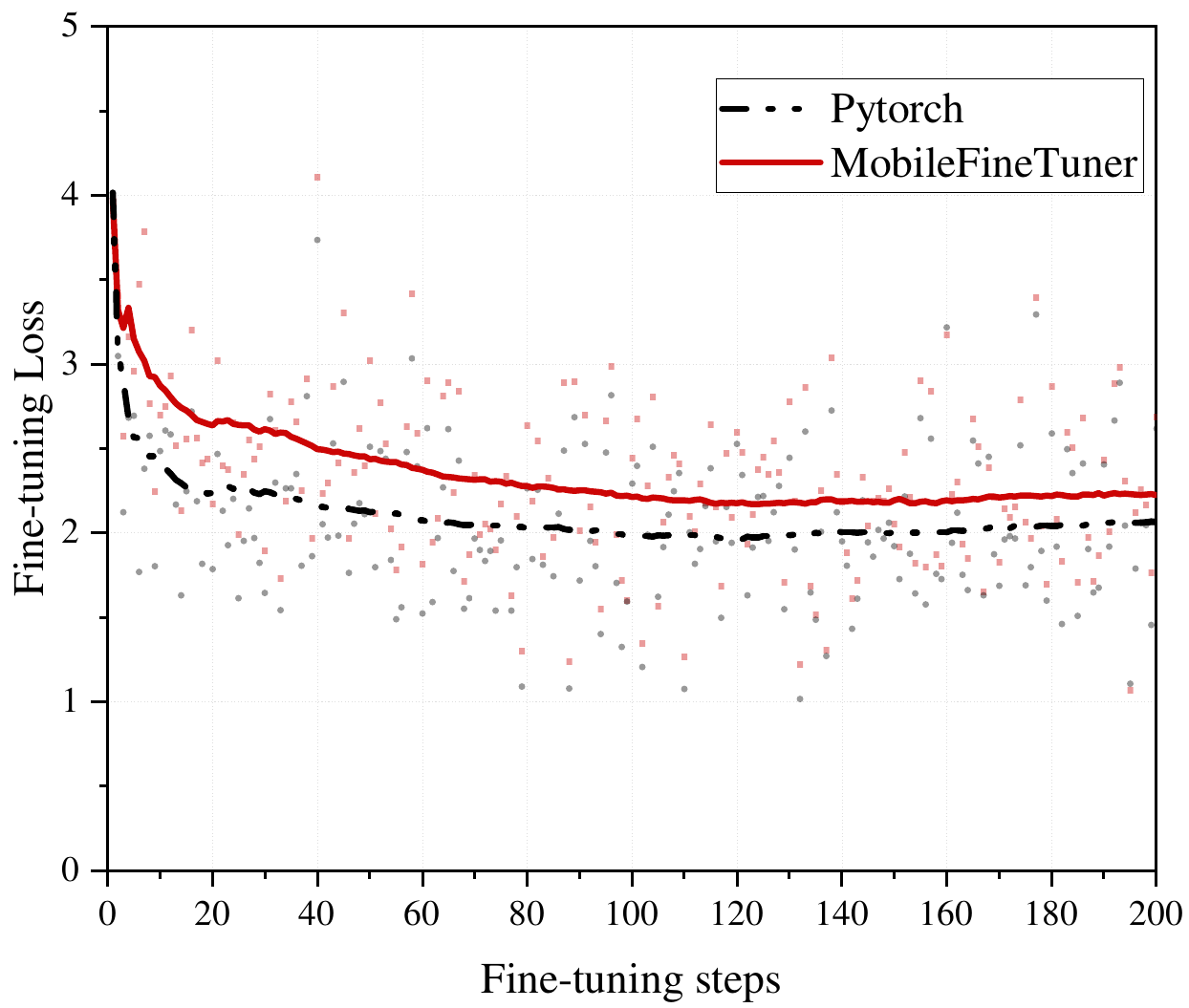}
{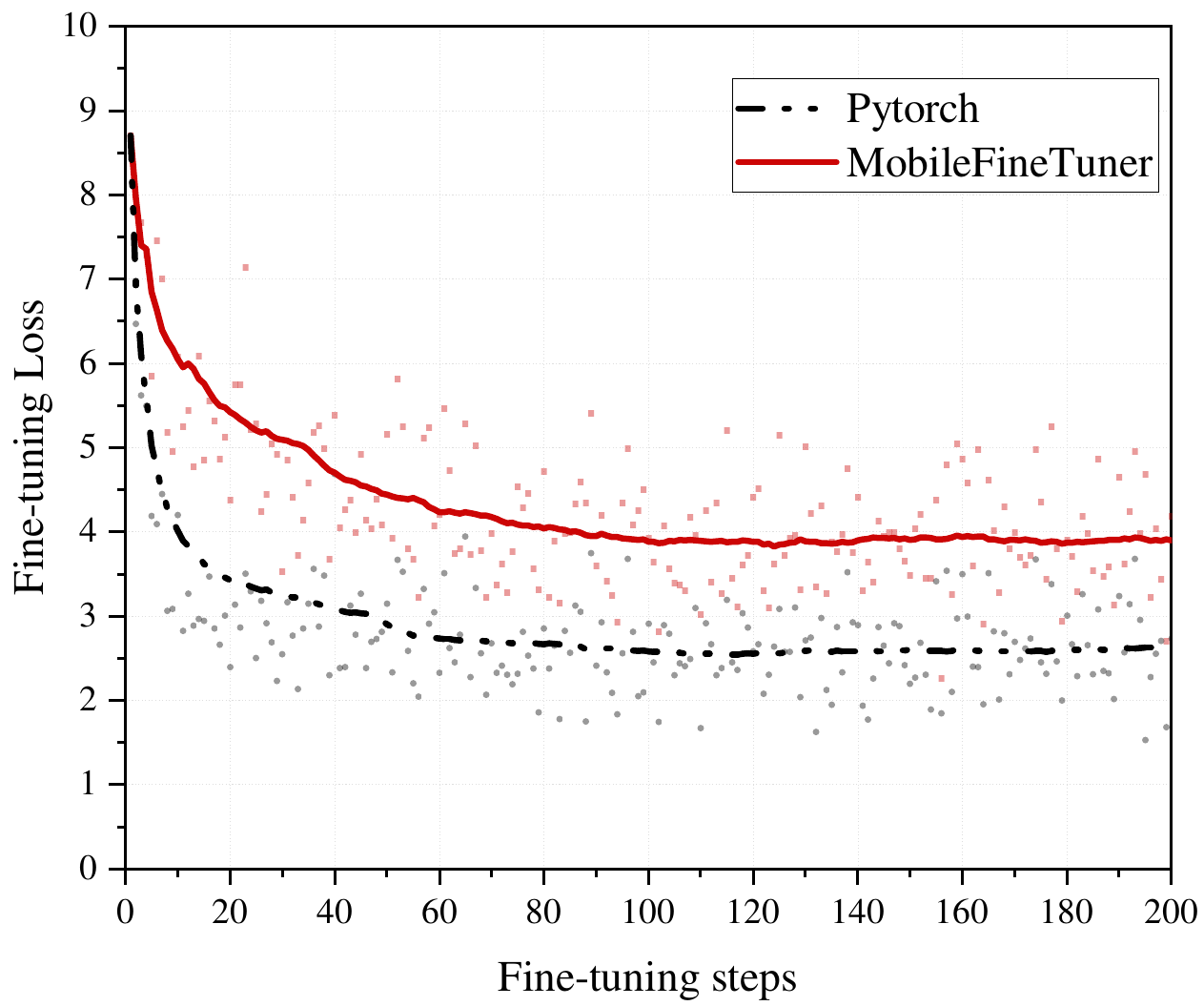}

\runtimegroupfive
{HellaSwag}
{fig:runtime-loss-hellaswag-seq128}
{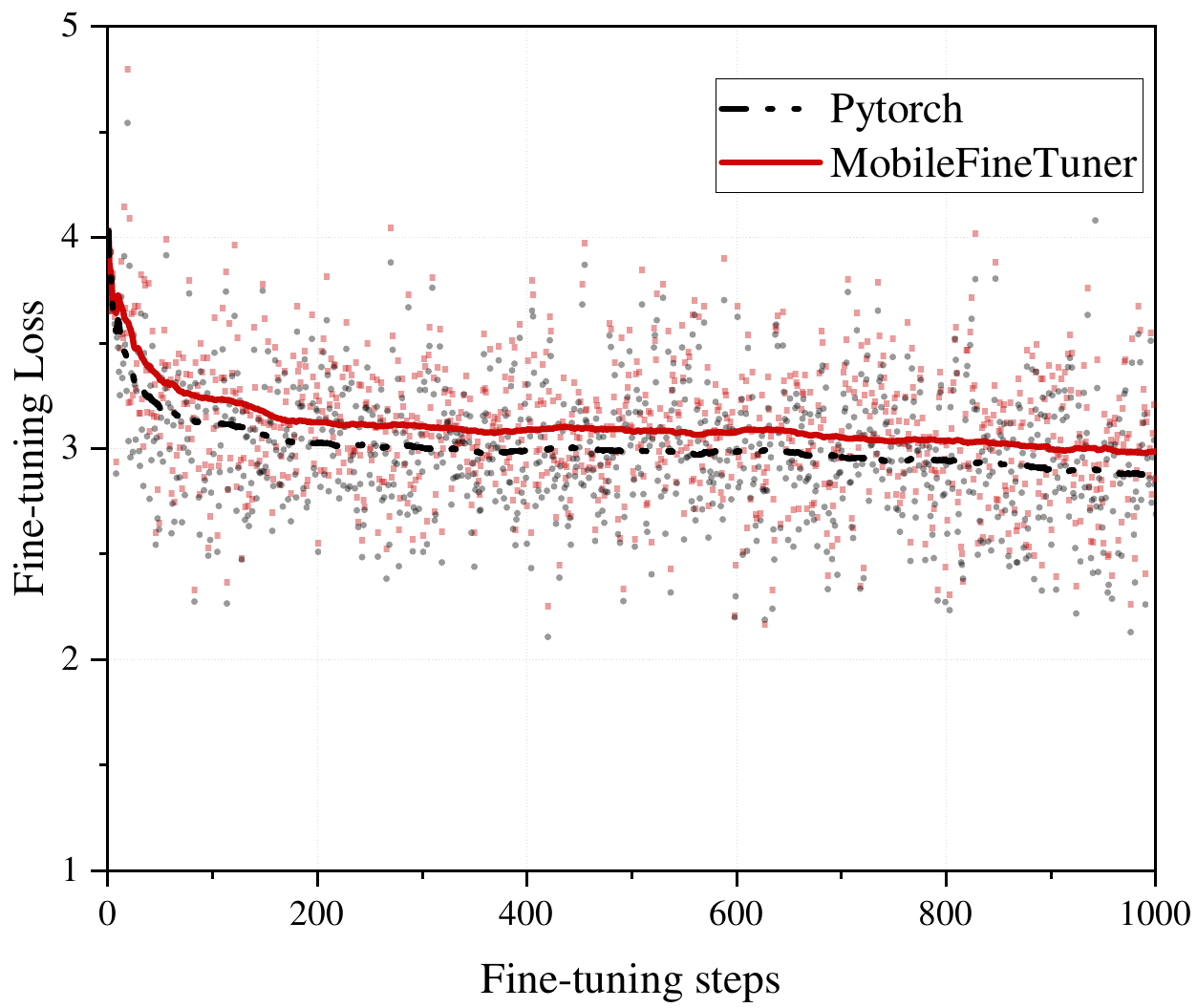}
{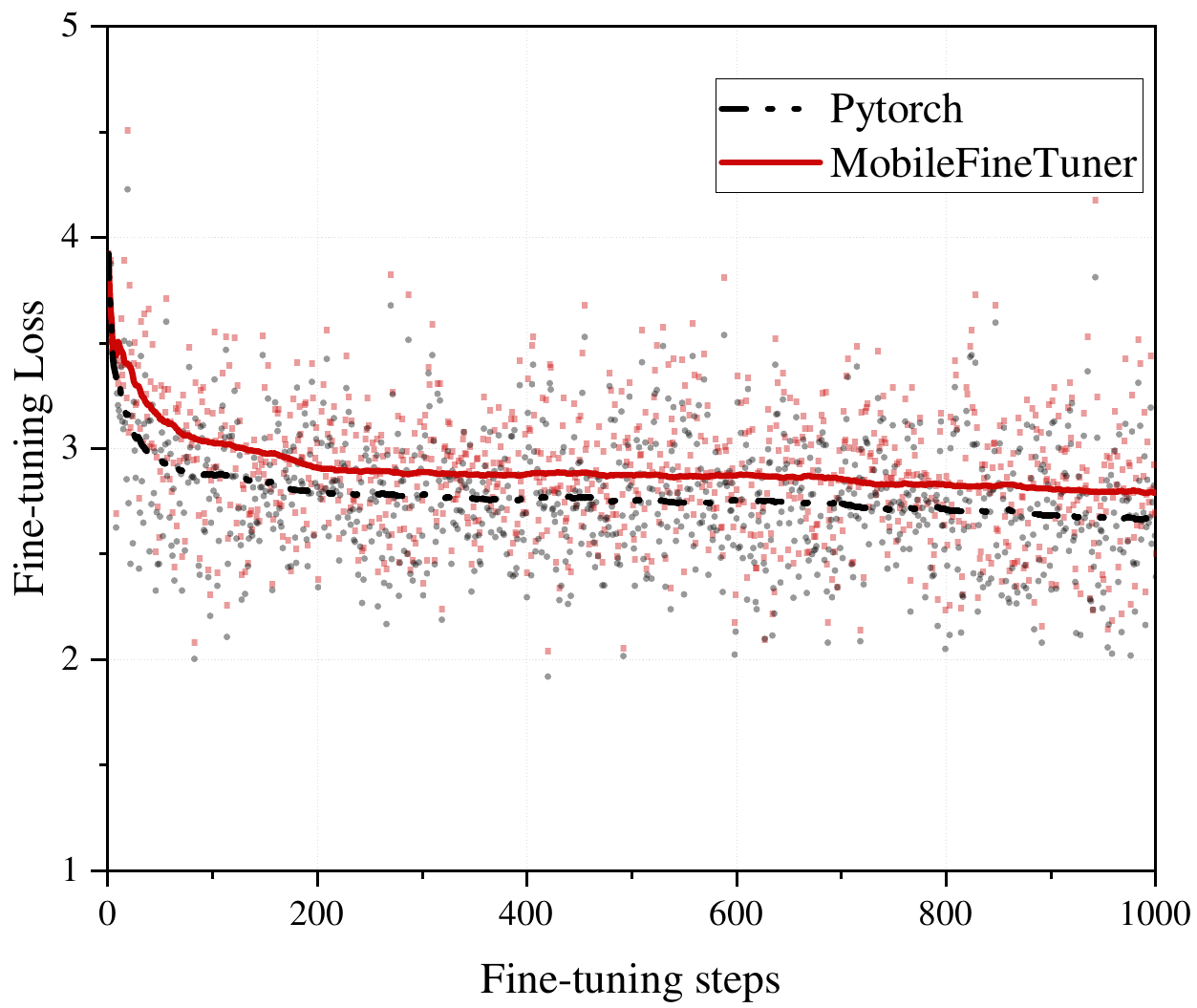}
{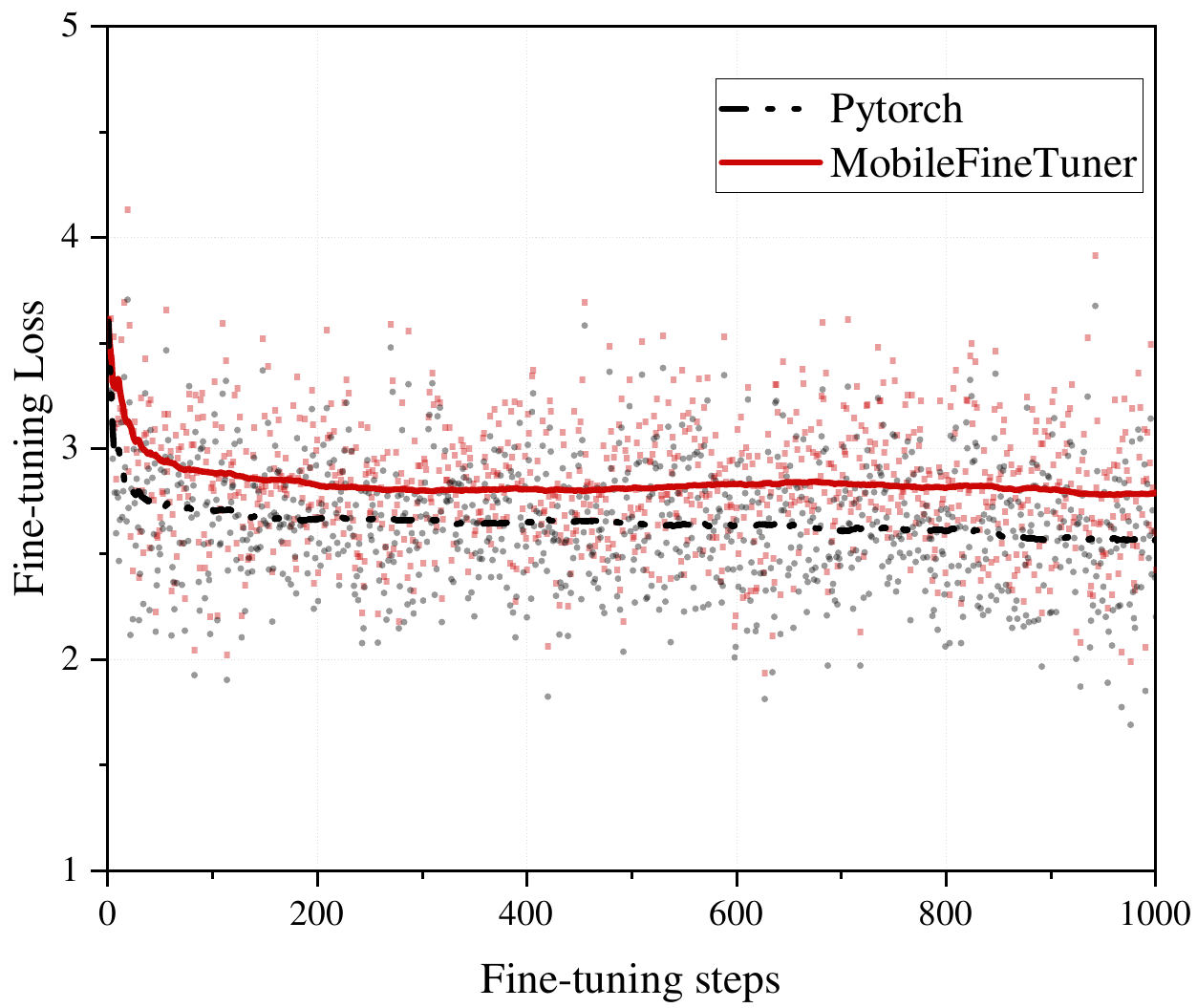}
{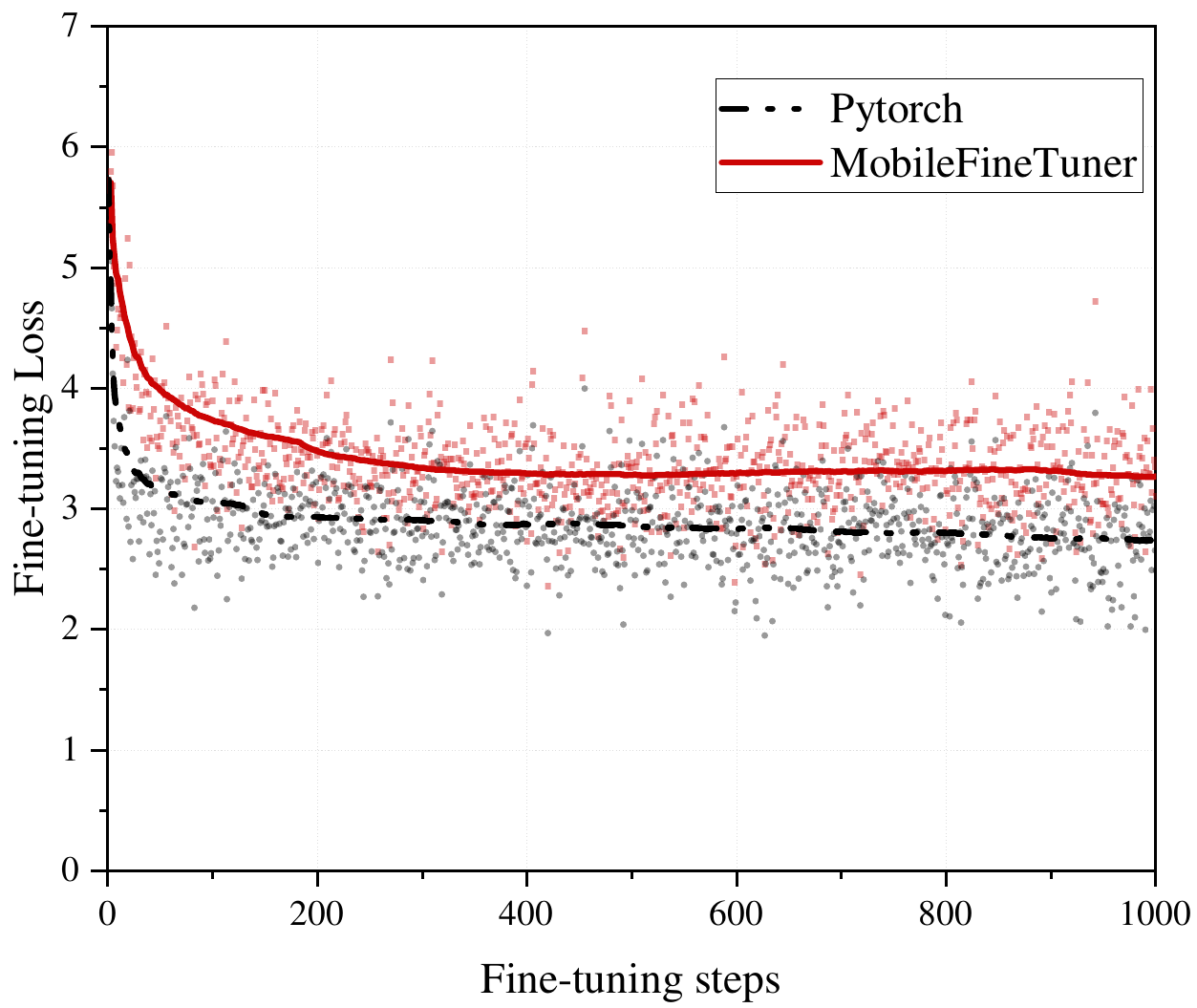}
{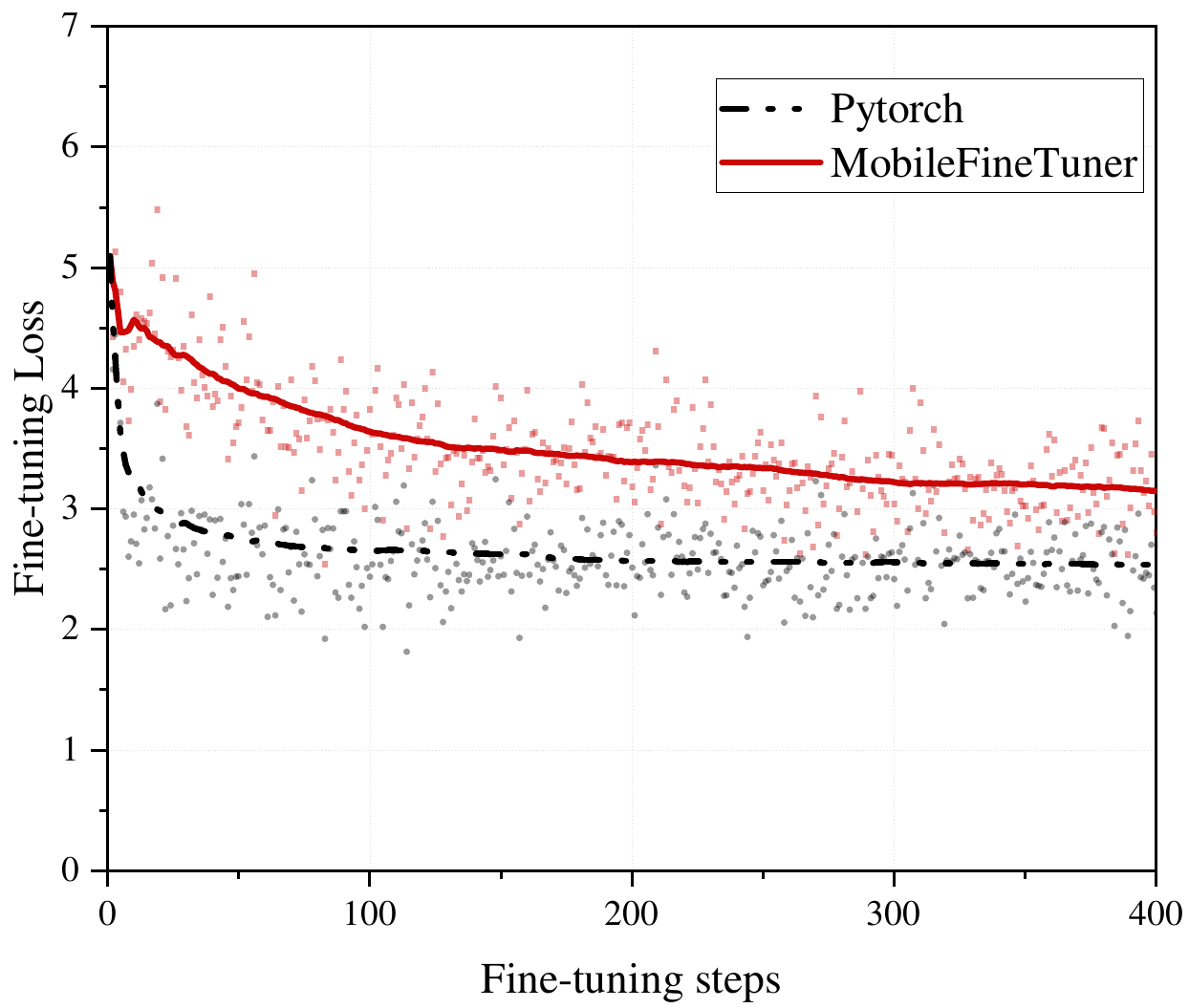}

\runtimegroupfour
{HellaSwag}
{fig:runtime-loss-hellaswag-seq256}
{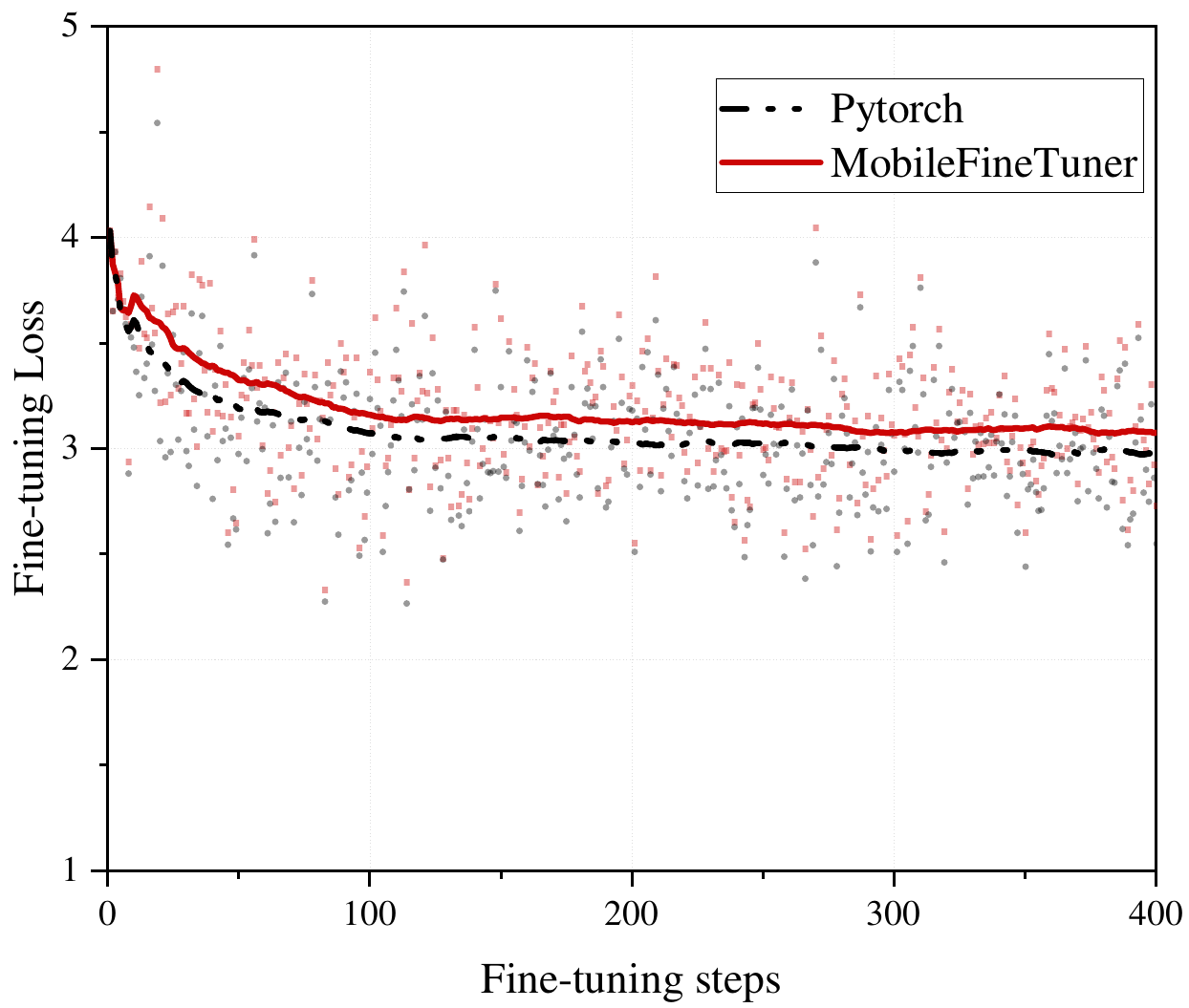}
{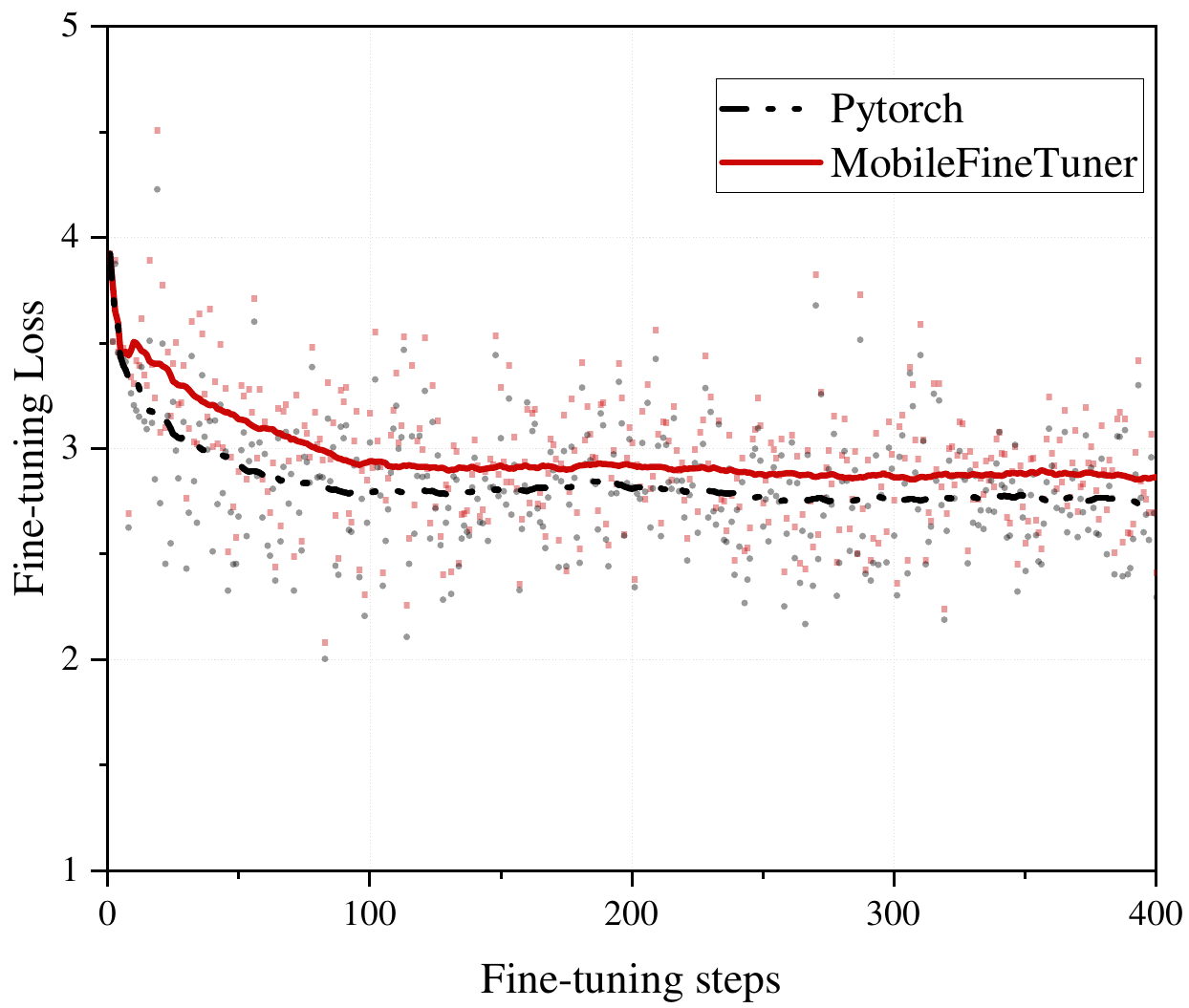}
{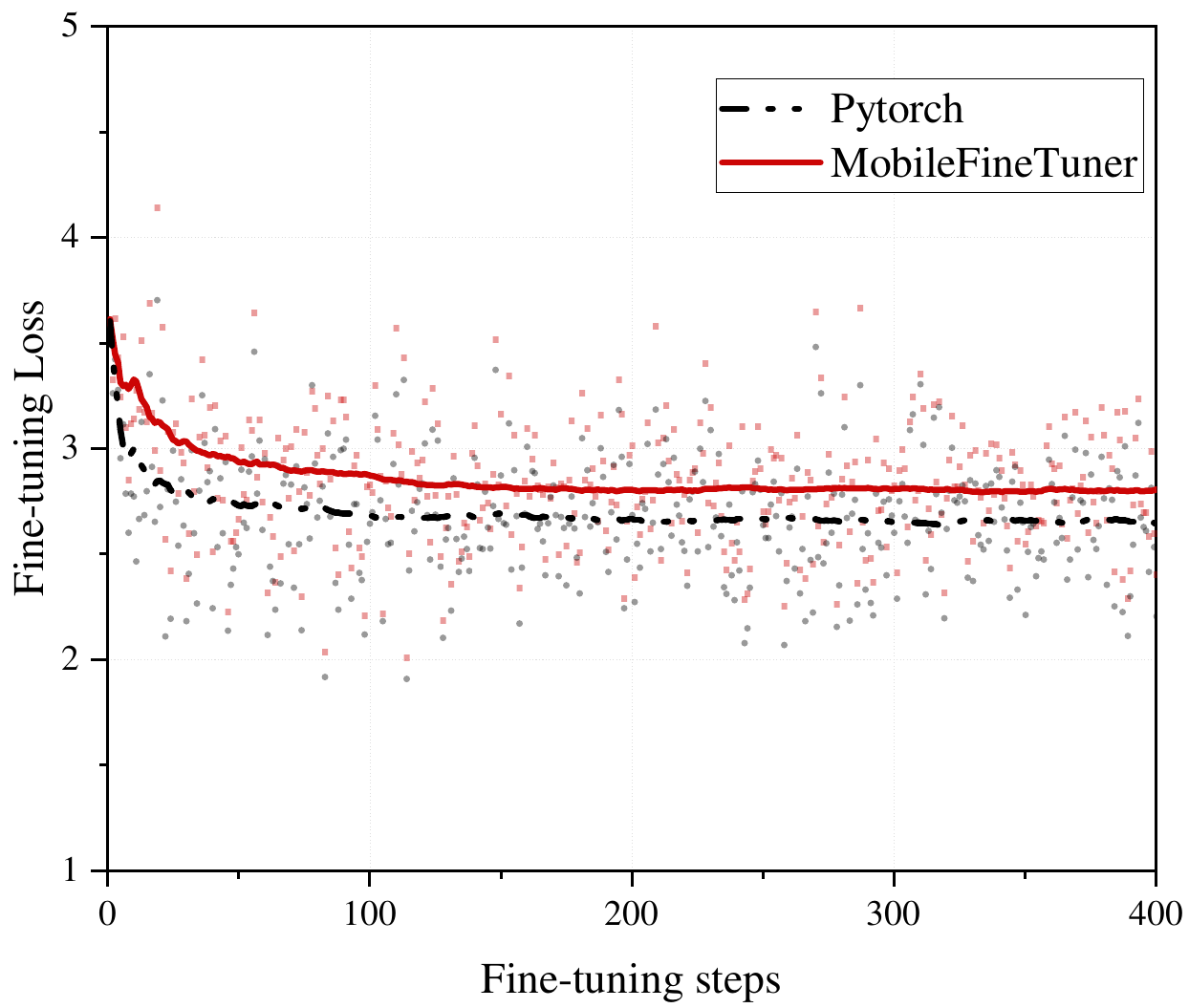}
{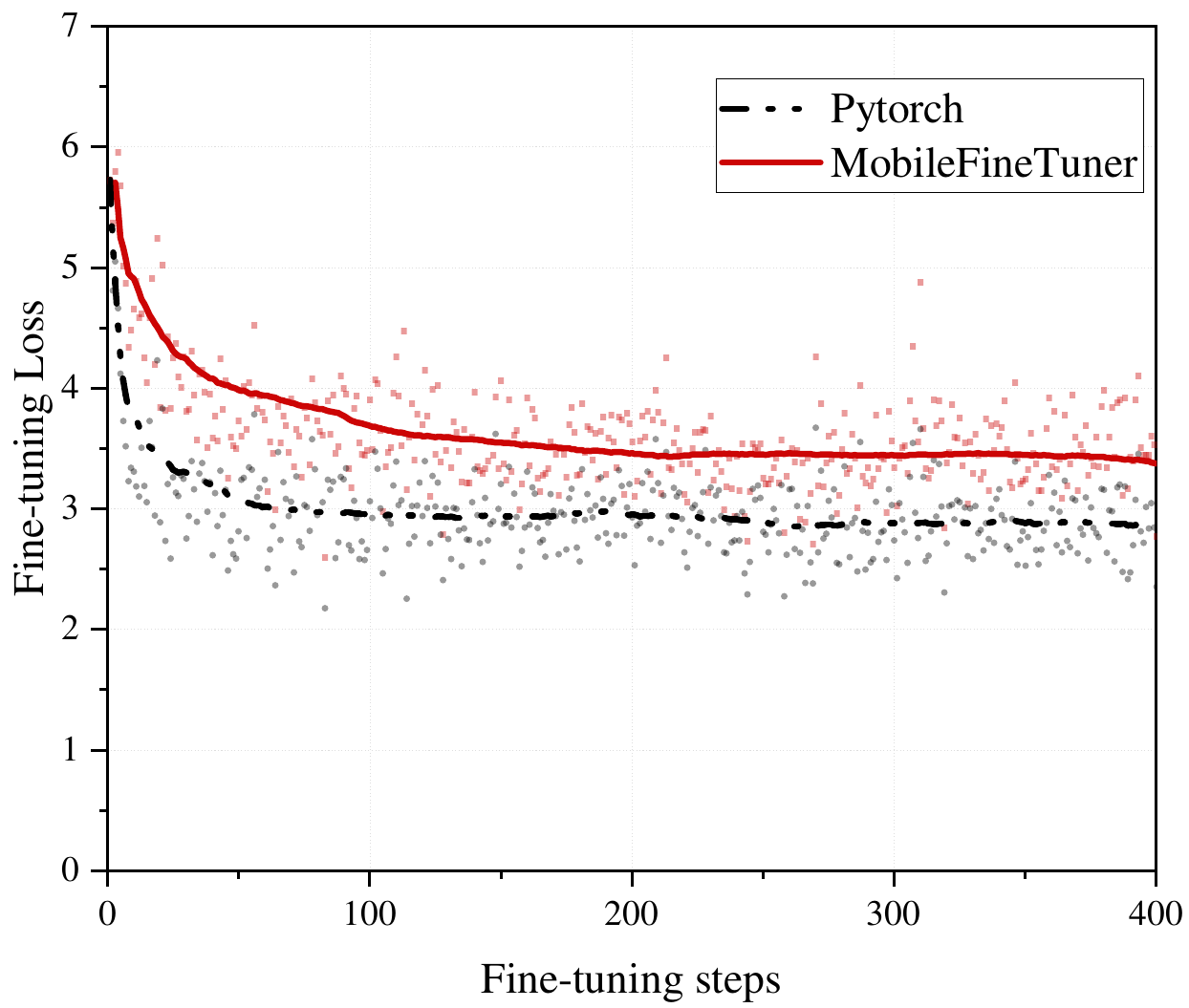}

\runtimegroupfive
{MMLU}
{fig:runtime-loss-mmlu-seq128}
{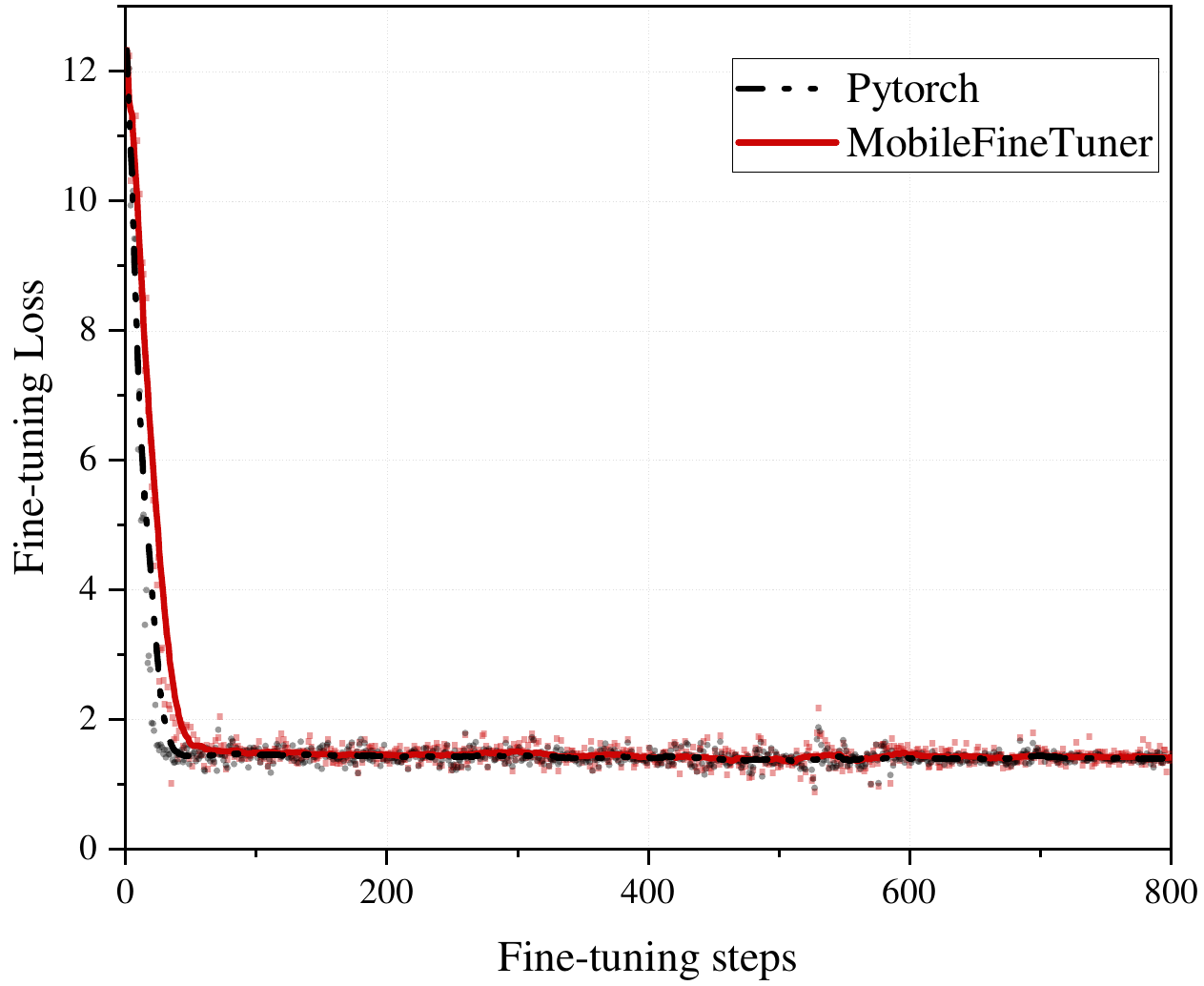}
{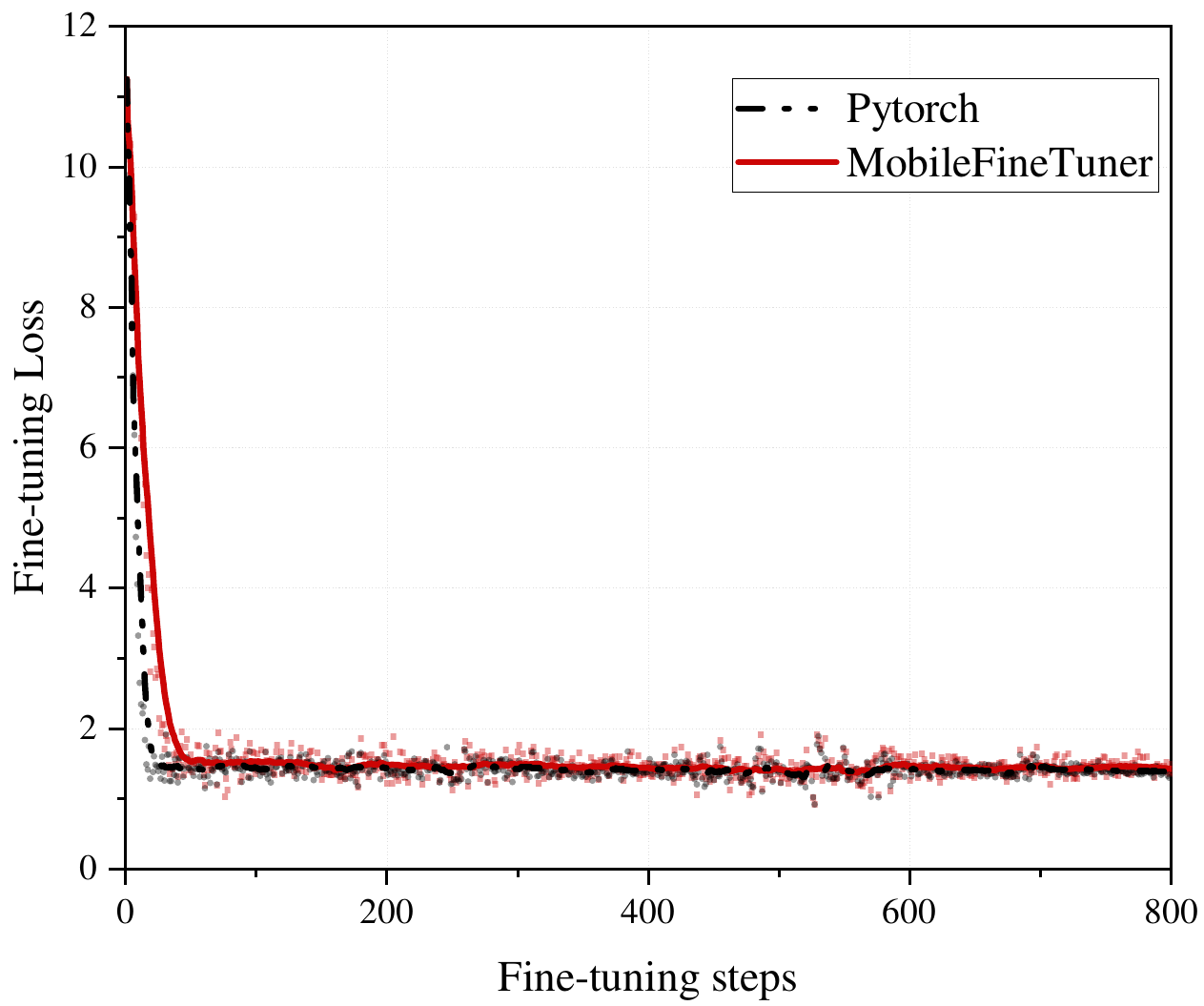}
{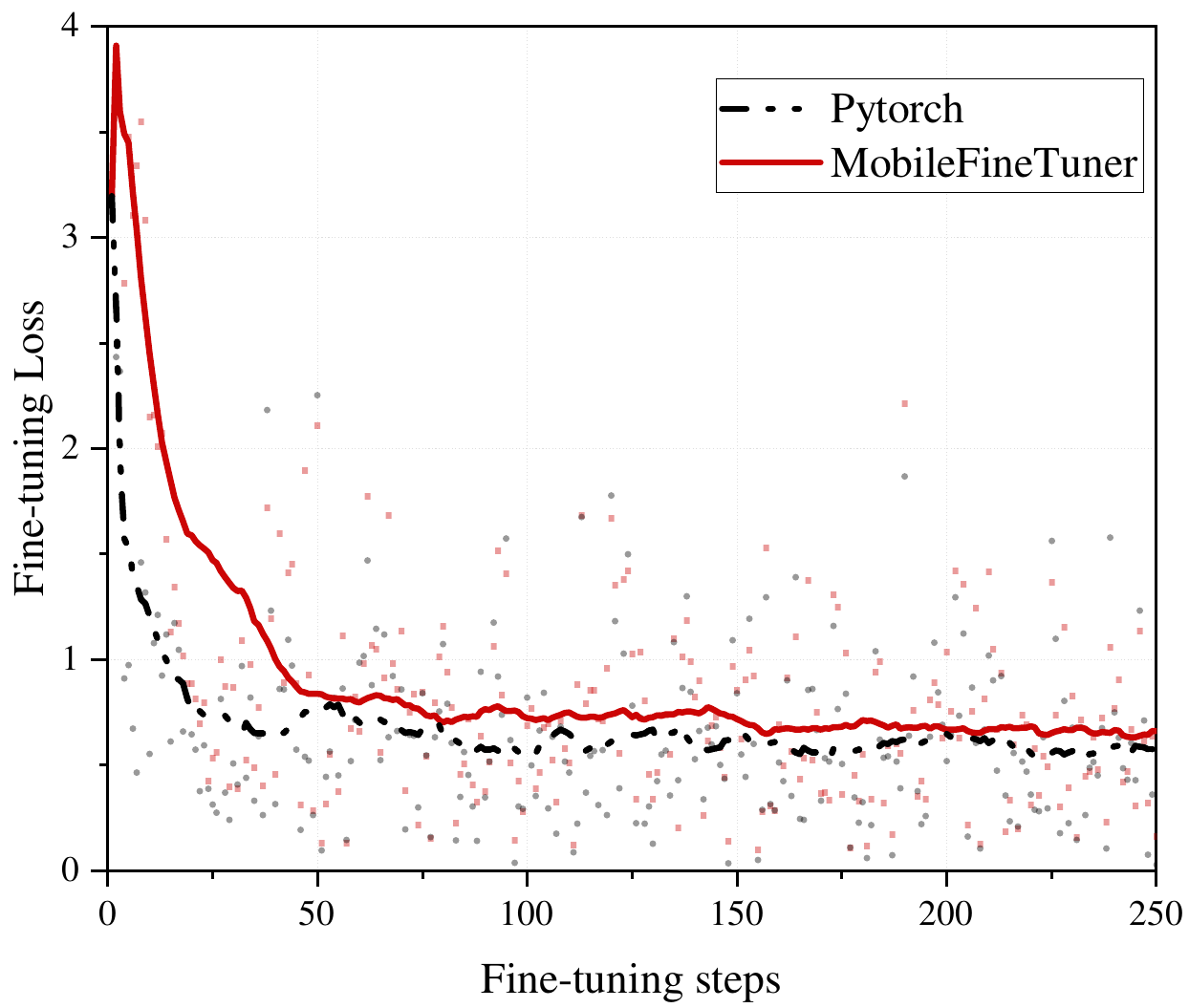}
{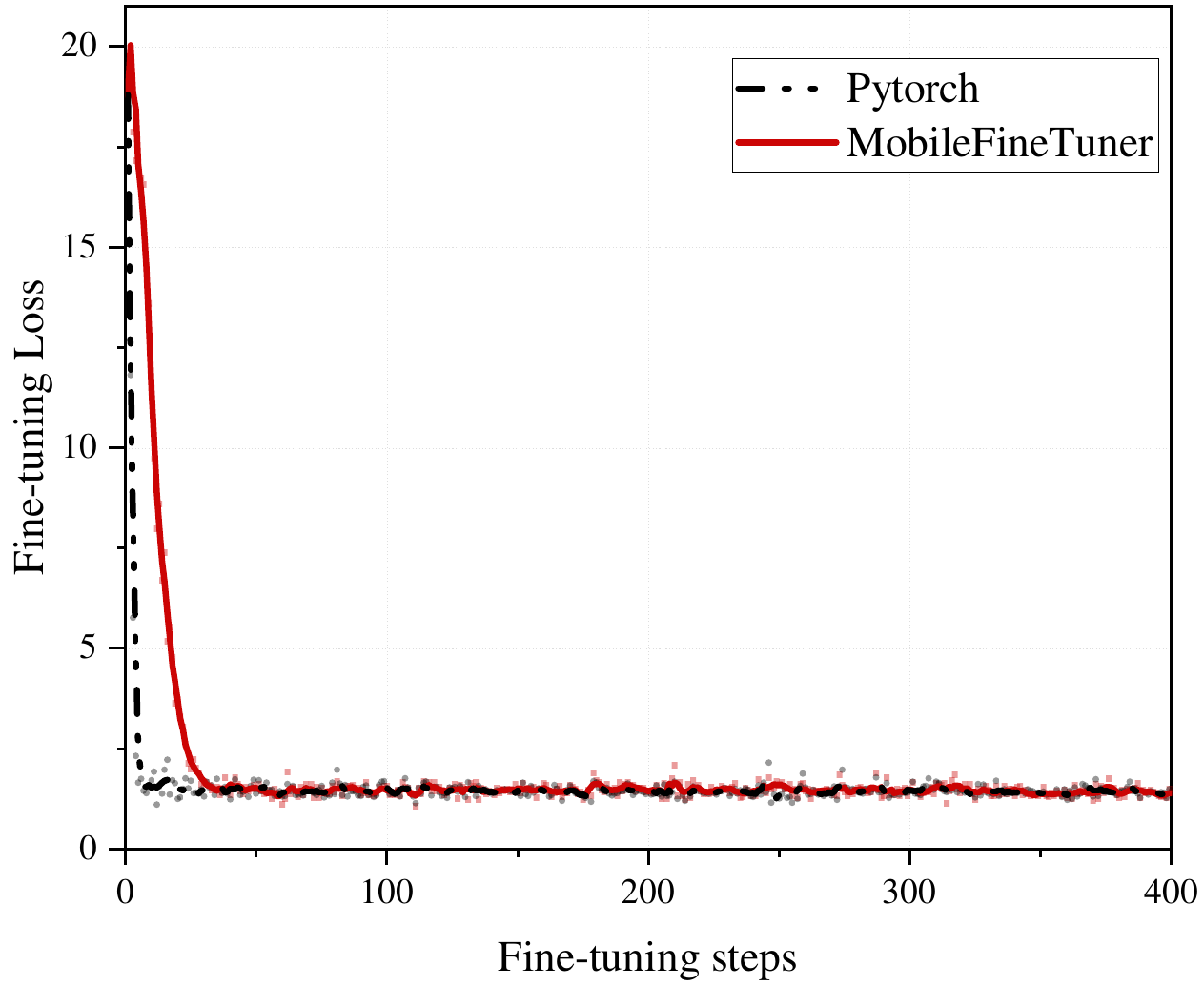}
{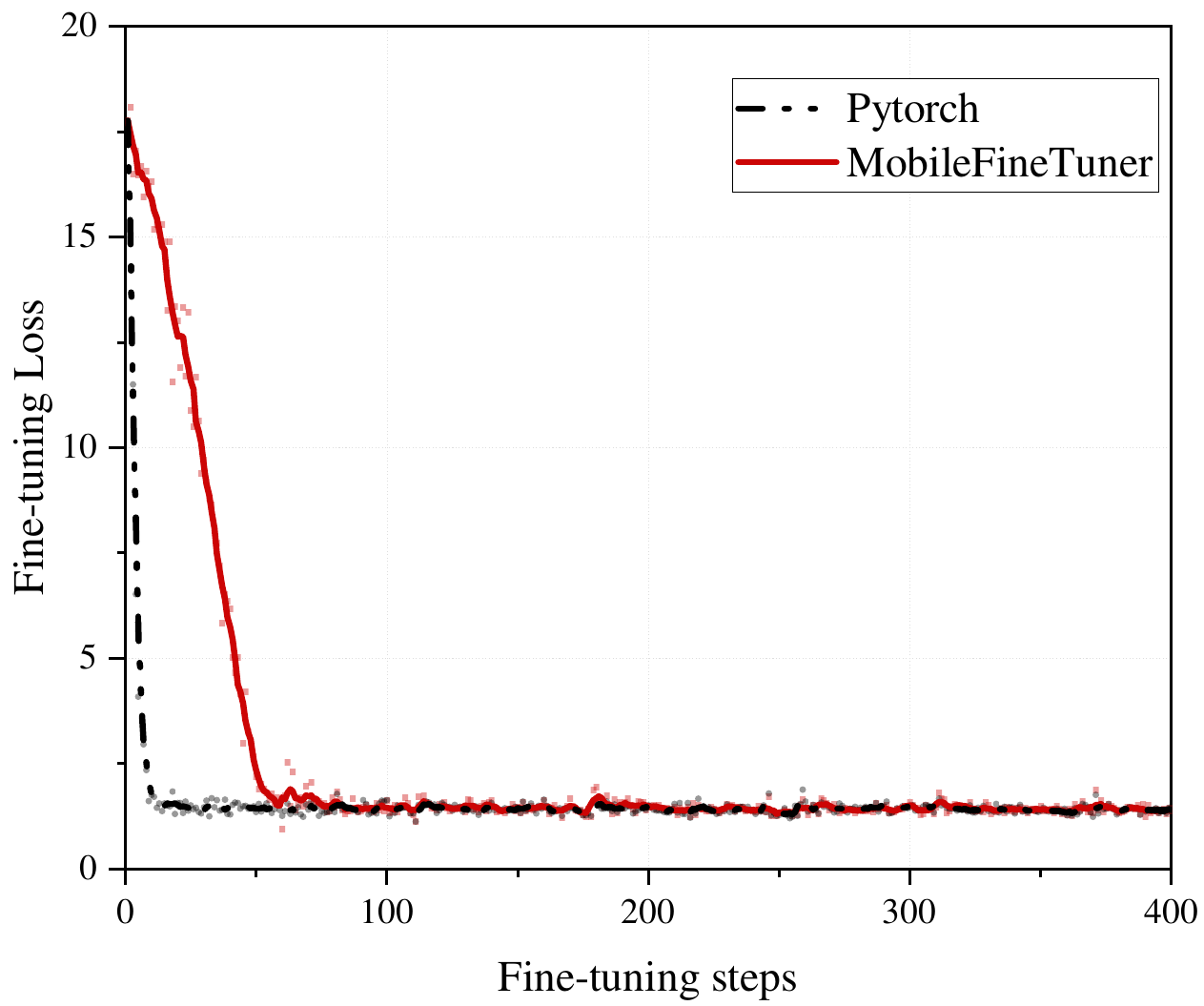}

\runtimegroupfour
{MMLU}
{fig:runtime-loss-mmlu-seq256}
{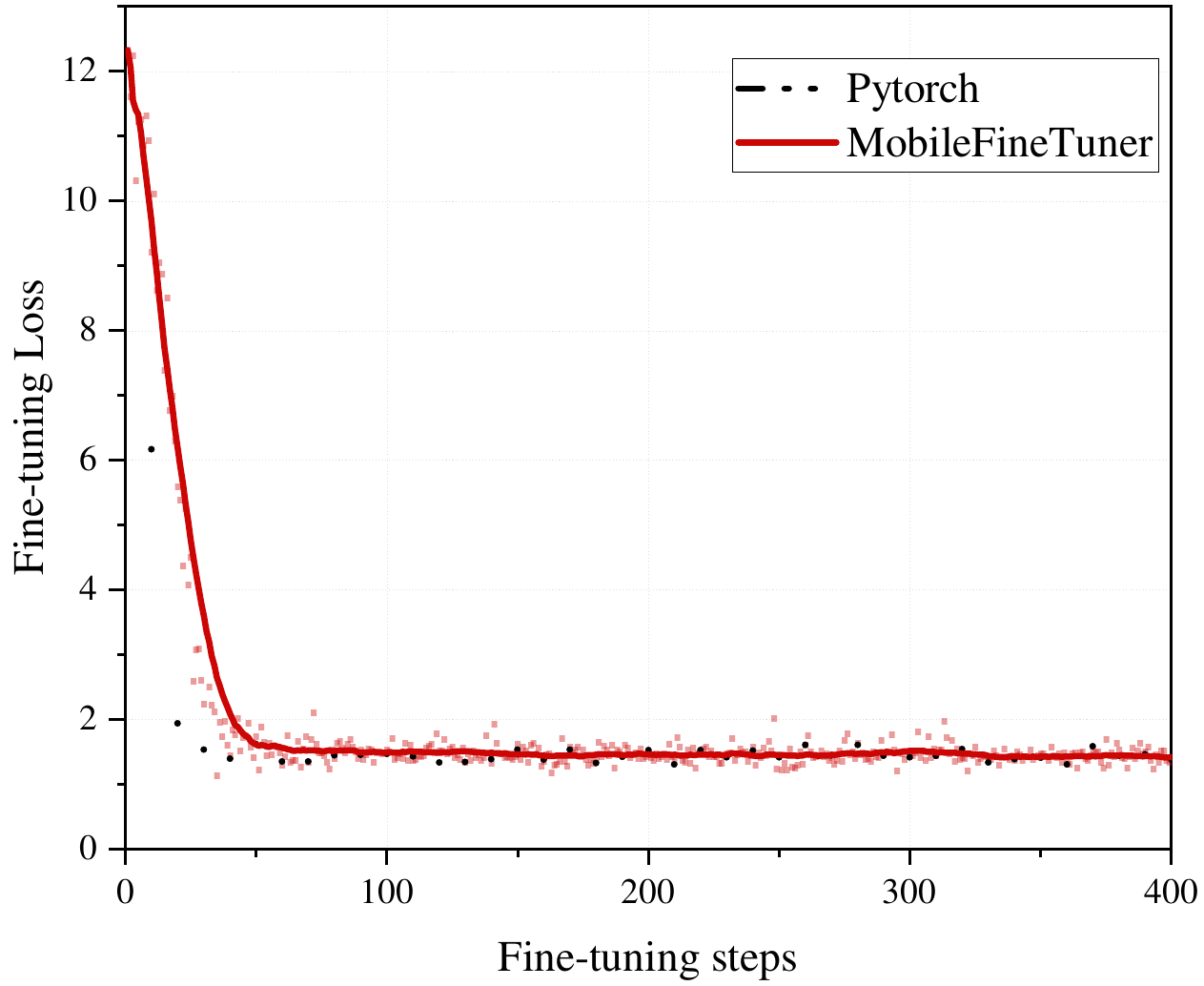}
{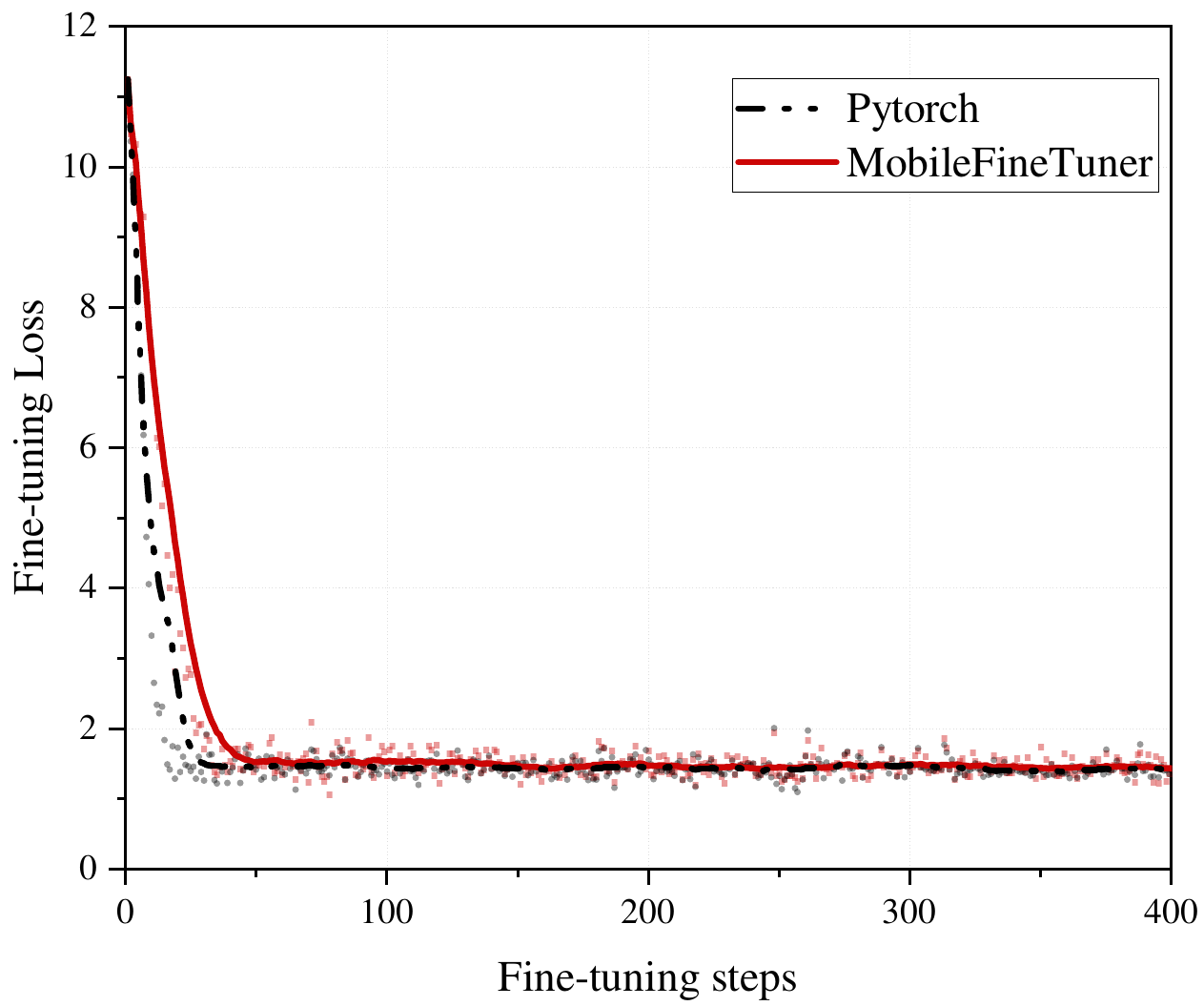}
{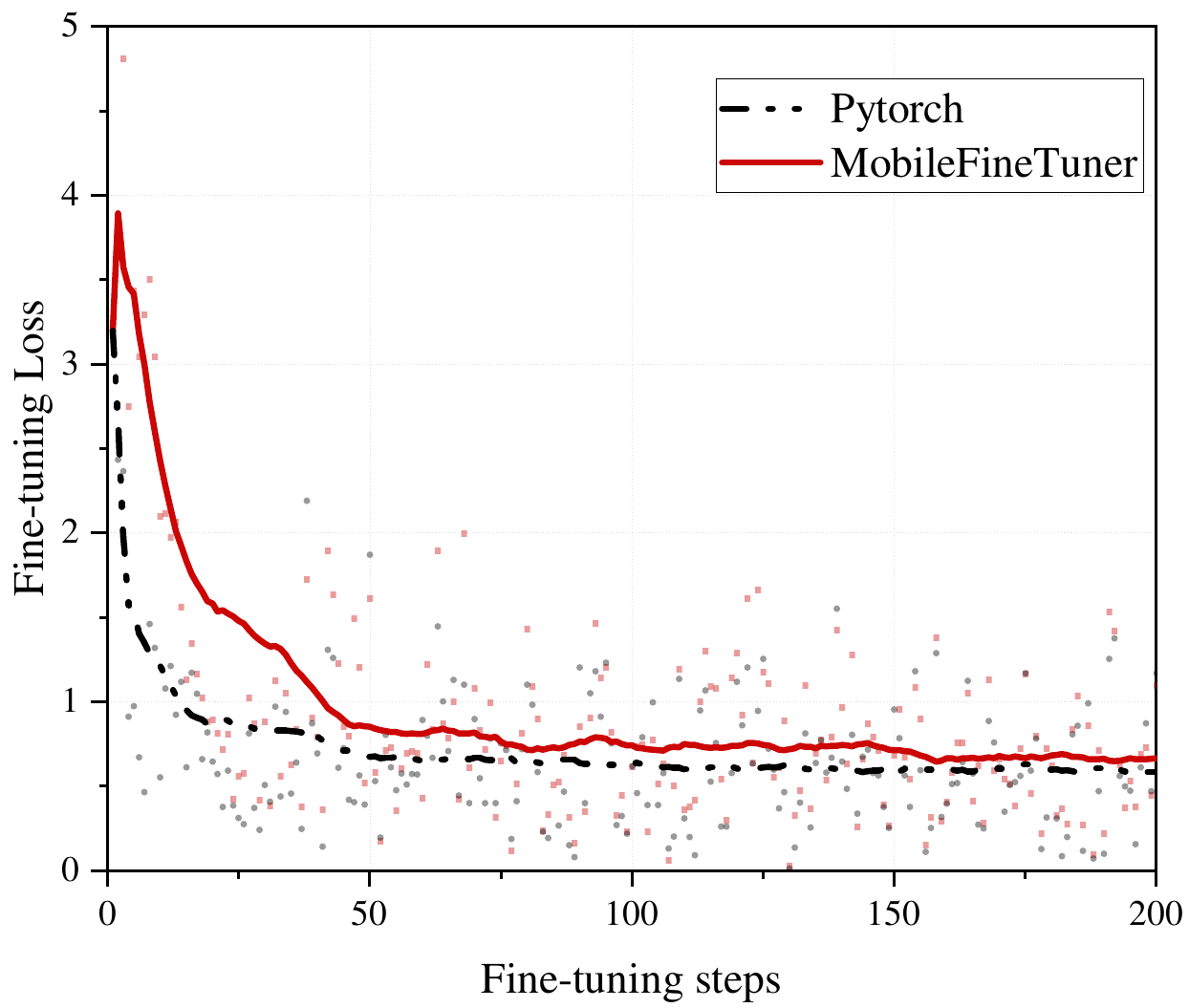}
{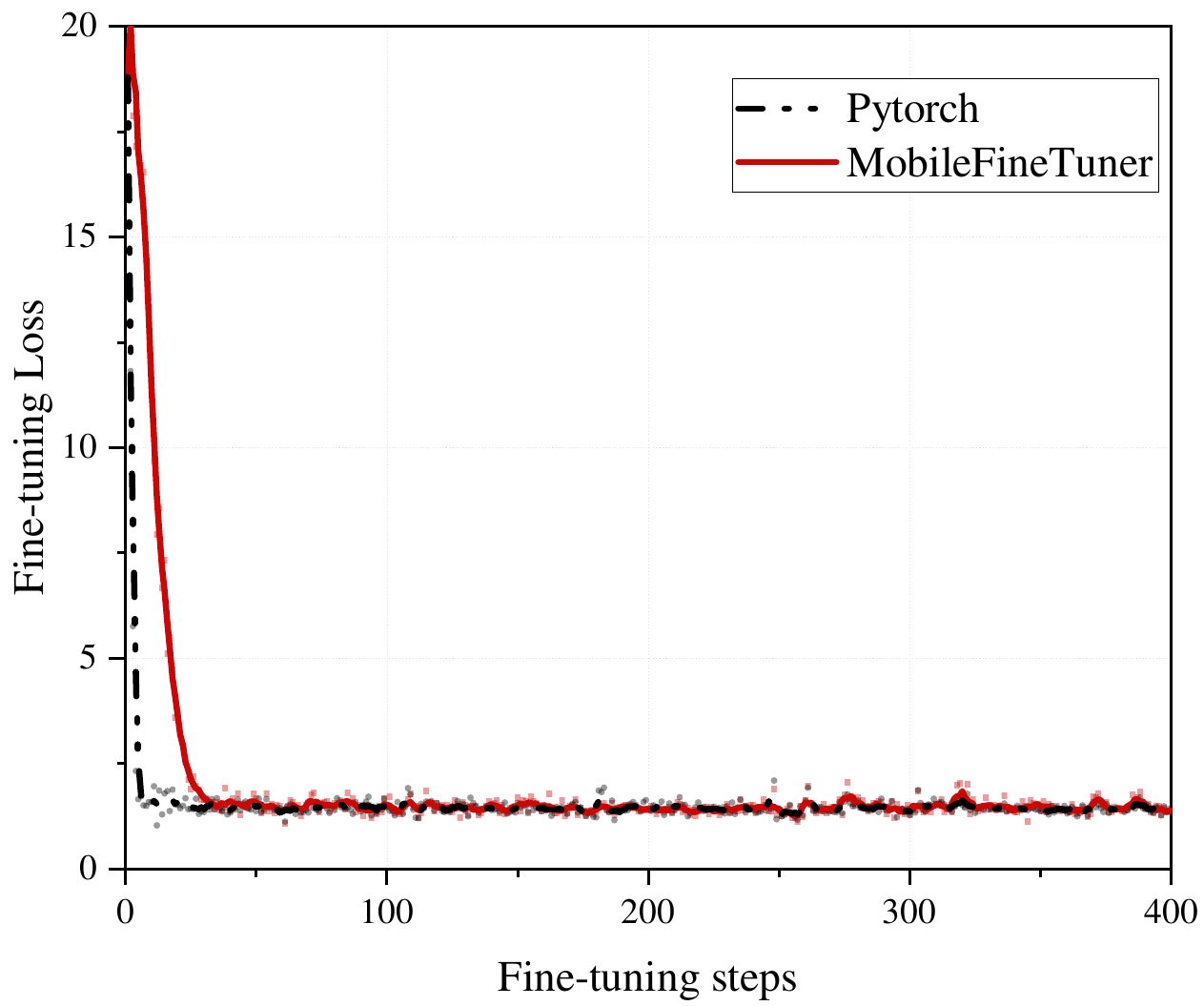}

\runtimegroupfive
{PIQA}
{fig:runtime-loss-piqa-seq128}
{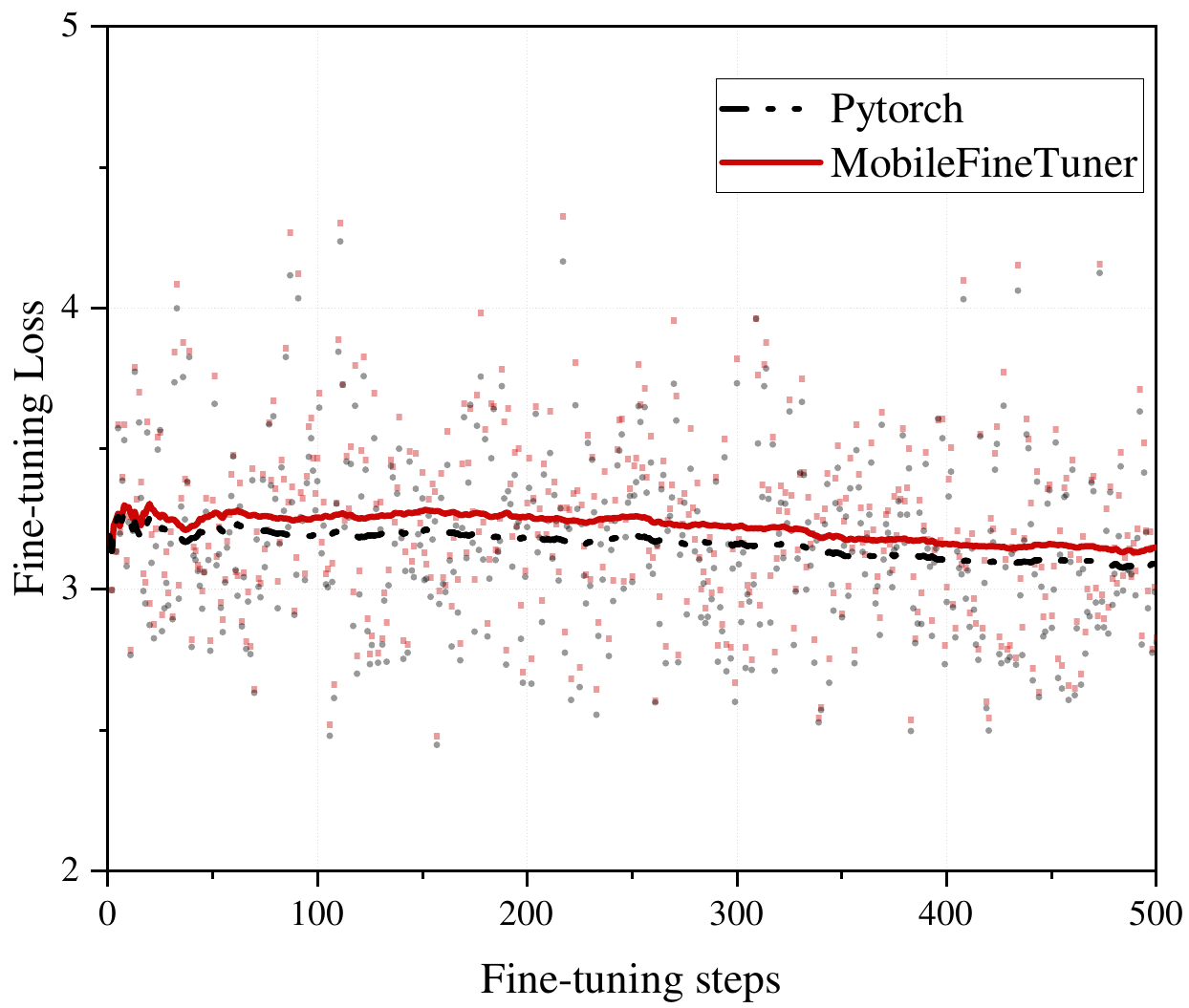}
{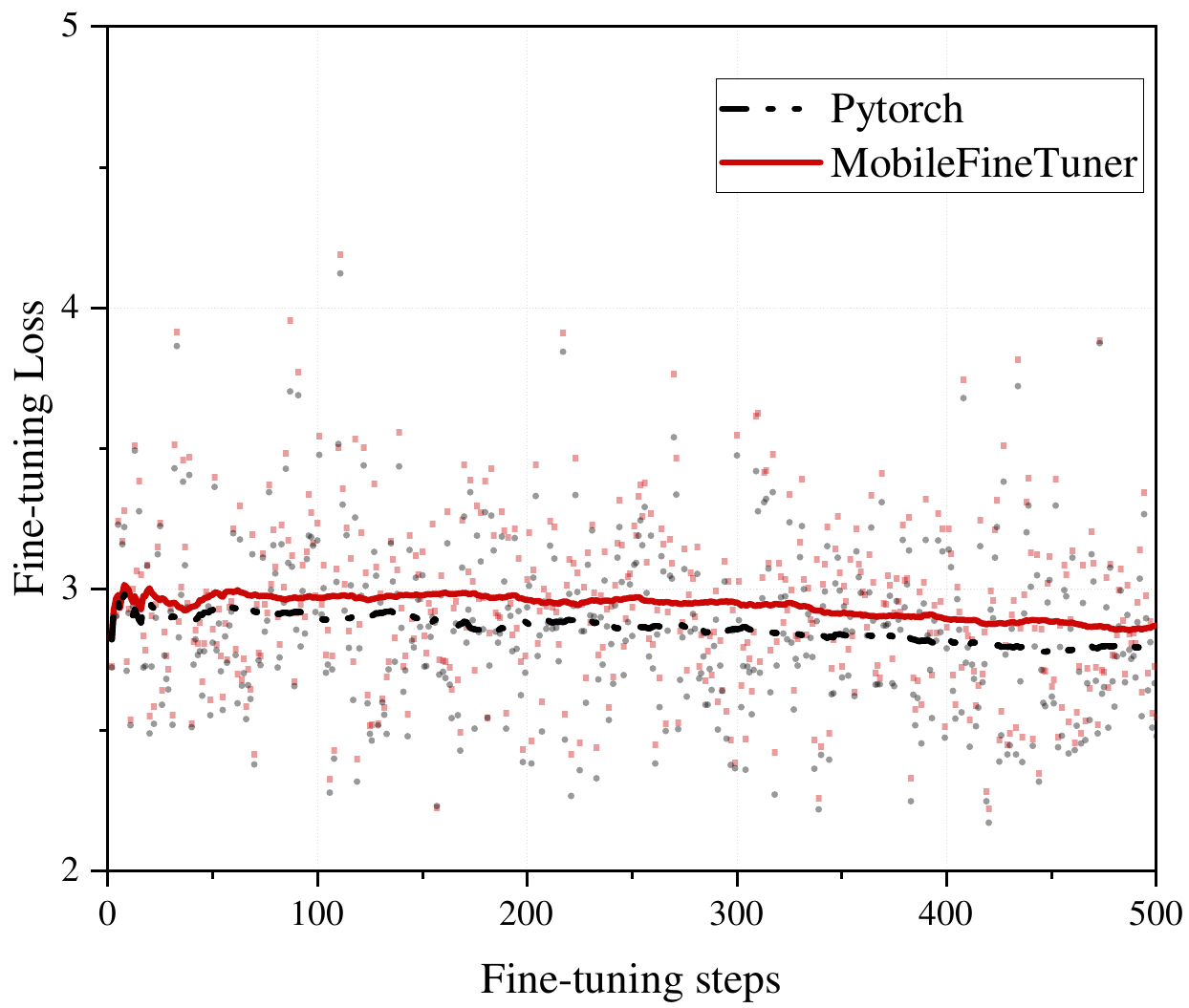}
{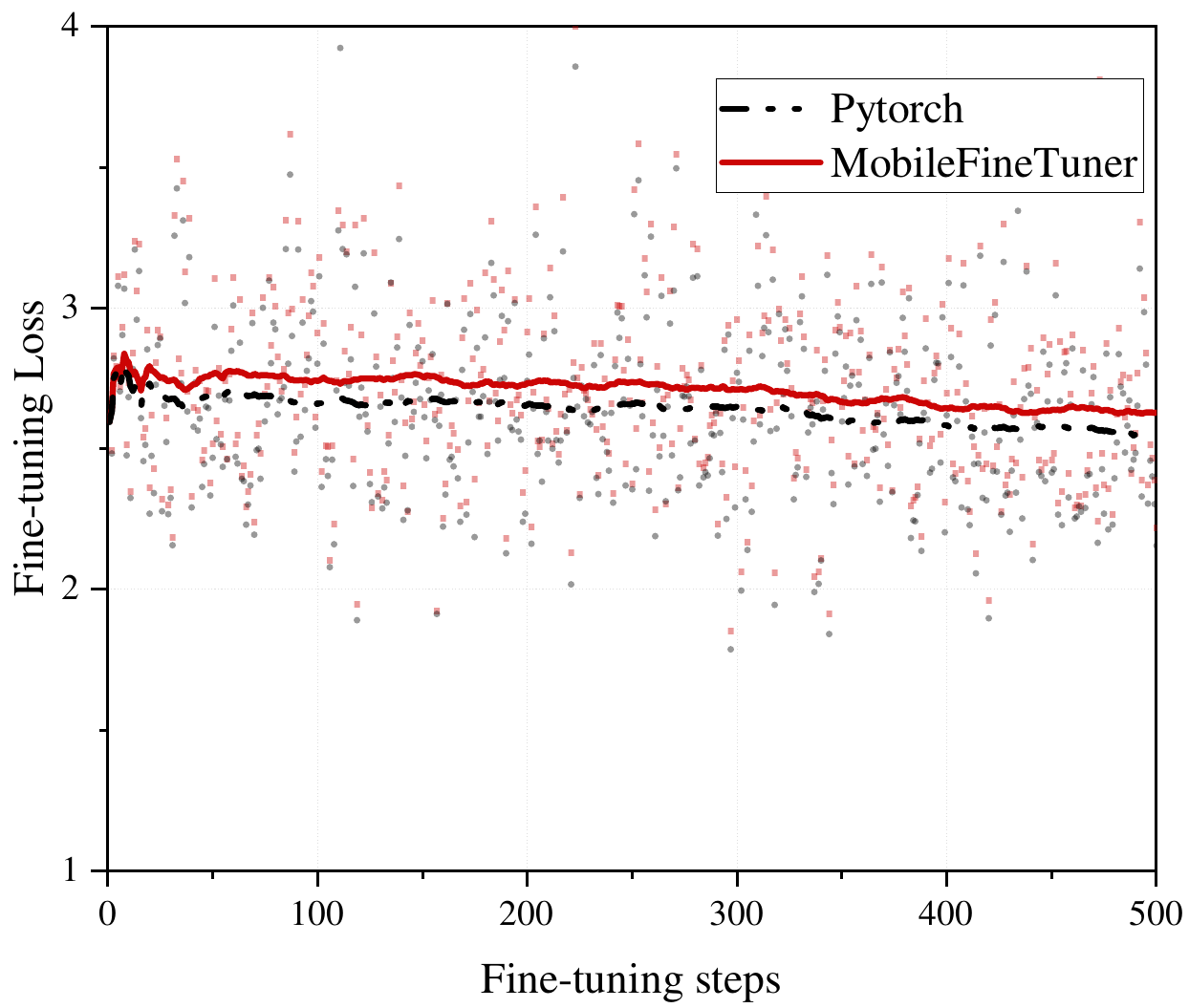}
{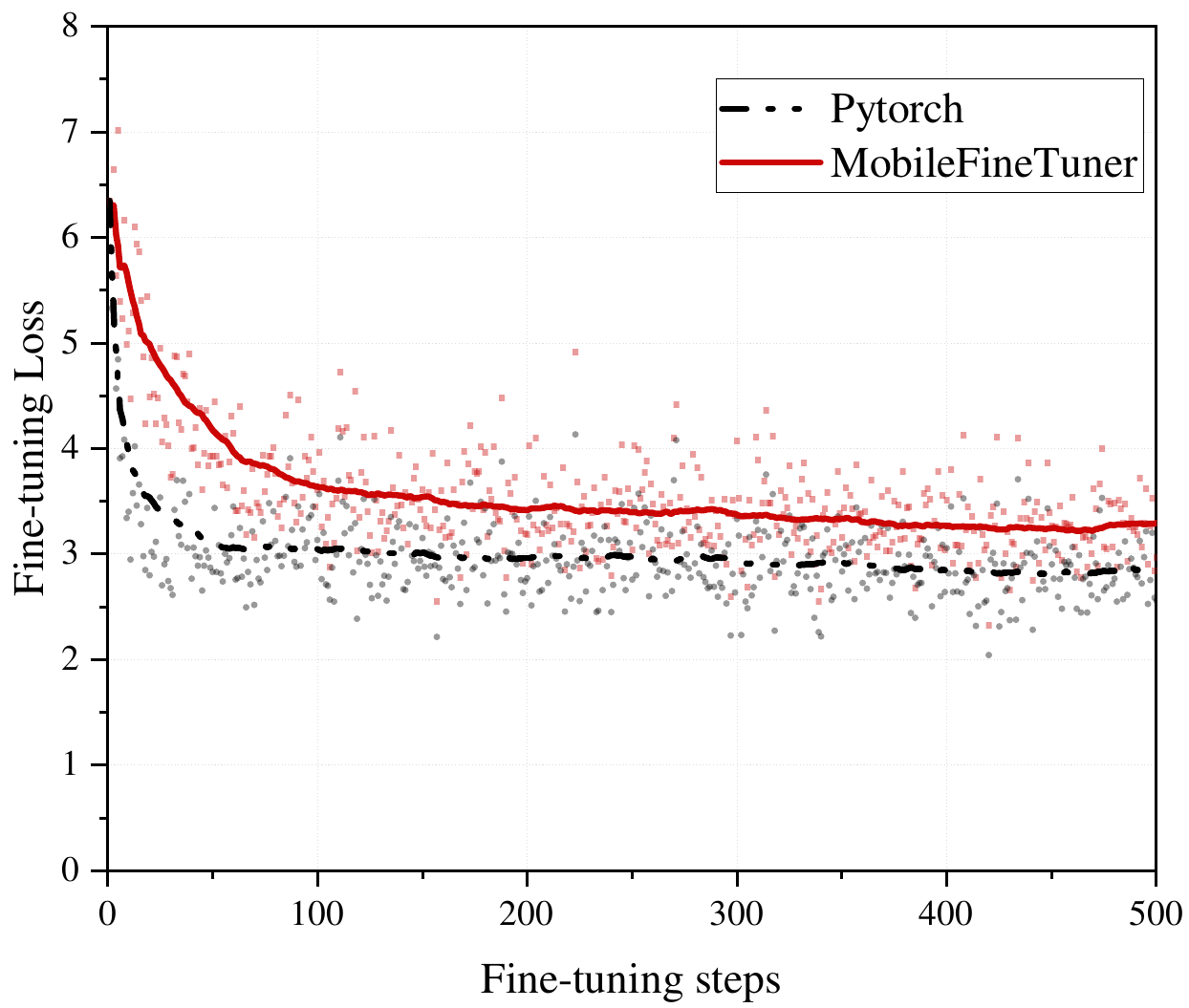}
{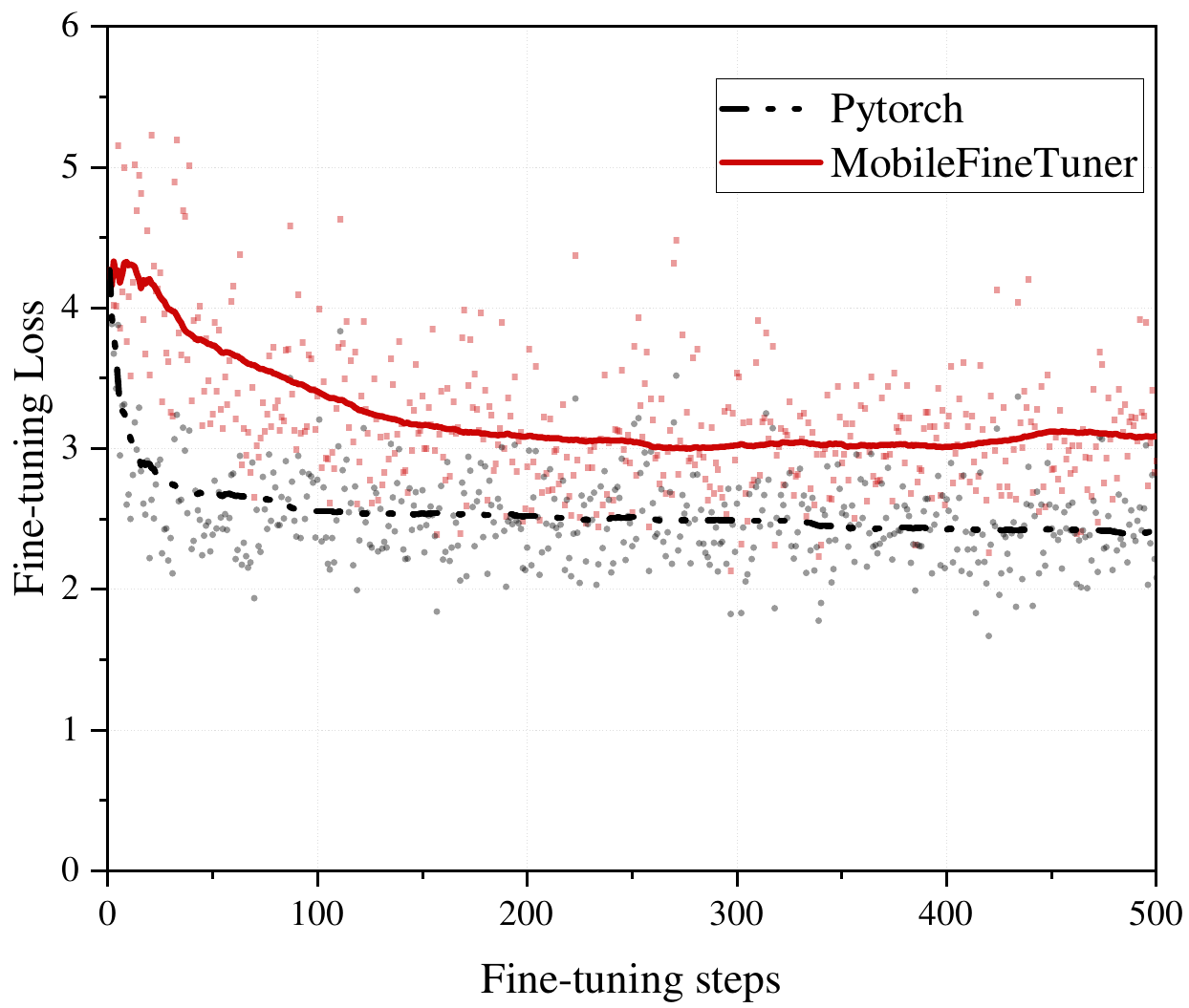}

\runtimegroupfour
{PIQA}
{fig:runtime-loss-piqa-seq256}
{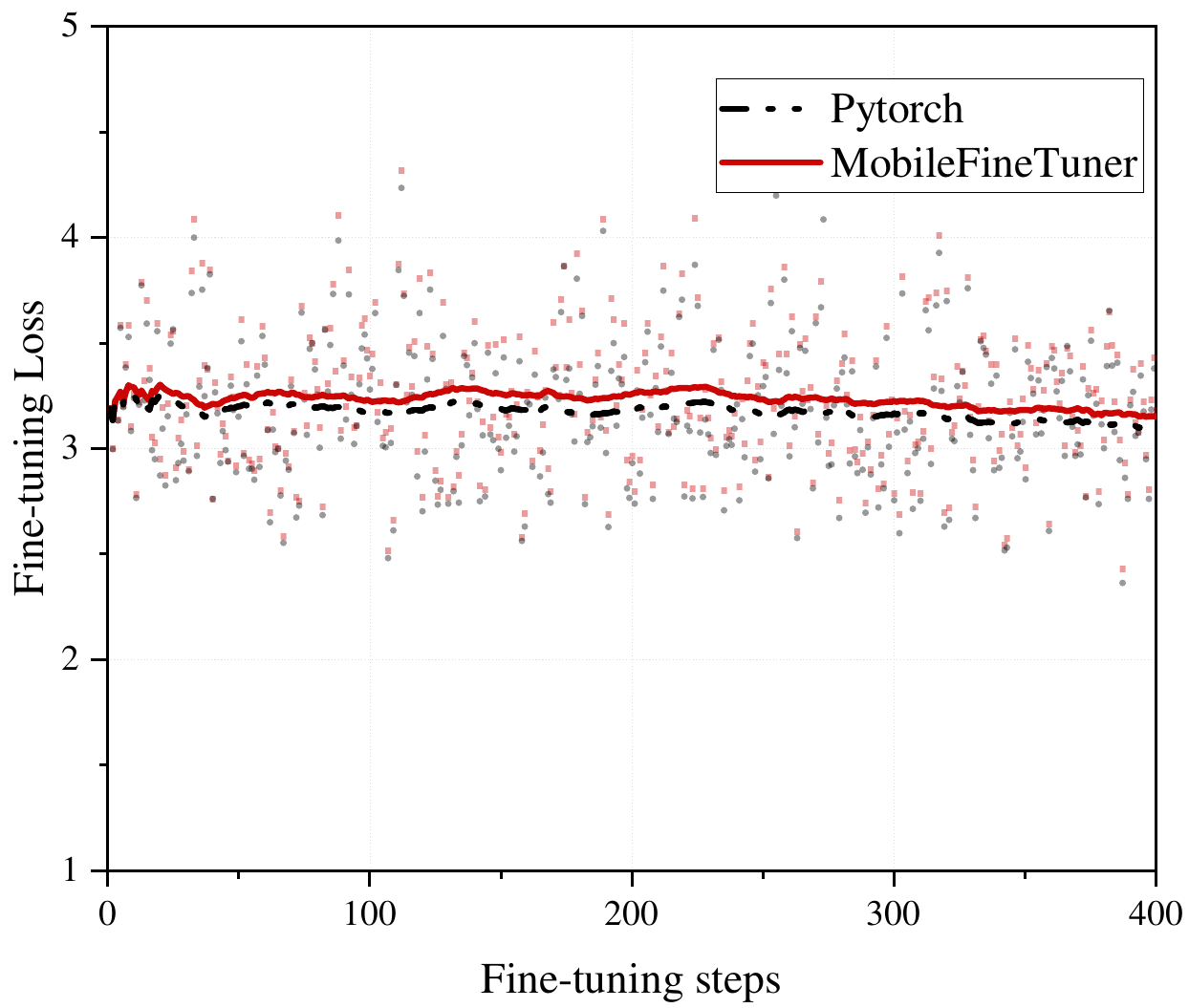}
{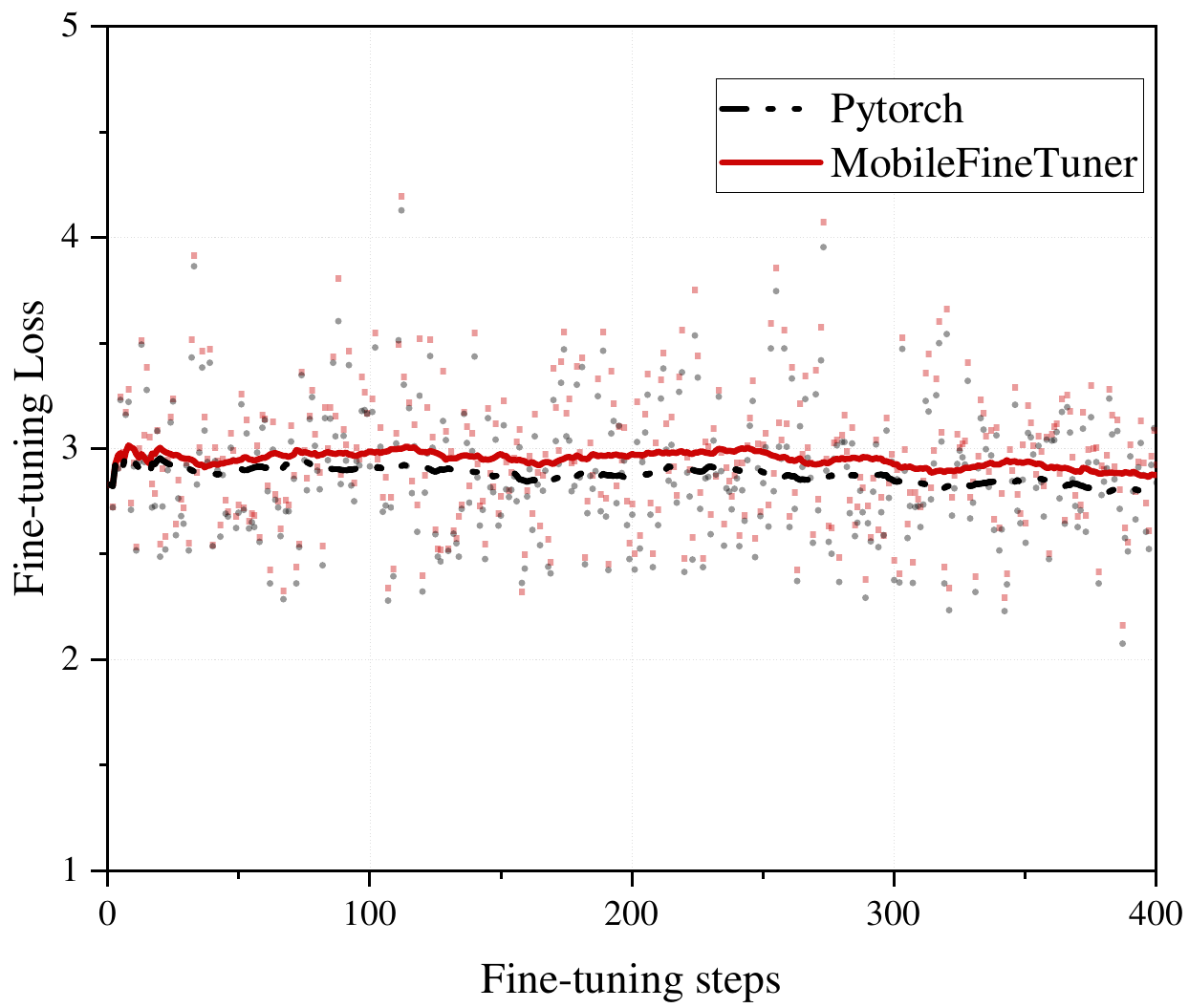}
{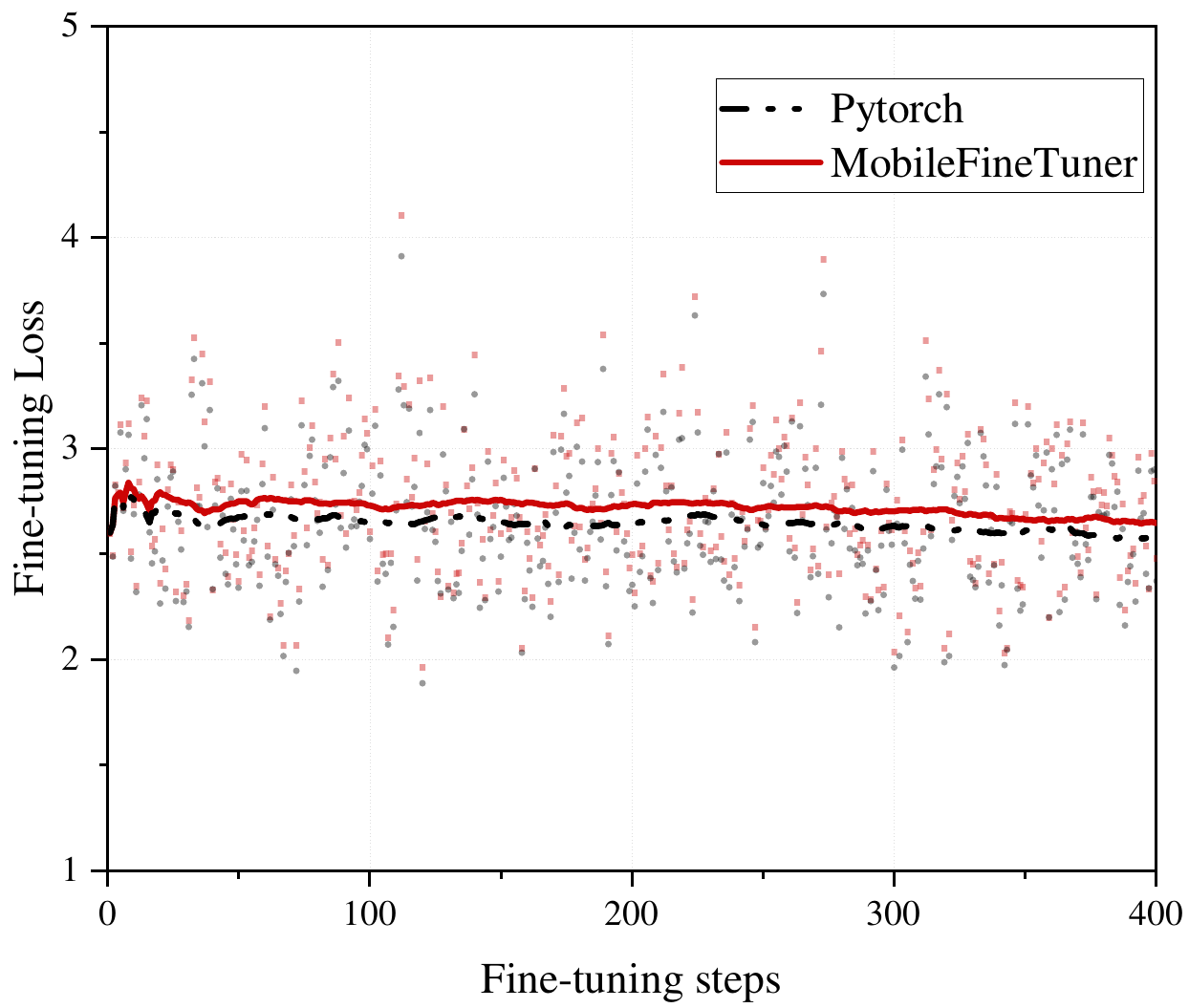}
{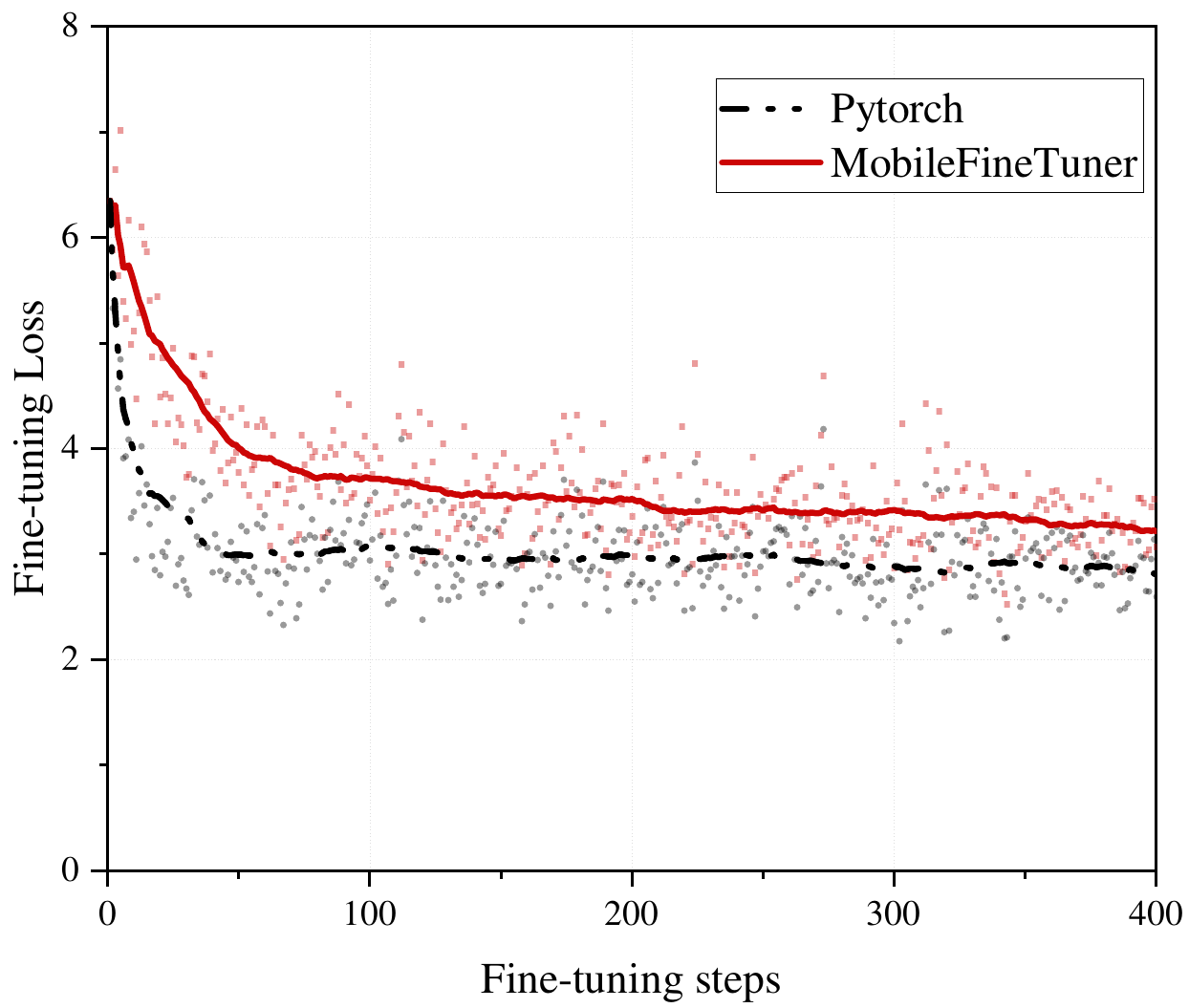}

\runtimegroupfive
{WikiText-2}
{fig:runtime-loss-wiki-seq128}
{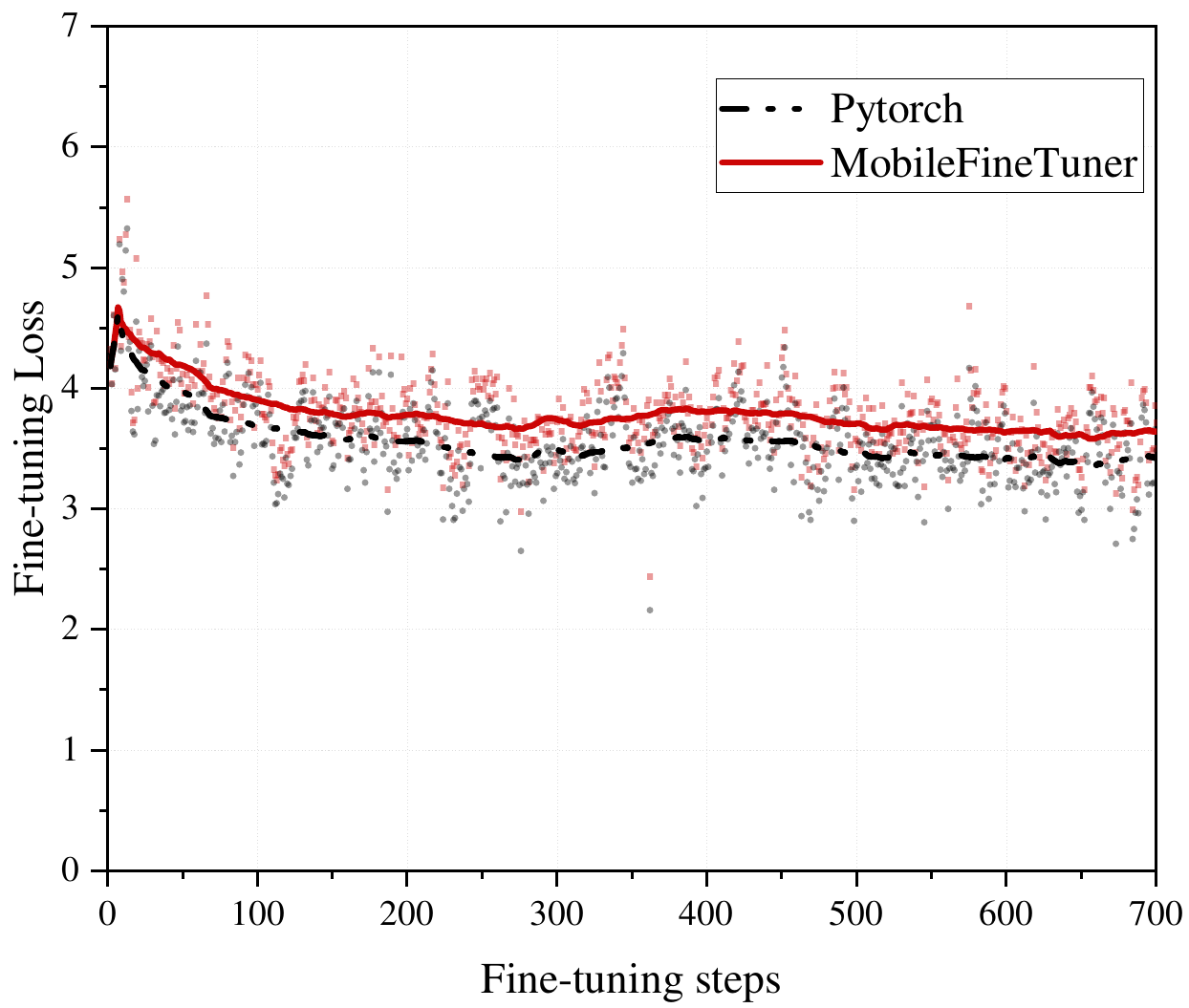}
{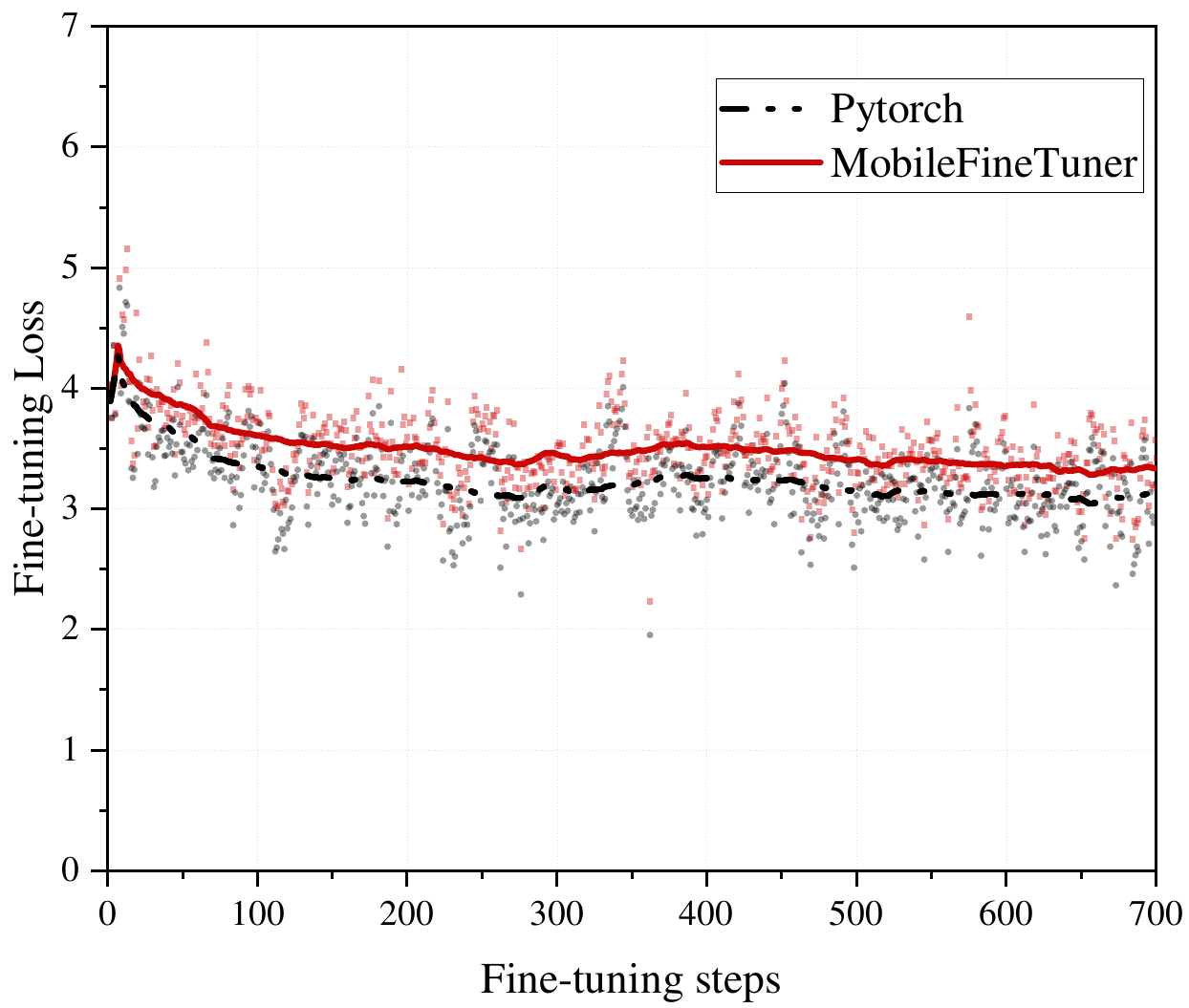}
{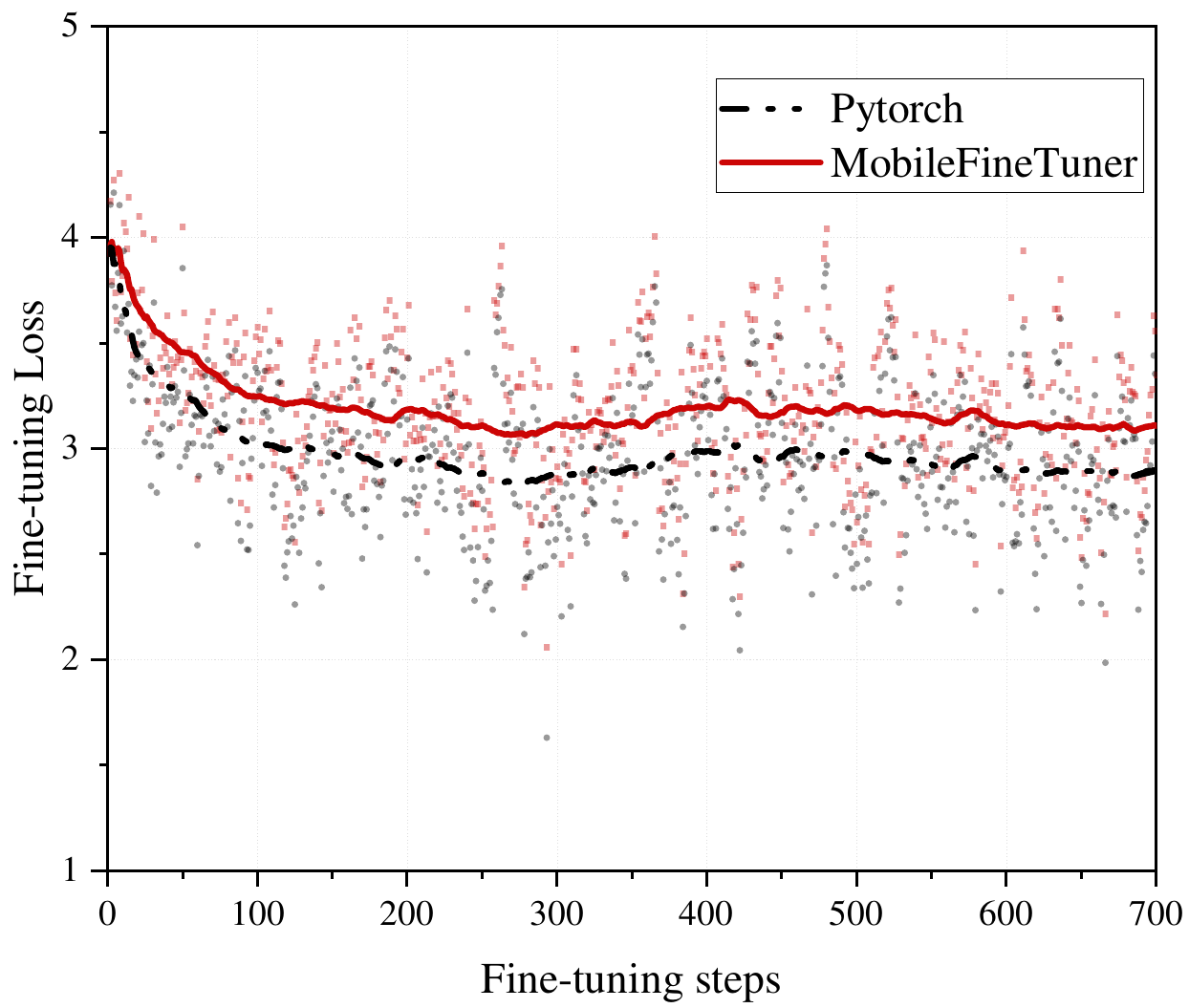}
{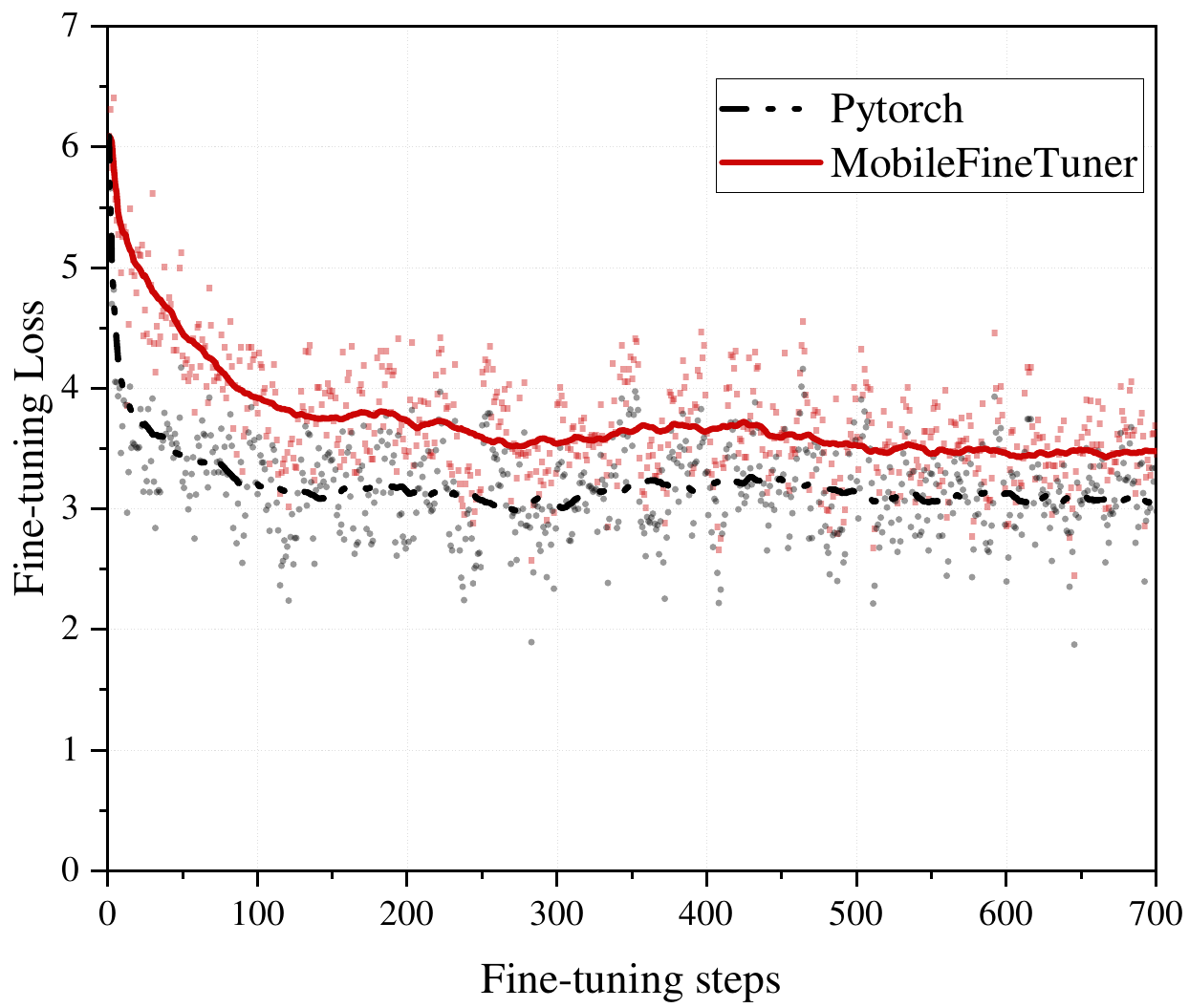}
{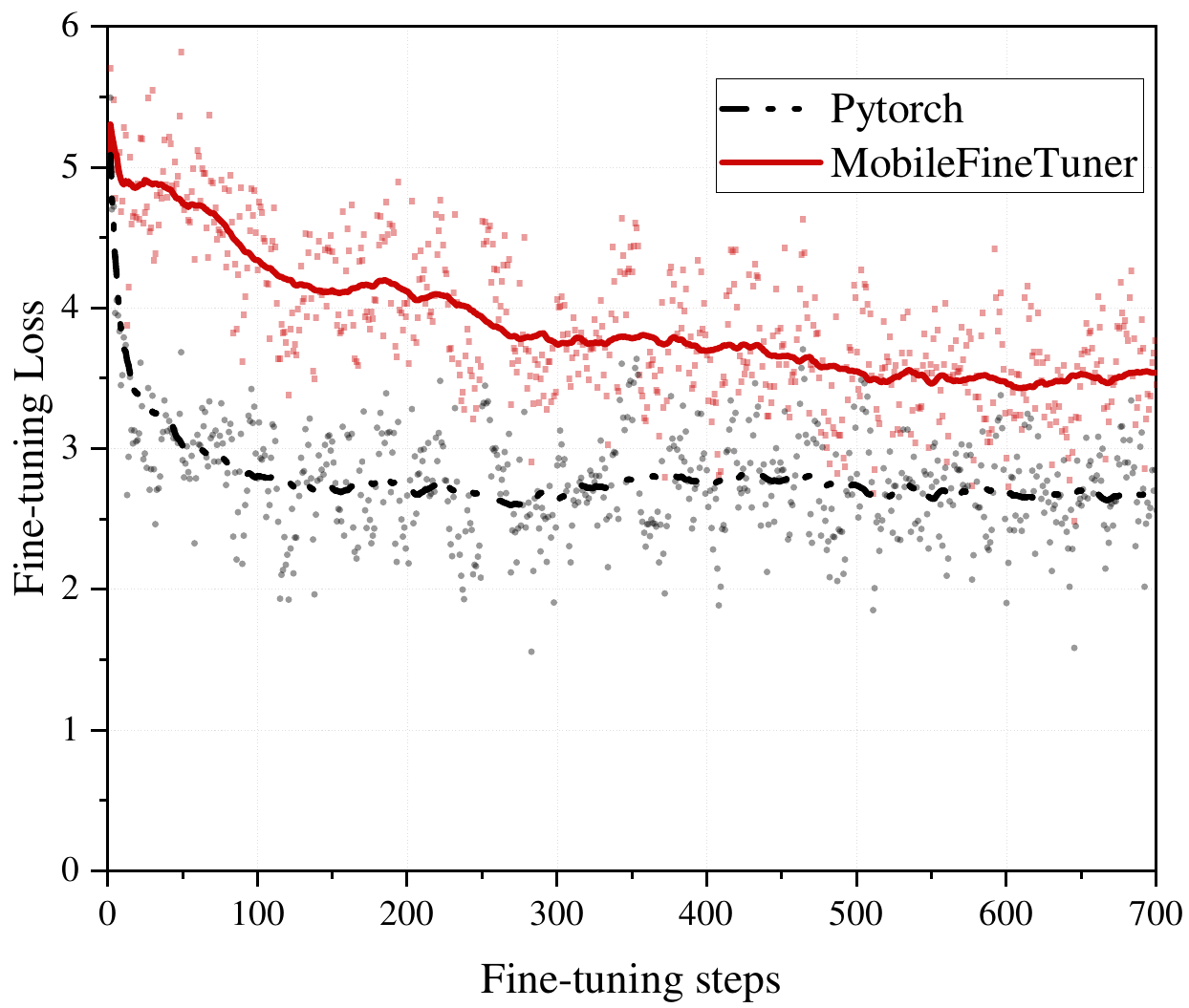}

\runtimegroupfour
{WikiText-2}
{fig:runtime-loss-wiki-seq256}
{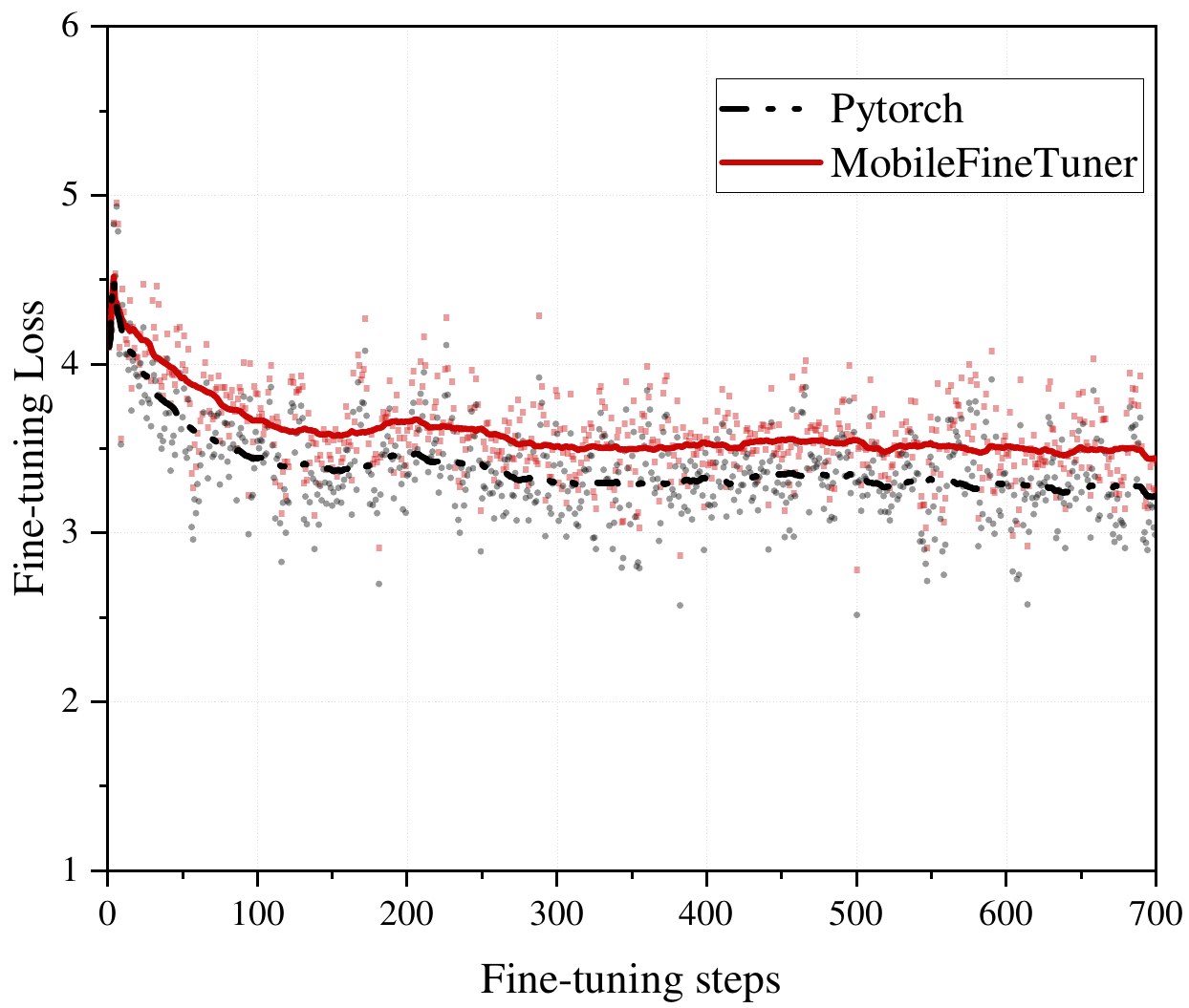}
{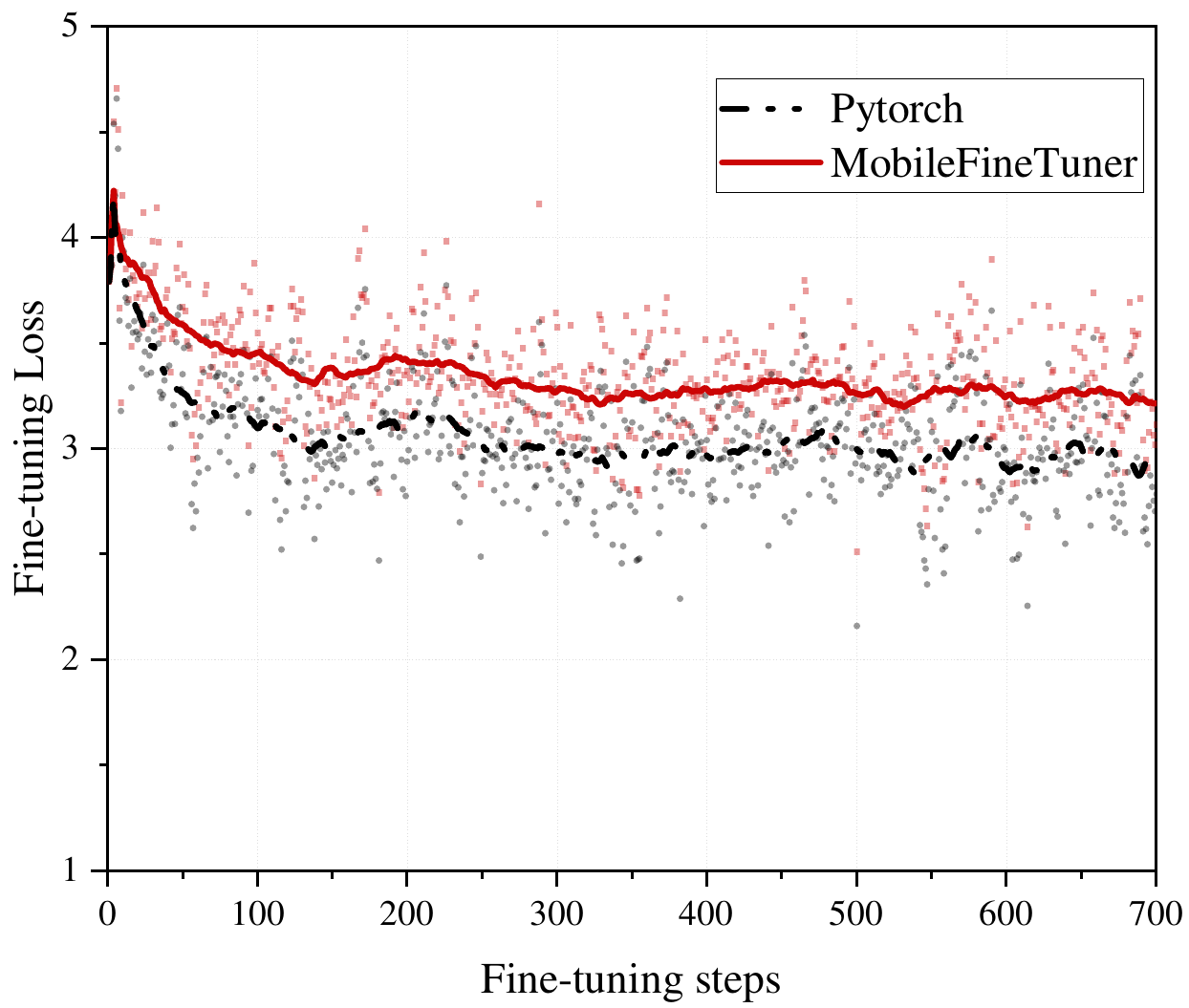}
{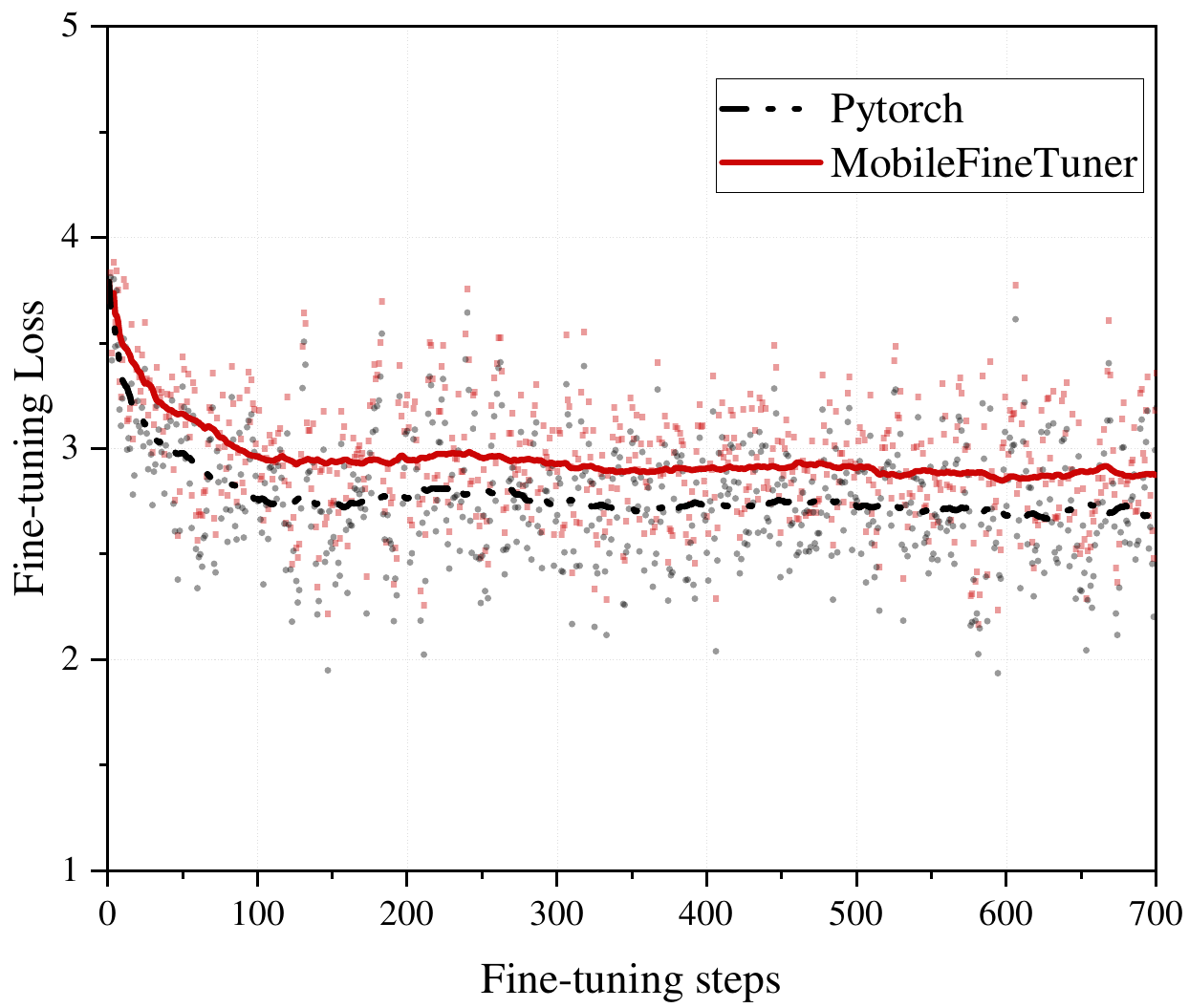}
{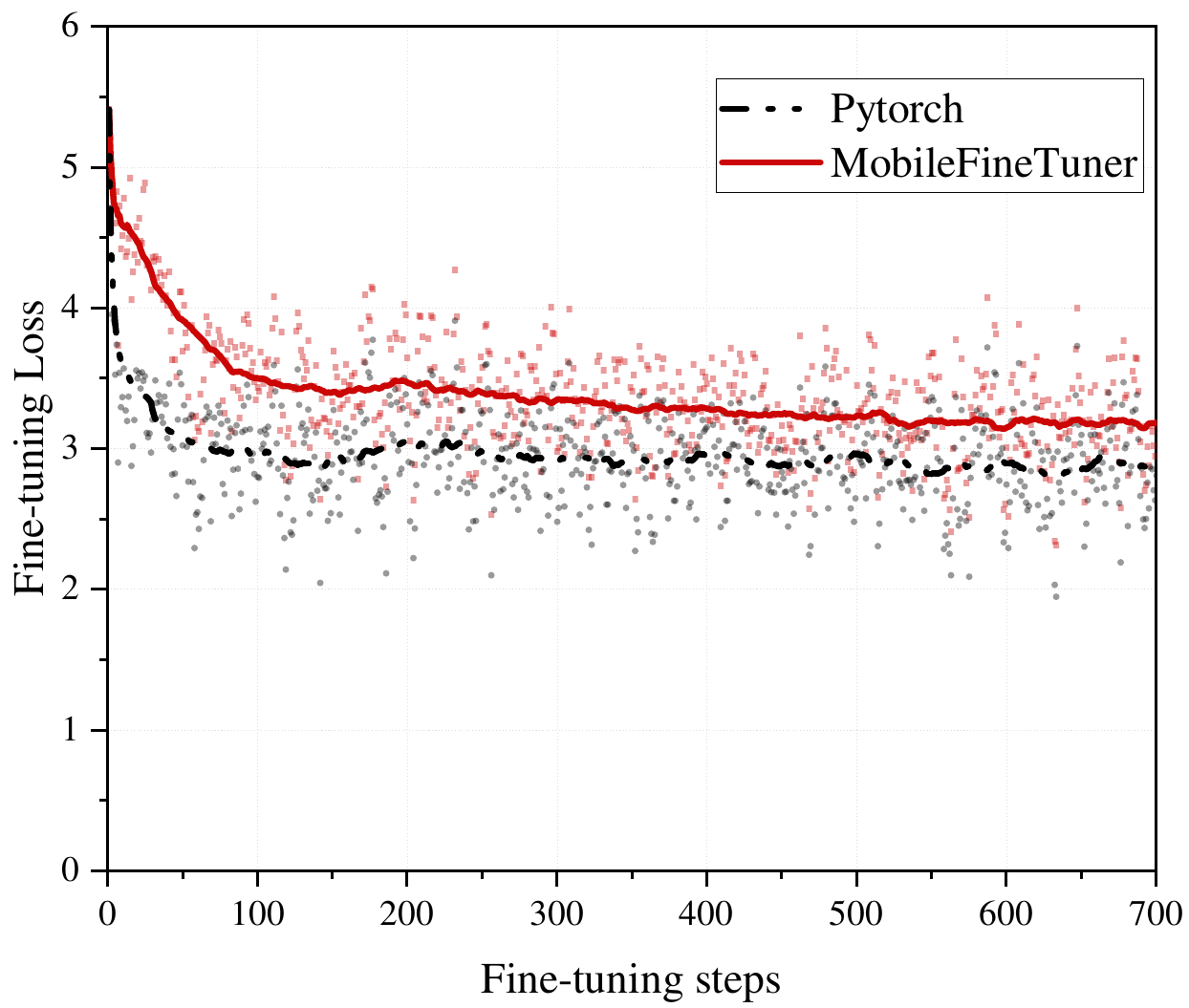}

\FloatBarrier
\clearpage

\noindent
\begin{minipage}{\textwidth}
\section{MobileFineTuner Screen Record during Fine-tuning Process}
\label{appD}

\centering

\begin{minipage}{0.28\textwidth}
    \centering
    \includegraphics[width=\linewidth]{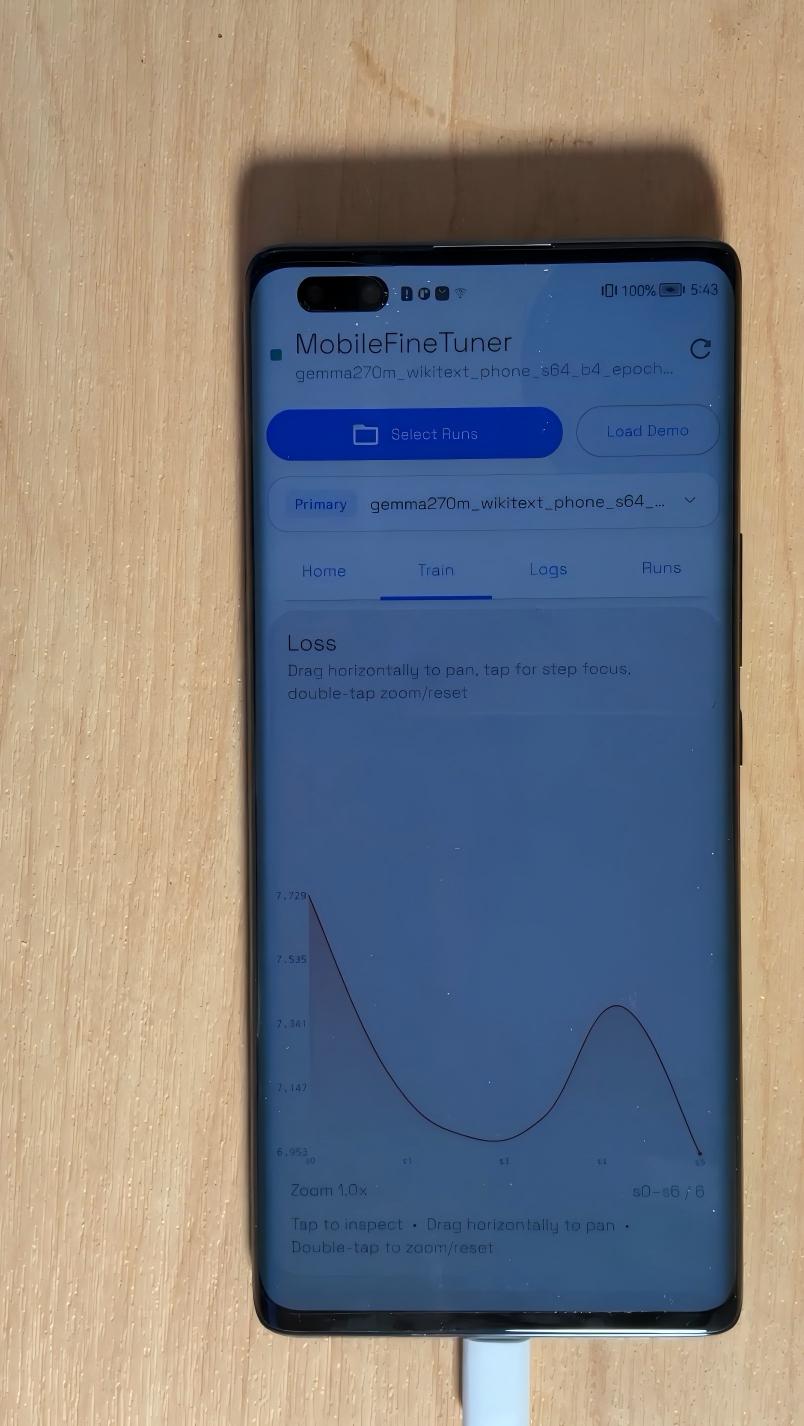}
    \vspace{2pt}
    {\small Timestamp: 05:43 am}
\end{minipage}
\hfill
\begin{minipage}{0.28\textwidth}
    \centering
    \includegraphics[width=\linewidth]{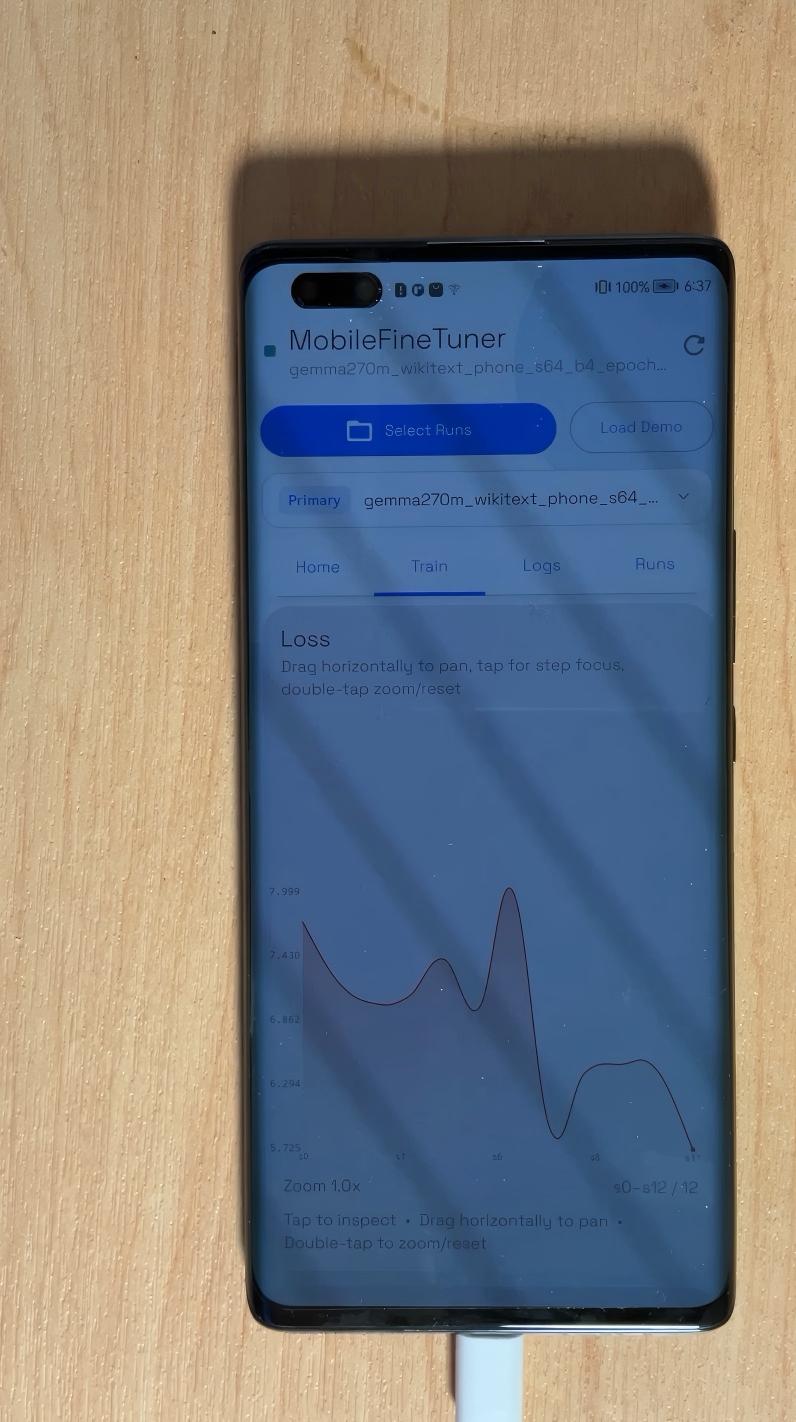}
    \vspace{2pt}
    {\small Timestamp: 06:37 am}
\end{minipage}
\hfill
\begin{minipage}{0.28\textwidth}
    \centering
    \includegraphics[width=\linewidth]{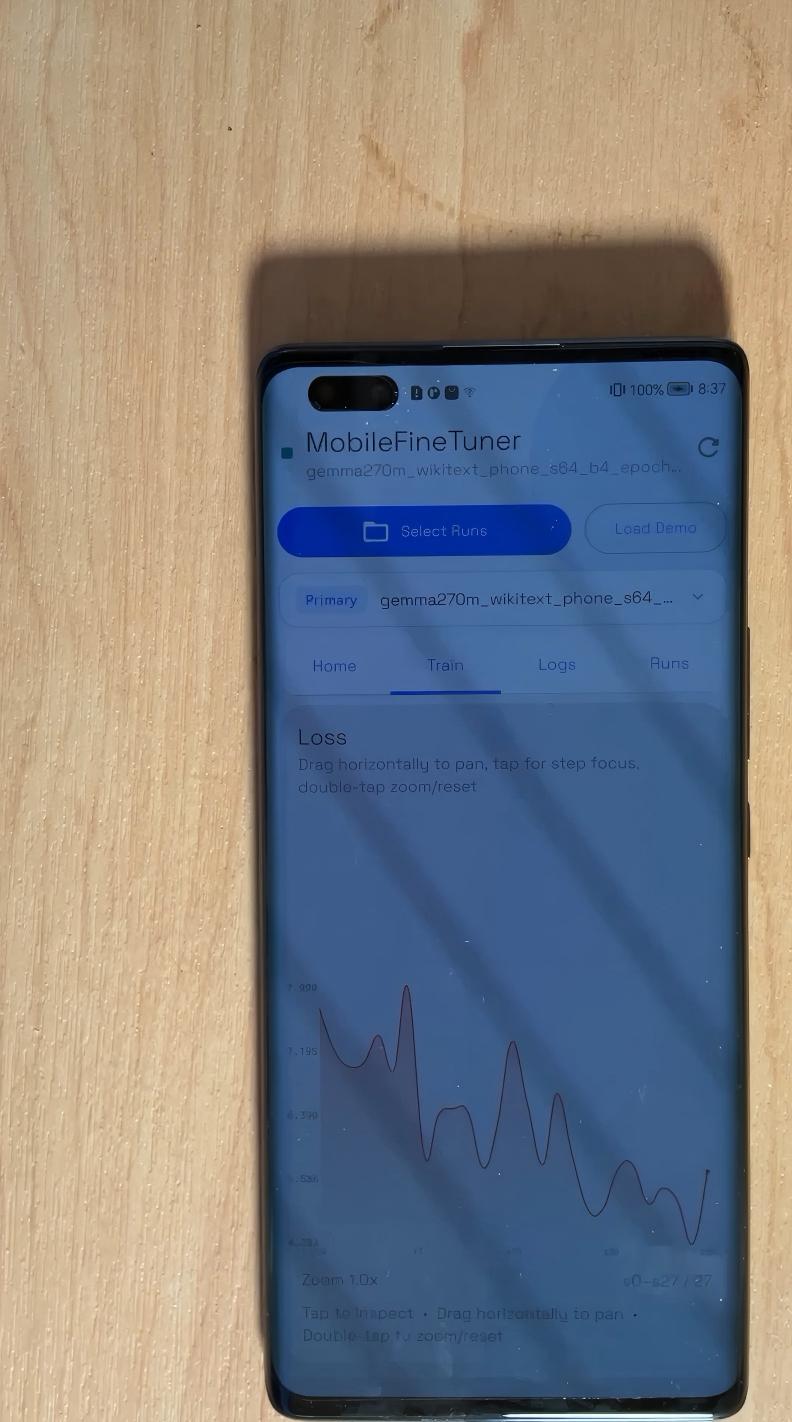}
    \vspace{2pt}
    {\small Timestamp: 08:37 am}
\end{minipage}

\vspace{2mm}

\begin{minipage}{0.28\textwidth}
    \centering
    \includegraphics[width=\linewidth]{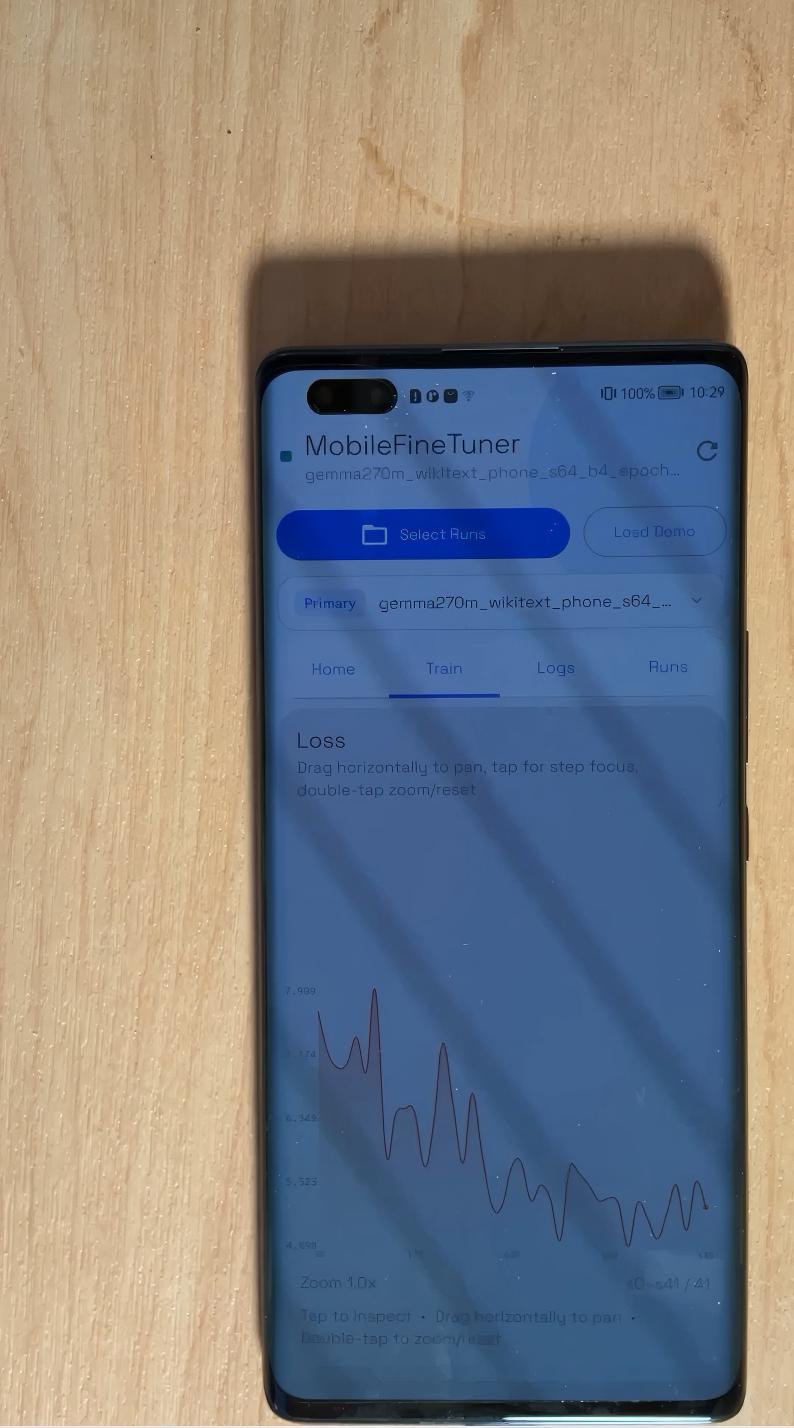}
    \vspace{2pt}
    {\small Timestamp: 10:29 am}
\end{minipage}
\hfill
\begin{minipage}{0.28\textwidth}
    \centering
    \includegraphics[width=\linewidth]{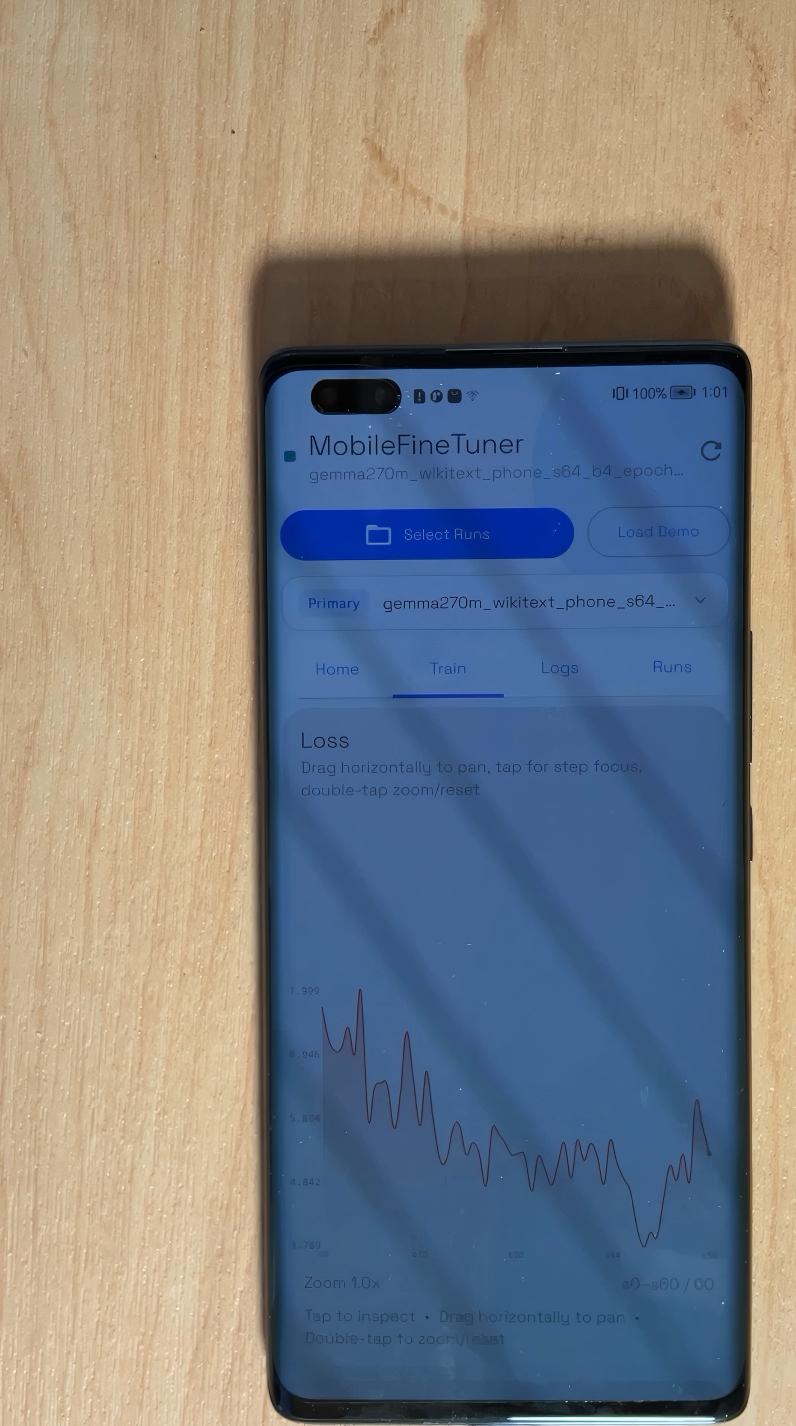}
    \vspace{2pt}
    {\small Timestamp: 1:01 pm}
\end{minipage}
\hfill
\begin{minipage}{0.28\textwidth}
    \centering
    \includegraphics[width=\linewidth]{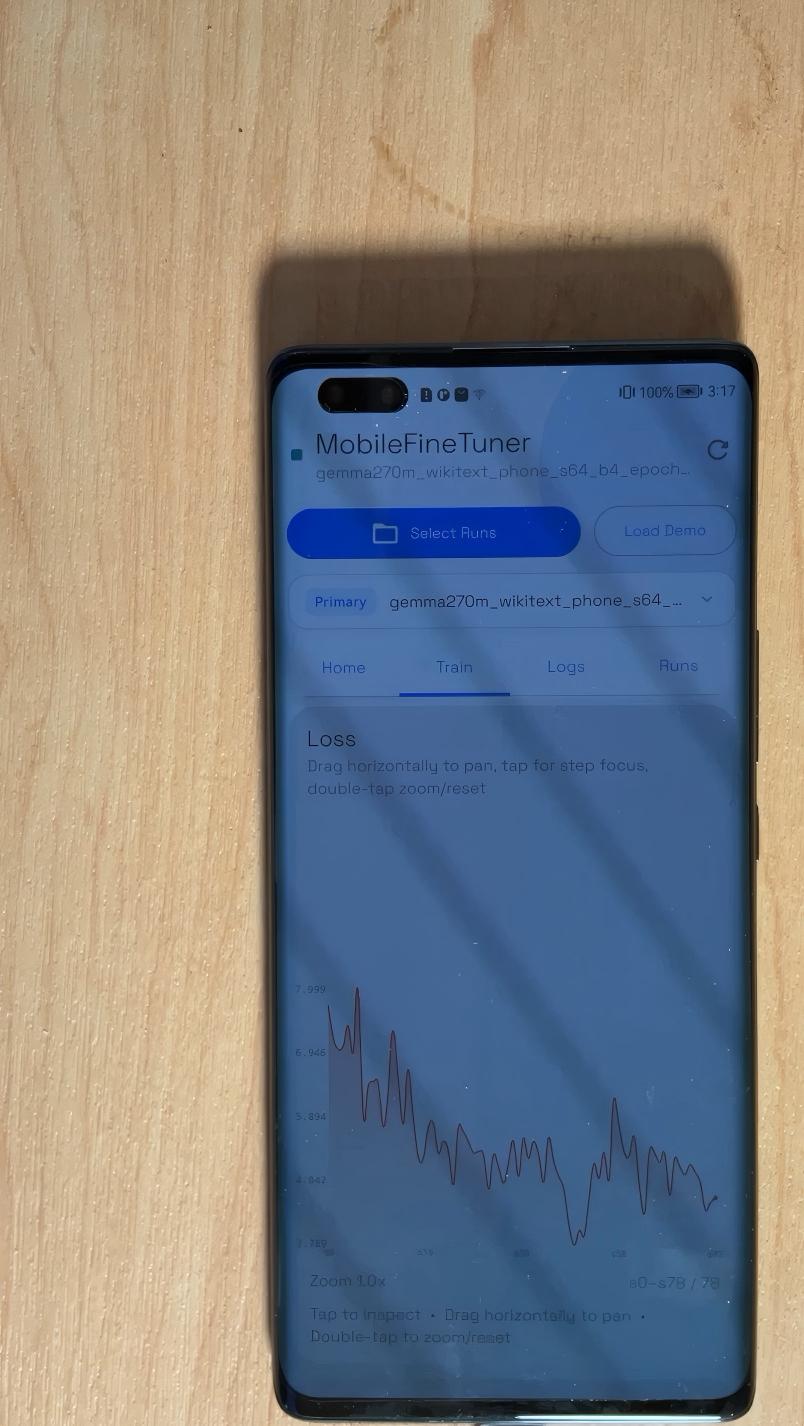}
    \vspace{2pt}
    {\small Timestamp: 03:17 pm}
\end{minipage}

\vspace{2mm}

\refstepcounter{figure}
\label{fig:mobilefinetuner_screen_record_gemma270m_wikitext_seq128}
{\small \textbf{Figure~\thefigure:} MobileFineTuner screen record during the fine-tuning process on Gemma3-270M with WikiText-2 and sequence length 128.}

\end{minipage}

\FloatBarrier
\clearpage

\section{Campus Health Agent QA Samples}
\label{appE}

\begin{table*}[h]
\centering
\caption{Representative QA samples from CHQA.}
\label{tab:chqa_samples}
\small
\begin{tabular}{p{0.15\textwidth}p{0.31\textwidth}p{0.22\textwidth}p{0.26\textwidth}}
\toprule
\textbf{Category} & \textbf{Historical Records} & \textbf{Question} & \textbf{Answer} \\
\midrule

\textbf{Activity Summary} &
[Recent records include 4 logged days. The user averaged 11,154 steps/day, with a peak of 15,461 steps. Recent movement is about 43\% higher than the previous stretch. Average active calories are 278 kcal/day.] &
Have I been moving enough recently? &
Yes. Your recent activity level looks strong, with an average of over 11,000 steps/day and a clear increase compared with your previous stretch. A good next step is to keep the pace steady rather than pushing for another peak immediately. \\

\midrule

\textbf{Goal Adjustment} &
[Recent records include 3 logged days. The user averaged 11,269 steps/day, with a peak of 15,461 steps. Recent movement is about 78\% higher than the previous stretch. Average active calories are 283 kcal/day.] &
If I keep it realistic, should my current step goal be higher or lower? &
A realistic goal would be around 10,000 to 10,500 steps/day. This is slightly below your recent average, so it remains achievable while still encouraging you to maintain your current activity level. \\

\midrule

\textbf{Habit Coaching} &
[Recent records include 4 logged days. The user averaged 11,154 steps/day, with a high-activity day reaching 15,461 steps. Recent movement is about 43\% higher than the previous stretch. Average active calories are 278 kcal/day.] &
Do my recent activity habits look regular? &
Your overall activity level is good, but the pattern still shows some fluctuation between regular days and peak days. For habit building, it is better to keep a stable daily floor than to rely on occasional very high-activity days. \\

\midrule

\textbf{Metric Insight} &
[Recent records include 4 logged days. The user averaged 11,154 steps/day. Recent movement is about 43\% higher than the previous stretch. Average active calories are 278 kcal/day.] &
Can you interpret my recent activity intensity? &
Your recent activity intensity looks relatively high. The combination of more than 11,000 steps/day and nearly 280 active calories/day suggests that you have been consistently active, not just accumulating light movement. \\

\midrule

\textbf{Plan Recommendation} &
[Recent records include 4 logged days. The user averaged 11,154 steps/day, with a peak of 15,461 steps. Recent movement is about 43\% higher than the previous stretch. Average active calories are 278 kcal/day.] &
Based on this step pattern, how far should I run tomorrow morning? &
A conservative morning run of 1.5--2.0 km would be reasonable, with easy walking before and after. Since your recent activity is already high, the goal should be to maintain consistency rather than add too much extra load. \\

\bottomrule
\end{tabular}
\end{table*}

\end{document}